\documentclass[journal]{IEEEtran}
\IEEEoverridecommandlockouts 

\usepackage{hyperref, url, amsmath, amssymb, booktabs, graphicx, bm, color, algorithm, tabularx}
\usepackage[noend]{algpseudocode}
\usepackage[caption=false]{subfig}
\graphicspath{{images/}}

\DeclareMathOperator*{\argmax}{\arg\max}

\newcolumntype{Y}{>{\centering\arraybackslash}X}

\newfloat{datacollect}{htbp}{lok}
\floatname{datacollect}{Data Collection}

\title{\LARGE \bf Characterizing Multidimensional Capacitive Servoing for\\Physical Human-Robot Interaction}

\author{Zackory Erickson, Henry M. Clever, Vamsee Gangaram, Eliot Xing, \\ Greg Turk, C. Karen Liu, and Charles C. Kemp
\thanks{This work was supported by NSF award IIS-1514258, NSF award IIS-2024444, NSF award DGE-1545287, and the National Science Foundation Graduate Research Fellowship Program under Grant No. DGE-1148903. Dr. Kemp owns equity in and works for Hello Robot, a company commercializing robotic assistance technologies.}
\thanks{Zackory Erickson is with the Robotics Institute, Carnegie Mellon University, Pittsburgh, PA., USA. (e-mail: \href{mailto:zackory@cmu.edu}{zackory@cmu.edu})}%
\thanks{Henry M. Clever, Vamsee Gangaram, Eliot Xing, and Charles C. Kemp are with the Healthcare Robotics Lab, Georgia Institute of Technology, Atlanta, GA., USA. (e-mail: \href{mailto:henryclever@gatech.edu}{henryclever@gatech.edu}, \href{mailto:vamsee.gangaram@gmail.com}{vamsee.gangaram@gmail.com}, \href{mailto:exing@gatech.edu}{exing@gatech.edu},
\href{mailto:charlie.kemp@bme.gatech.edu}{charlie.kemp@bme.gatech.edu})}%
\thanks{Greg Turk is with the School of Interactive Computing, Georgia Institute of Technology, Atlanta, GA., USA. (e-mail: \href{mailto:turk@cc.gatech.edu}{turk@cc.gatech.edu})}%
\thanks{C. Karen Liu is with the Department of Computer Science, Stanford University, Stanford, CA., USA. (e-mail: \href{mailto:karenliu@cs.stanford.edu}{karenliu@cs.stanford.edu})}%
}

\begin{document}

\maketitle

\begin{abstract}
Towards the goal of robots performing robust and intelligent physical interactions with people, it is crucial that robots are able to accurately sense the human body, follow trajectories around the body, and track human motion. 
This study introduces a capacitive servoing control scheme that allows a robot to sense and navigate around human limbs during close physical interactions.
Capacitive servoing leverages temporal measurements from a multi-electrode capacitive sensor array mounted on a robot's end effector to estimate the relative position and orientation (pose) of a nearby human limb.
Capacitive servoing then uses these human pose estimates from a data-driven pose estimator within a feedback control loop in order to maneuver the robot's end effector around the surface of a human limb.
We provide a design overview of capacitive sensors for human-robot interaction and then investigate the performance and generalization of capacitive servoing through an experiment with 12 human participants.
The results indicate that multidimensional capacitive servoing enables a robot's end effector to move proximally or distally along human limbs while adapting to human pose. Using a cross-validation experiment, results further show that capacitive servoing generalizes well across people with different body size.

\end{abstract}

\begin{IEEEkeywords}
Capacitive servoing, 
physical human-robot interaction,
capacitive sensors,
pose estimation,
physically assistive robotics.
\end{IEEEkeywords}

\section{Introduction}
\label{sec:intro}



\IEEEPARstart{P}{hysical} human-robot interaction and robotic assistance presents an opportunity to benefit the lives of many people, including the millions of older adults and people with physical disabilities, who have difficulty performing activities of daily living (ADLs) on their own~\cite{okoro2018prevalence, mitzner2014identifying}. However, accurately sensing the human body and tracking motion remains challenging for assistive robots, especially under the presence of visual occlusions.

\begin{figure}
\centering
\includegraphics[width=0.48\textwidth, trim={4cm 1cm 4cm 0cm}, clip]{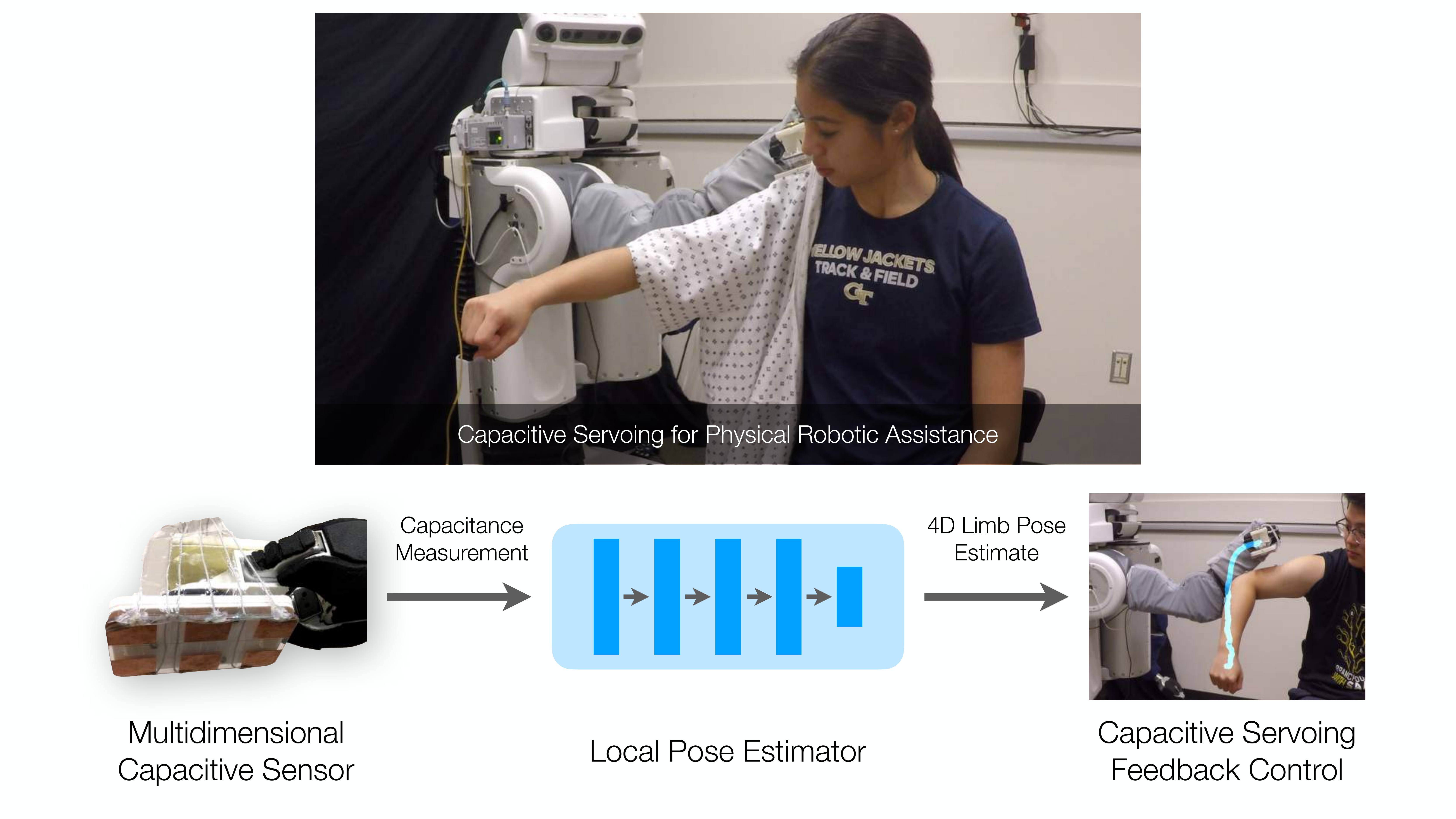}
\vspace{-0.2cm}
\caption{\label{fig:intro}(Top) A PR2 robot uses capacitive servoing to track human motion while pulling a hospital gown onto a participant's arm. (Bottom) The capacitive servoing process. We use a multidimensional capacitive sensor in order to observe capacitance measurements during physical human-robot interaction. We feed these measurements to a trained pose estimation model which provides a relative 4D pose estimate for a nearby human limb. We then use a feedback controller in order for a robot's end effector to follow the contours along a human limb and adapt to human motion in real time.}
\end{figure}

In this article, we introduce and characterize a multidimensional capacitive servoing technique that enables a robot to track the local pose of a person's limb in real time during physical human-robot interaction (Fig.~\ref{fig:intro}). Capacitive servoing consists of a feedback control loop that leverages measurements from an array of conductive capacitive electrodes mounted to a robot's end effector. Each electrode generates an electric field, which in turn allows a robot to sense nearby conductive materials, such as a human body. A key benefit of capacitive sensing is the ability for robots to sense the human body through some opaque materials, including garments and wet cloth. Such materials often visually occlude a robot's sight of a person during caregiving tasks, such as dressing~\cite{gao2015user} and bathing.

For a robot to perform servoing around human limbs, we train a local pose estimation model that takes a time series of capacitance measurements from an array of six capacitive electrodes and estimates a relative pose offset between a capacitive sensor mounted at the robot's end effector and a person's limb. This pose estimate includes the relative position, $\bm{p} = (p^{}_y, p^{}_z)$, as well as the pitch and yaw orientation, $\bm{\theta} = (\theta^{}_y, \theta^{}_z)$, between the capacitive sensor array and a nearby human limb.
Yaw rotation is defined with respect to the central axis of the limb, whereas the relative position and pitch orientation are defined with respect to a coordinate frame at a nearby point on the surface of the limb. Using this representation, a robot is able to traverse its end effector along the upper surface of human limbs.

We then use these estimates with feedback control, which allows a robot to move its end effector along trajectories that follow the contours of a person's limbs.
By conducting a study with 12 human participants, we investigate performance of this new servoing technique through several metrics, including task success rates, generalization, sensing ranges, and pose estimation error.
Our results demonstrate that capacitive servoing can enable a robot to maintain its end effector on average within 1~cm from a target goal near a human limb, while simultaneously traversing the limb. Across various human limb poses and limb sizes, contact between the robot's end effector and a human limb occurred for only one participant and lasted for under 0.2 seconds.

The authors' prior work in~\cite{erickson2019multidimensional} introduced a multidimensional capacitive sensor, which a robot used to provide dressing and bathing assistance to four able-bodied participants. 
Since this prior work was application-driven, it \emph{did not} investigate the general capabilities and limitations of capacitive sensing for physical human-robot interaction. 
The current article significantly builds on work in~\cite{erickson2019multidimensional} by characterizing a capacitive servoing control scheme for physical human-robot interaction and for controlling a robot's end effector to move along the contours of human limbs.
In comparison to prior work, in this article we make the following contributions:
\begin{itemize}
\item We provide a formalized control scheme for capacitive servoing along with an intuition and design overview of multi-electrode capacitive sensors for sensing the pose of human limbs along multiple axes.
\item We conducted a human-robot interaction study with 12 able-bodied participants where we investigated how capacitive servoing can enable a robot to control its end effector both proximally and distally along the contours of a human limb, which is the key component for performing several robotic caregiving tasks around the body.
\item We characterize the sensing ranges of an end effector-mounted capacitive sensor and the generalization of human limb pose estimation across people with different body size.
\end{itemize}

\section{Related Work}
\label{sec:related_work}

\subsection{Capacitive Sensing}

One of the most common applications for capacitive sensors are within consumer devices, such as touchscreens~\cite{lee1985multi, barrett2010projected}, virtual reality controllers~\cite{lee2019torc, higgins2018hand}, and wearables/clothing~\cite{cheng2010active, holleis2008evaluating}. Grosse-Puppendahl et al.~\cite{grosse2017finding} provide an overview of capacitive sensing within human-computer interaction, which includes a discussion of sensing modes, commonly used conductive materials, and capacitive sensing applications. Zimmerman et al.~\cite{zimmerman1995applying} also introduced several modes of operation and applications of capacitive sensing in human-computer interaction.

Within robotics, capacitive sensors have frequently been used as a form of proximity sensing~\cite{ye2020reviewproximity}. Navarro et al.~\cite{navarro2021proximity} provide a comprehensive overview of capacitive and proximity sensing in human-centered robotics, including applications, multimodal sensing techniques from literature that incorporate capacitance, and future challenges. Capacitive sensors have been mounted along the surface of robotic arms for avoiding collisions with people and stationary objects in the robot's environment~\cite{schlegl2013virtual, xia2016multi, hoffmann2016environment, tsuji2019proximity, ding2020collision, poeppel2020robust}. Non-contact capacitive sensing has also seen applications in controlling prostheses and exoskeletons~\cite{zheng2016noncontact, crea2019controlling}, tracking metallic objects near an industrial robot~\cite{xia2018tri}, teleoperation of mobile manipulators~\cite{stetco2020gesture}, and non-contact material recognition~\cite{kirchner2008capacitive, alagi2018material}. When affixed to the human body as wearable sensors, capacitive proximity sensors have also been used for limb motion recognition~\cite{zheng2018forearm}, activity recognition~\cite{cheng2013designing, chen2013locomotion}, and health monitoring~\cite{wang2017flexible}. In comparison to prior capacitive sensing methods, we characterize a multidimensional capacitive sensor array for physical human-robot interaction, which enables a mobile manipulator to estimate relative human limb pose and perform servoing along the human body.

Capacitive sensors have also been widely used for contact-based tactile sensing with use cases in texture recognition~\cite{muhammad2011capacitive} and for creating tactile artificial skins~\cite{cotton2009multifunctional, damilano2017robust}, which can serve as a robotic skin for collision monitoring~\cite{phan2011capacitive}. Capacitive sensors can be designed and integrated into artificial robotic skins to measure forces applied to the robot's body~\cite{ulmen2010robust, viry2014flexible, ji2016design}. Capacitive tactile sensors have also been designed for robot fingertips and hands~\cite{schmitz2011methods, maslyczyk2017highly, cataldi2018carbon, hashizume2019capacitive, gruebele2020stretchable}, which can provide high resolution force sensing during object grasping and dexterous manipulation. Capacitive sensing has been used in tandem with pneumatic sensing to estimate forces and deformation in a soft robotic finger~\cite{navarro2020model}. A number of works have introduced and developed capacitive tactile proximity sensors, which are capable of both sensing conductive objects from a distance away and measuring applied contact forces~\cite{lee2009dual, goeger2010tactile, navarro2013methods, navarro20146d, alagi2016versatile}. 

In order to navigate a robot's end effector along the surface of stationary objects, Navarro et al.~\cite{navarro20146d} introduced an arrangement of 2~$\times$~2 capacitive sensors on the inside of a robot's parallel jaw gripper. The authors presented a closed-loop controller (proximity servoing) which operated on pairwise differences in sensor values. Using this setup, the robot would continually realign its end effector to the shape and orientation of an object, while it closed its end effector to grasp a flat plate or cylindrical rod.
Navarro et al.~\cite{navarro20163d} have also fitted a cylindrical robot end effector with an array of capacitive sensors for following the contours of stationary metallic objects. 
The authors estimate curvature by fitting circles to linear position estimates from the co-located capacitive sensors and then compute an angle between the capacitive sensors and the object curvature for proportional control.
Similar to these two prior approaches, our work also uses capacitive sensing to define control trajectories for a robot's end effector. However, rather than sensing objects between a parallel jaw gripper or servoing around stationary metallic objects, we explore capacitive sensing for physical interaction with people, including following the contours along human limbs and tracking human limb motion. Additionally, we estimate the position and orientation offset between a robot's end effector and a human limb using a temporal data-driven approach on capacitance measurements.

\subsection{Servoing}

One of the most prominent servoing techniques is visual servoing~\cite{espiau1992new, hutchinson1996tutorial}, which uses computer vision techniques to extract information from visual data as feedback to control a robot. Two common approaches are \emph{image-based visual servoing} that uses features directly available in image data, and \emph{position-based visual servoing}, which estimates the 3D pose of a robot relative to a calibrated camera pose for computing a kinematic error used in control. For a more in-depth overview on visual servoing methods, we refer the reader to~\cite{chaumette2006visual, chaumette2007visual, janabi2010comparison, chaumette2016visual}. 

Towards robots that can interact with the human body, visual servoing has been applied in medical robotics contexts, such as tracking human body parts during surgery~\cite{azizian2014visual, azizian2015visual} and commanding robotic surgical tools~\cite{krupa2003autonomous, osa2010framework, dahroug2017visual}. Visual servoing has also been used for collision avoidance with a human collaborator wearing motion capture equipment~\cite{garcia2009visual}, to track human trajectories during human-robot collaboration~\cite{pomares2011direct}, and for human-robot object transfers~\cite{wang2015visual}. 

Beyond visual servoing, other modalities such as force or capacitive sensing have also been used in the control loop. Reichl et al.~\cite{reichl2013electromagnetic} present a servoing approach for tracking medical instruments within the human body by using electromagnetic sensors that do not require line of sight. Lepora et al.~\cite{lepora2017exploratory} use tactile servoing to navigate along edges of planar objects. Sutanto et al.~\cite{sutanto2019learning} learn a dynamics model from human demonstrations for tactile servoing with a BioTac tactile sensing finger. Ding et al.~\cite{ding2019proximity, ding2020collision} attach time-of-flight (ToF) and capacitive proximity sensor cuffs to a robot arm for active collision avoidance during robot motion while following a task trajectory. 

Wistort and Smith~\cite{wistort2008electric} introduced a capacitive servoing (electric field servoing) approach for object manipulation by implementing a transmitter and receiver electrode (E-field sensor) on each of the three fingertips of a Barrett Hand. The authors demonstrate capacitive servoing for preshaping the grasp of an object, and for aligning the robot's end effector above stationary objects, prior to grasping. In following work, Mayton et al.~\cite{mayton2010electric} integrated these E-field sensors onto a WAM arm and proposed strategies for grasping objects using closed-loop servo control and object handovers to people.
In comparison to other servoing methods, in this article we characterize a multidimensional capacitive servoing technique that enables a robot to navigate its end effector along the contours of human limbs, regardless of visual occlusions and without requiring contact with a person's body.



\subsection{Human Pose Estimation for Physical Assistance}

A number of approaches have been proposed for sensing human pose during physical human-robot interaction. One common pose estimation method during physical interaction is using visual features from an RGB or depth camera. For example, several works have used depth sensing for estimating human pose~\cite{yamazaki2014bottom, klee2015personalized, chance2018elbows, joshi2019framework} and tracking cloth features~\cite{koganti2017bayesian, gabas2016robot, zhang2020learning} while a robot helps a person or mannequin dress on a clothing garment. Jim{\'e}nez et al.~\cite{jimenez2020perception} provide an overview of various perception techniques for tracking cloth during assistive robotic tasks. Zlatintsi et al.~\cite{zlatintsi2018multimodal, zlatintsi2020isupport} present a bathroom setup with three Microsoft Kinects to capture RGB and depth information for human gesture and pose recognition in a robot-assisted bathing context. However, the human body can often become visually occluded during physical human-robot interaction tasks, where the robot's own arms may intersect line of sight of a person, or clothing can occlude the body during dressing assistance.

In healthcare and assistive robotics settings, sensing approaches that rely on force sensors~\cite{gao2016iterative}, pressure mats~\cite{casas2018human, clever20183d, clever2020bodies}, capacitive sensors~\cite{erickson2018tracking, kapusta2019personalized, clegg2020learning}, and feature extraction from occluded depth images~\cite{achilles2016patient, chance2018elbows} have been applied to recover visually occluded human pose. 

As an alternative to visual features, haptic measurements have been used to detect errors in robot-assisted dressing tasks when cloth gets caught on the human body~\cite{chance2016assistive, chance2017quantitative}. Haptic and kinematic measurements from a robot's end effector have been used during robot-assisted dressing to estimate the forces a fabric garment applied onto a person's stationary body~\cite{erickson2017does, erickson2018deep}. Several works have also combined force measurements with data from a depth camera to optimize personalized robotic dressing trajectories for a sleeveless jacket~\cite{gao2016iterative, zhang2017personalized, zhang2019probabilistic}. In contrast, capacitive sensing has the advantage of tracking human pose without applying force to a person's body and can directly sense the human body through many visually opaque materials, including cloth.

\section{Sensor, Model, and Controller}
\label{sec:method}

In this section, we describe the capacitive servoing technique, including the multidimensional capacitive sensor and data-driven human pose estimation models. Along with this, we provide the process for collecting capacitance data for pose estimation, and the capacitive servoing feedback control loop for controlling a robot's end effector to move along the contours of a human limb.

\subsection{Multidimensional Capacitive Sensor Design}

\begin{figure}
\centering
\includegraphics[width=0.23\textwidth, trim={5cm 4cm 4cm 4cm}, clip]{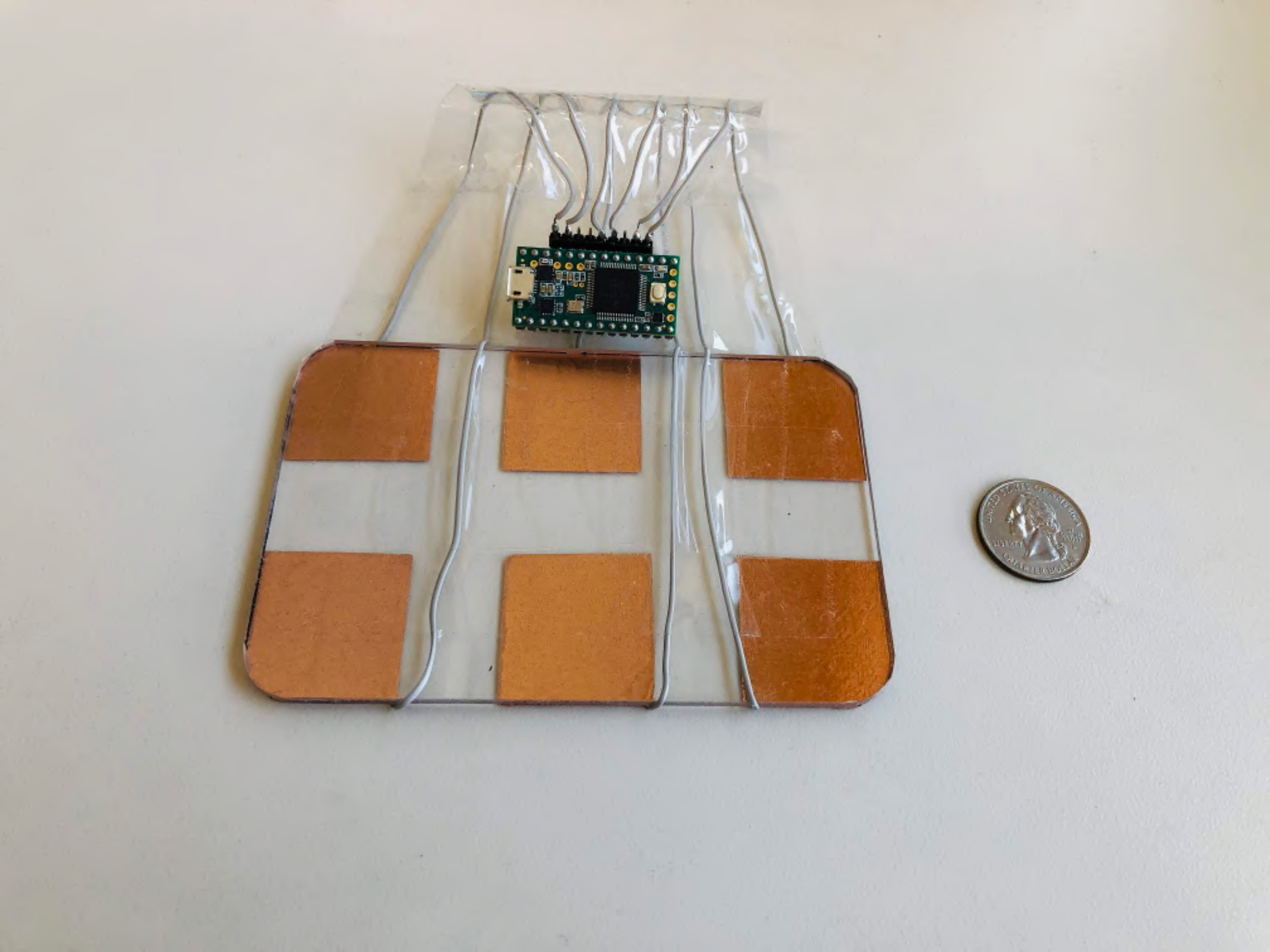}
\includegraphics[width=0.23\textwidth, trim={8cm 4.5cm 12cm 0cm}, clip]{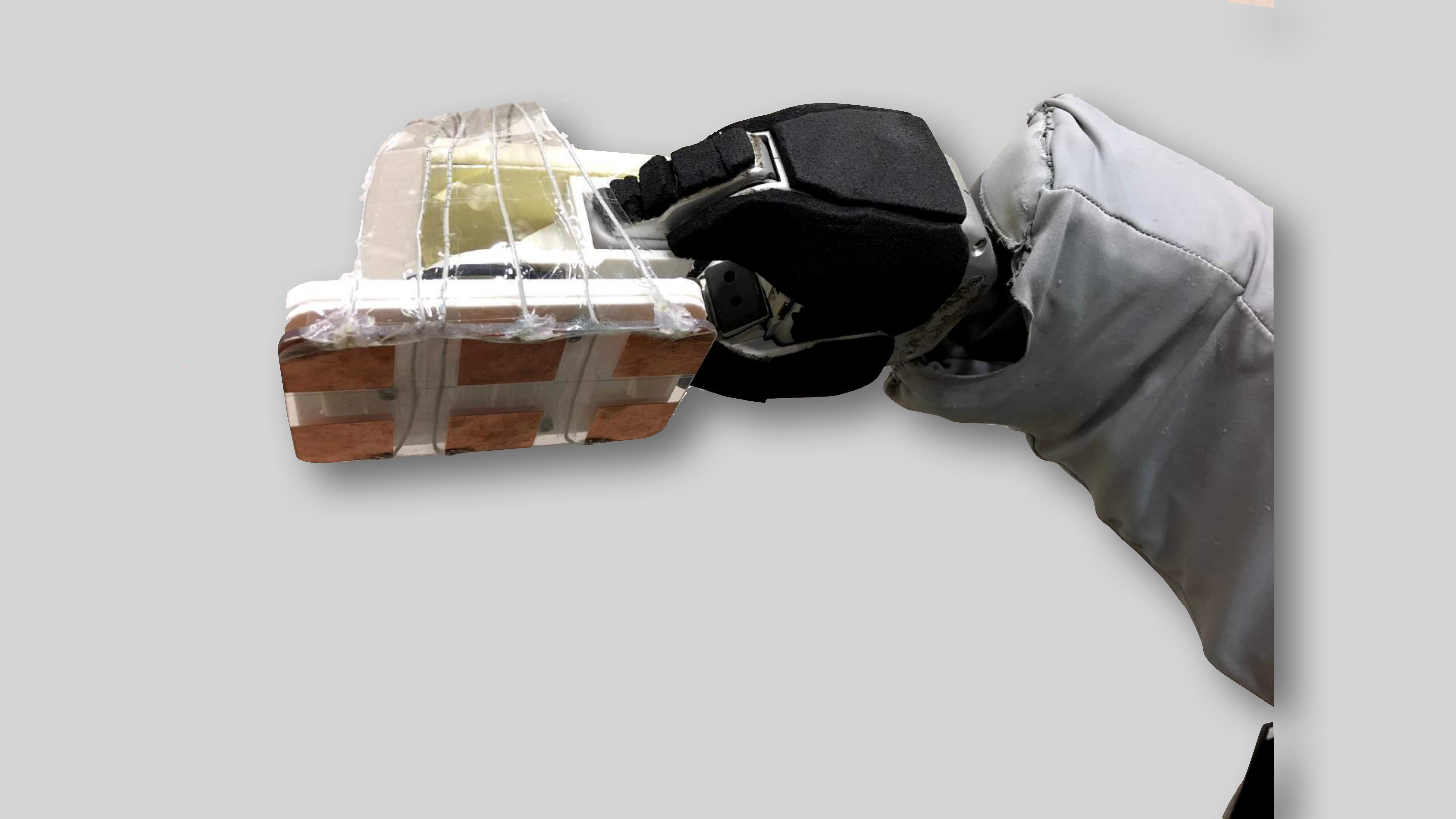}
\caption{\label{fig:sensor}(Left) Six-electrode capacitive sensor connected to a Teensy microcontroller board, with a U.S. quarter shown for sizing. (Right) Capacitive sensor mounted to the bottom of an assistive tool, which allows a robot to more easily hold task relevant objects (garments and washcloths).}
\end{figure}

\begin{figure*}
\centering
\subfloat[\label{subfig:aa}]{\includegraphics[width=0.23\textwidth, trim={7cm 6cm 9cm 2cm}, clip]{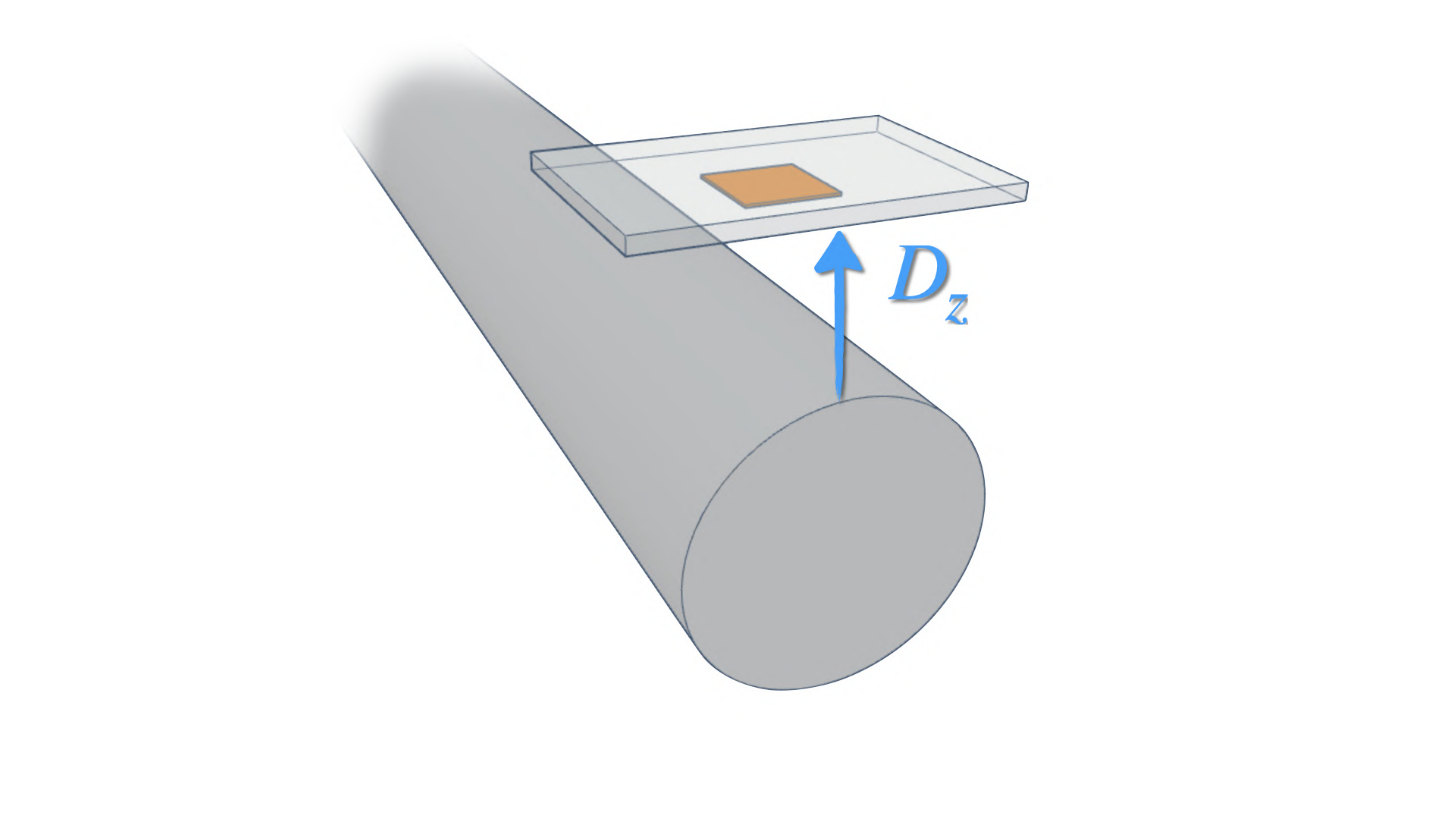}}\ 
\subfloat[\label{subfig:bb}]{\includegraphics[width=0.23\textwidth, trim={11cm 5cm 5cm 3cm}, clip]{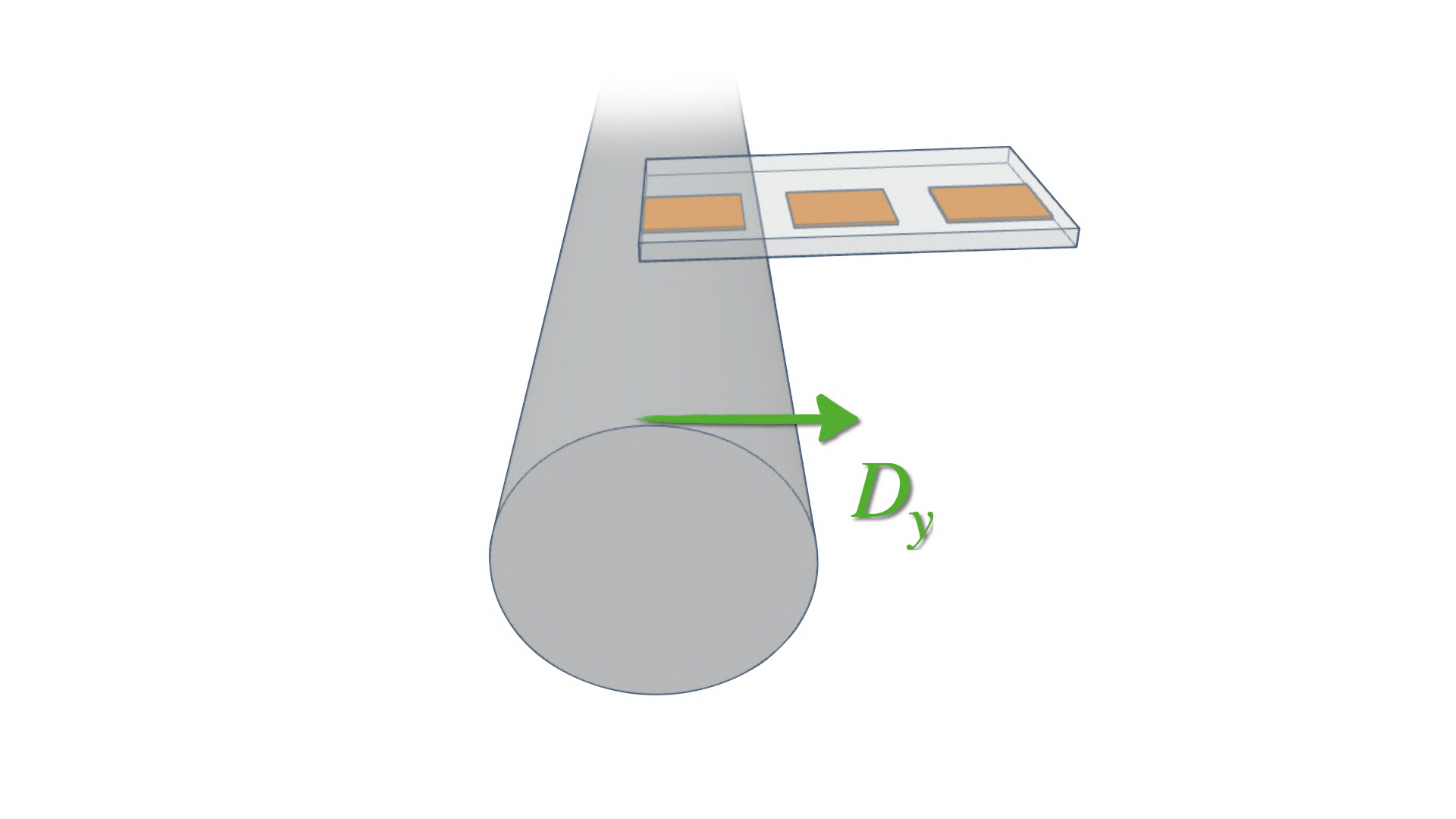}}\ 
\subfloat[\label{subfig:cc}]{\includegraphics[width=0.23\textwidth, trim={17cm 15cm 12cm 1cm}, clip]{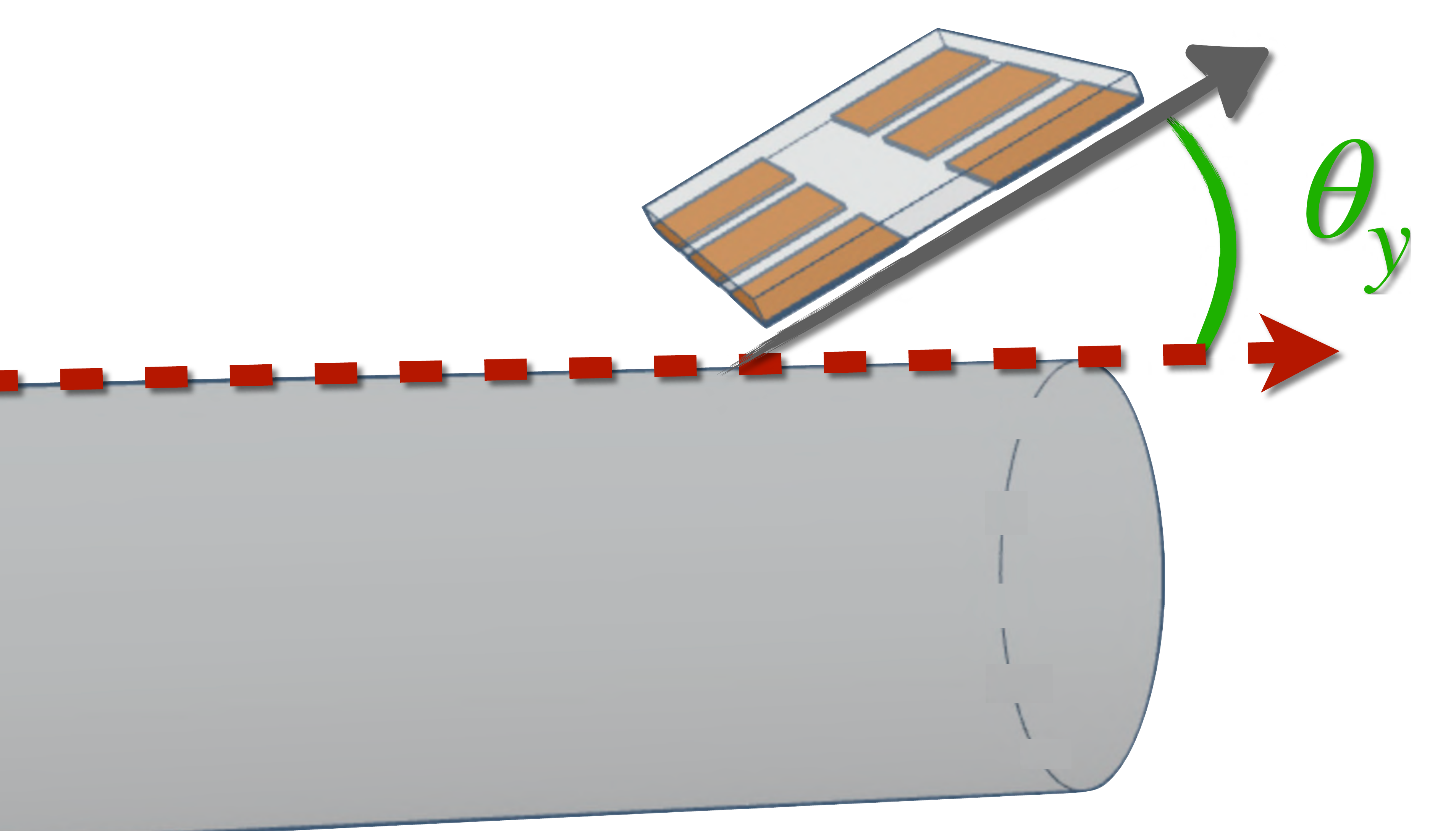}}\ 
\subfloat[\label{subfig:dd}]{\includegraphics[width=0.23\textwidth, trim={4cm 1cm 4cm 1cm}, clip]{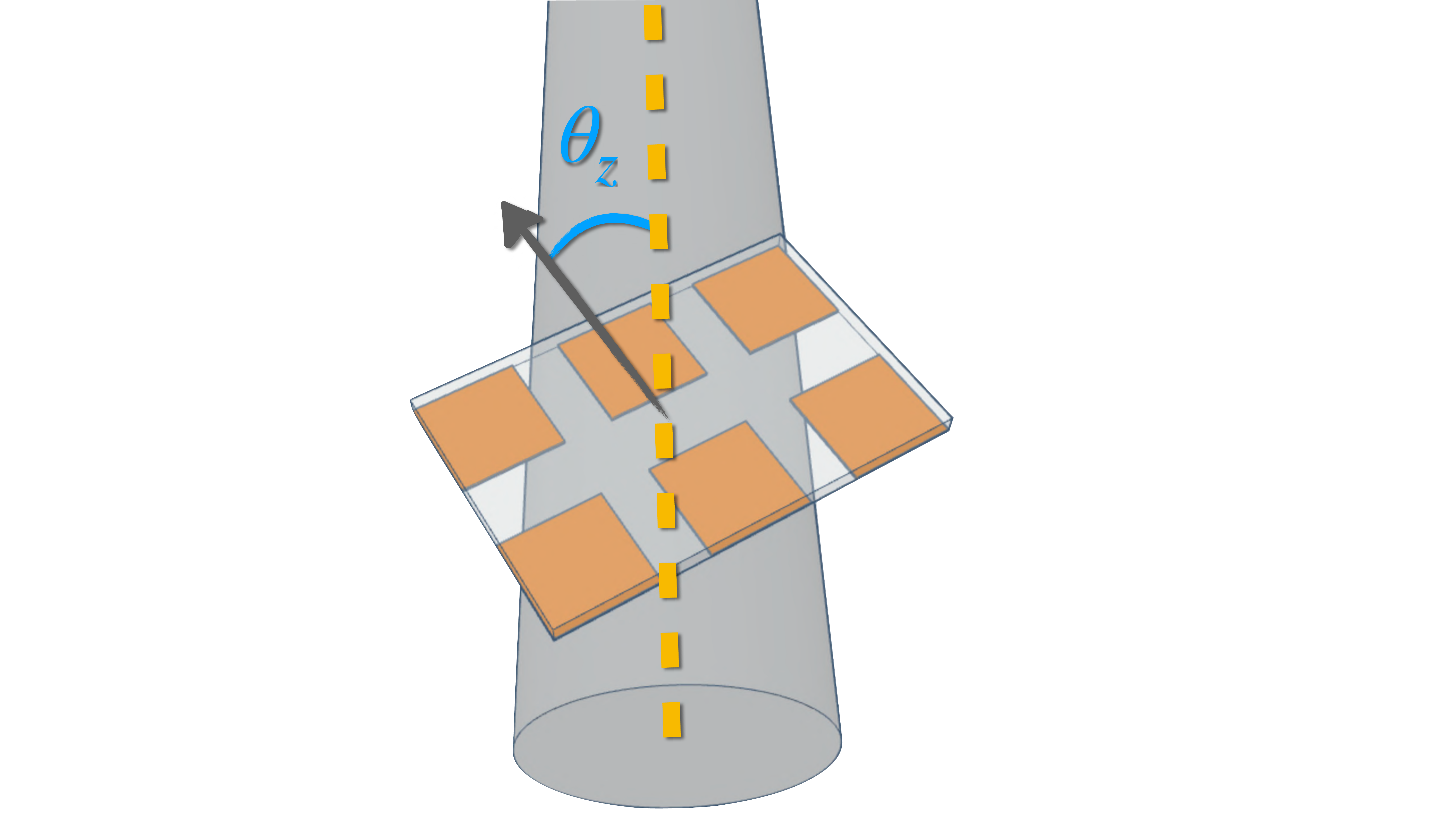}}\ 
\caption{\label{fig:abstract_cap_axes}(a) Sensing vertical displacement $D^{}_z$ between the capacitive sensor and surface of a cylinder is feasible with only a single electrode. (b) A row of three capacitive electrode enables sensing both lateral displacement $D^{}_y$ and vertical displacement. (c), (d) Adding a second row of electrodes allows for sensing pitch rotation $\theta_y$ to the surface of the cylinder and yaw rotation $\theta^{}_z$ with respect to the central axis.}
\end{figure*}


Our capacitive sensor design consists of six positively charged capacitive electrodes, arranged in a 3-by-2 grid and adhered to a polycarbonate plate (Fig.~\ref{fig:sensor}). The entire sensor with the polycarbonate mounting plate is a 11.5 cm $\times$ 8.5 cm $\times$ 1 mm rectangular surface with rounded corners. Each electrode is a 3~cm $\times$ 3~cm square of copper foil, which we connected to a commercially available Teensy 3.2 microcontroller board. The Teensy has dedicated hardware components for capacitive proximity sensing, including an analog voltage comparator and a 1~pF internal reference capacitor. 
With this setup, we can sample capacitance measurements at frequencies over 100~Hz from all six electrodes simultaneously. We then attached this capacitive sensor to the bottom of an assistive tool held by a robot's end effector, shown in~Fig.~\ref{fig:sensor}. This assistive tool enables a robot to more easily hold task relevant objects, such as fabric garments or washcloths.


\subsection{Multidimensional Capacitive Sensing Intuition}
\label{sec:multicap}

\begin{figure}
\centering
\includegraphics[width=0.48\textwidth, trim={0cm 7cm 0cm 4cm}, clip]{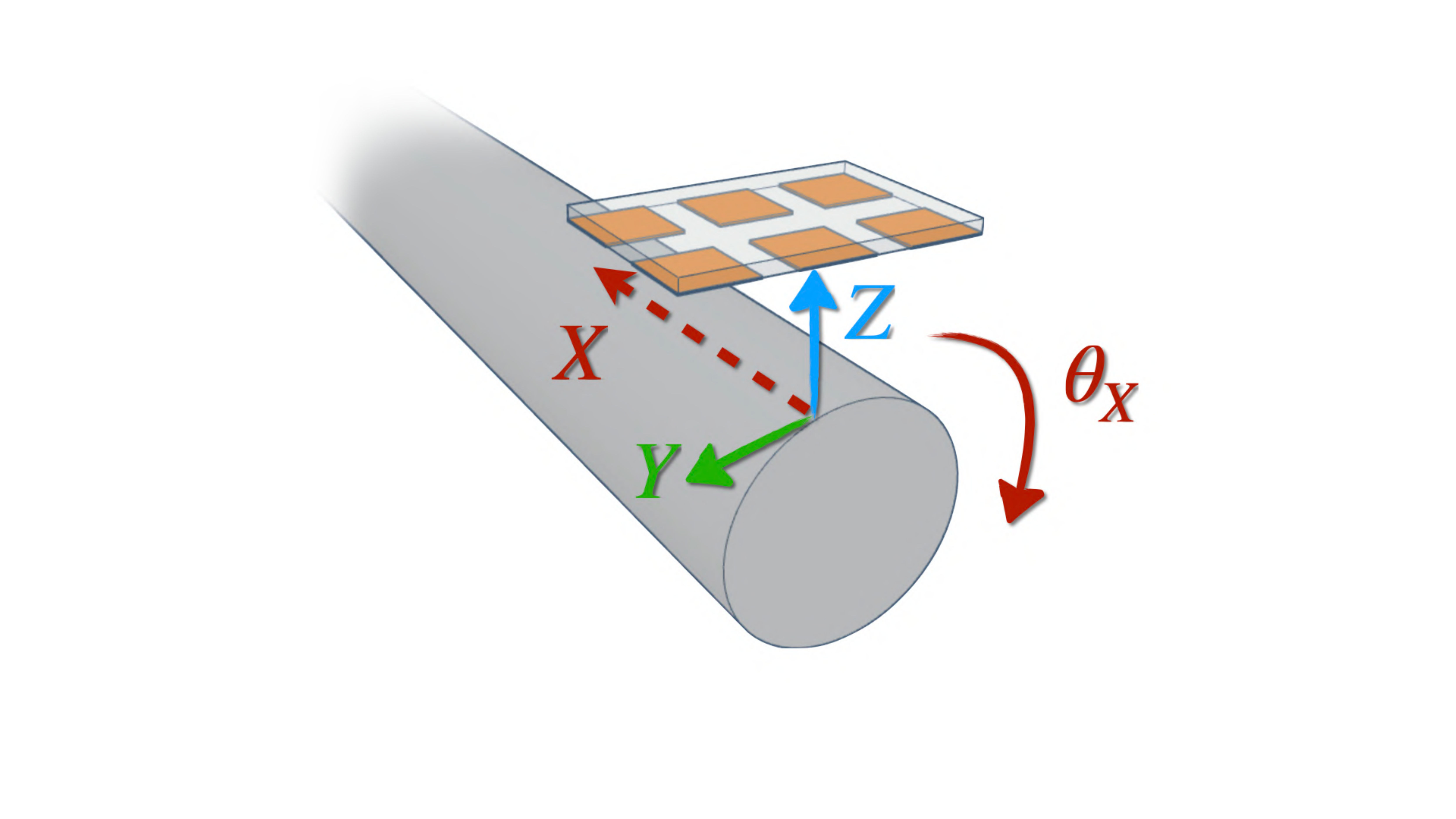}
\caption{\label{fig:abstract_cap}A multidimensional capacitive sensor above an infinitely long cylinder, which is an abstract representation of a human limb. We sense the relative translation and orientation between the capacitive sensor array and the origin of a coordinate frame at a point on the surface of the cylinder (limb). Note that sensing translation and rotation with respect to the $X$-axis along a cylinder ($X$ and $\theta_X$ shown in red) is infeasible.}
\end{figure}

Capacitive sensing is frequently used as a means to sense a nearby human body for human-computer interaction~\cite{grosse2017finding} and some human-robot interaction contexts~\cite{navarro2013methods, hoffmann2016environment}. A simple \emph{self-capacitance} capacitive sensor consists of a conductive electrode connected to a power source, which in turn generates an electric field. An external conductive object, such as a human limb, that intersects this electric field causes an increase in capacitance that can be measured from the electrode. Capacitance, or the change in electric charge stored on the electrode (given a fixed electric potential), is then inversely proportional to the distance between the electrode and the human limb. 

To gain an intuition for this multidimensional capacitive sensor design for sensing human limb pose, we can begin by representing a human limb as an abstract cylinder, as shown in Fig.~\ref{fig:abstract_cap_axes} and Fig.~\ref{fig:abstract_cap}. As demonstrated in~\cite{erickson2018tracking, kapusta2019personalized}, only a single electrode is needed to determine the vertical distance between the sensor and the surface of a human limb, along the Z-axis. This scenario is depicted in Fig.~\ref{fig:abstract_cap_axes} (a) with a single capacitive electrode vertically above the abstract cylinder. By adding two additional electrodes in parallel, the sensor can also measure lateral displacement along the Y-axis, as displayed in Fig.~\ref{fig:abstract_cap_axes}~(b). Finally, by adding a second row of three electrodes, this six-electrode capacitive sensor can measure pitch rotation to the surface of the limb ($\theta_y$, Fig.~\ref{fig:abstract_cap_axes} (c)) and yaw rotation with respect to the central axis of the limb ($\theta_z$, Fig.~\ref{fig:abstract_cap_axes} (d)).

\begin{figure}
\centering
\includegraphics[width=0.48\textwidth, trim={10cm 8cm 6cm 5cm}, clip]{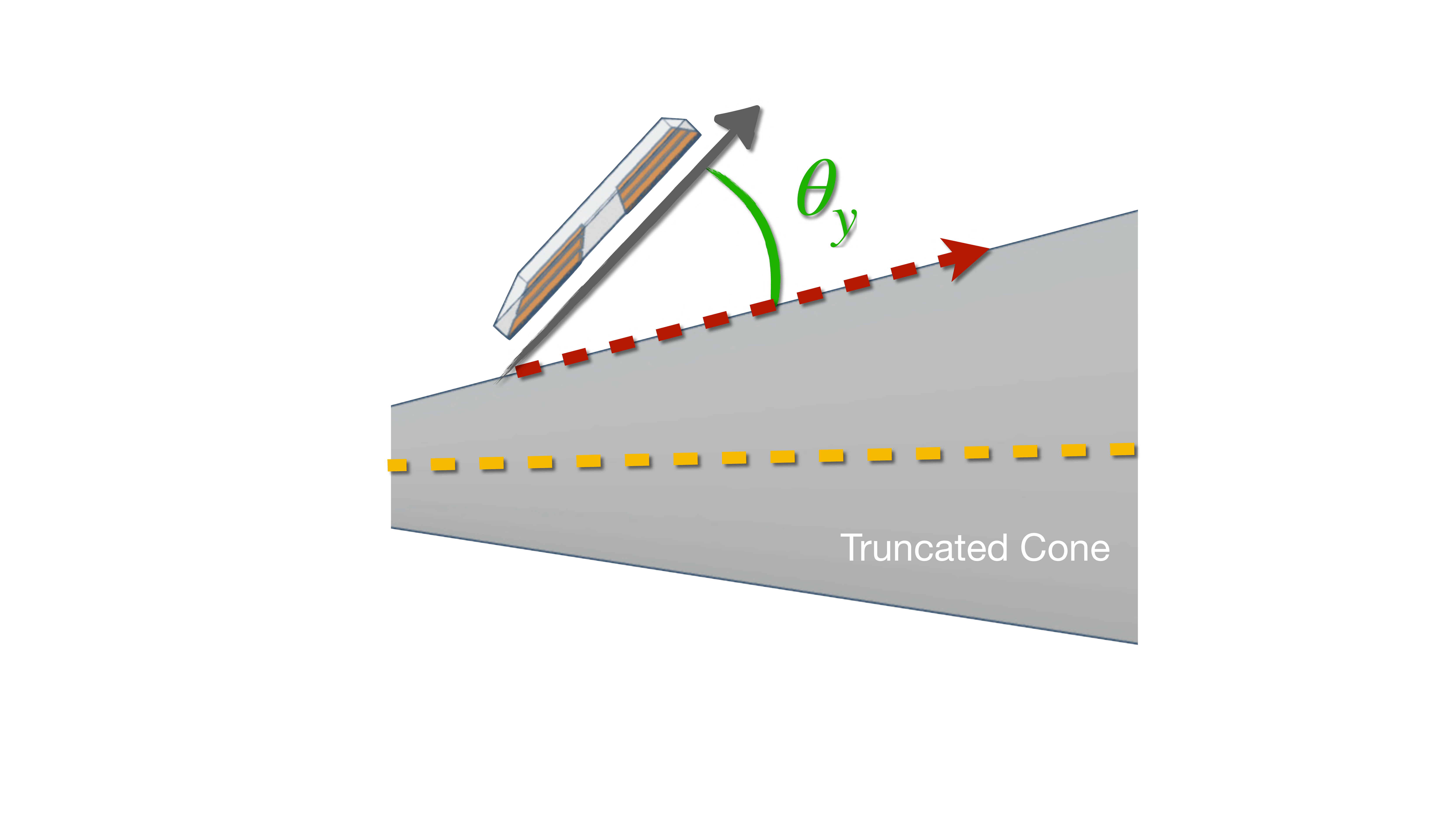}
\vspace{-0.3cm}
\caption{\label{fig:abstract_cap_truncated_cone}$\theta^{}_y$ is represented geometrically as the orientation between the sensor and the local limb surface (red dashed line) rather than to the limb's central axis (yellow dashed line). This representation allows capacitive servoing to generalize to non-cylindrical limbs which can be more accurately modeled as a truncated cone (conical frustum).}
\end{figure}

Overall, we can use this 3 $\times$ 2 grid of electrodes to sense the 4-dimensional relative pose of a human limb (or conductive cylindrical object) with respect to the sensor. This pose includes the 2-dimensional translation and 2-dimensional rotation between the capacitive sensor array and a nearby limb. The 2-dimensional translation ($D^{}_y$, $D^{}_z$) and pitch rotation ($\theta_y$) are defined with respect to a coordinate frame at a point on the surface of the limb. We define this point as the highest vertical point on a cross section of the limb---a cross section that is perpendicular to the limb's central axis and intersects the center of the capacitive sensor (see Fig.~\ref{fig:abstract_cap}). Yaw rotation ($\theta_z$) is then defined with respect to the central axis of the limb. Note that we do not sense translation or rotation (roll, $\theta_x$) around the X-axis of the limb due to a cylinder's self-similarity along its central axis, as depicted in Fig.~\ref{fig:abstract_cap}. Fig.~\ref{fig:abstract_cap_truncated_cone} highlights the importance of geometrically modeling pitch rotation to the surface of a limb rather than the central axis. Doing so allows capacitive servoing to generalize to non-cylindrical limbs that are instead more accurately modeled as a truncated cone (conical frustum).

\begin{figure}
\centering
\includegraphics[width=0.48\textwidth, trim={0cm 0cm 0cm 7cm}, clip]{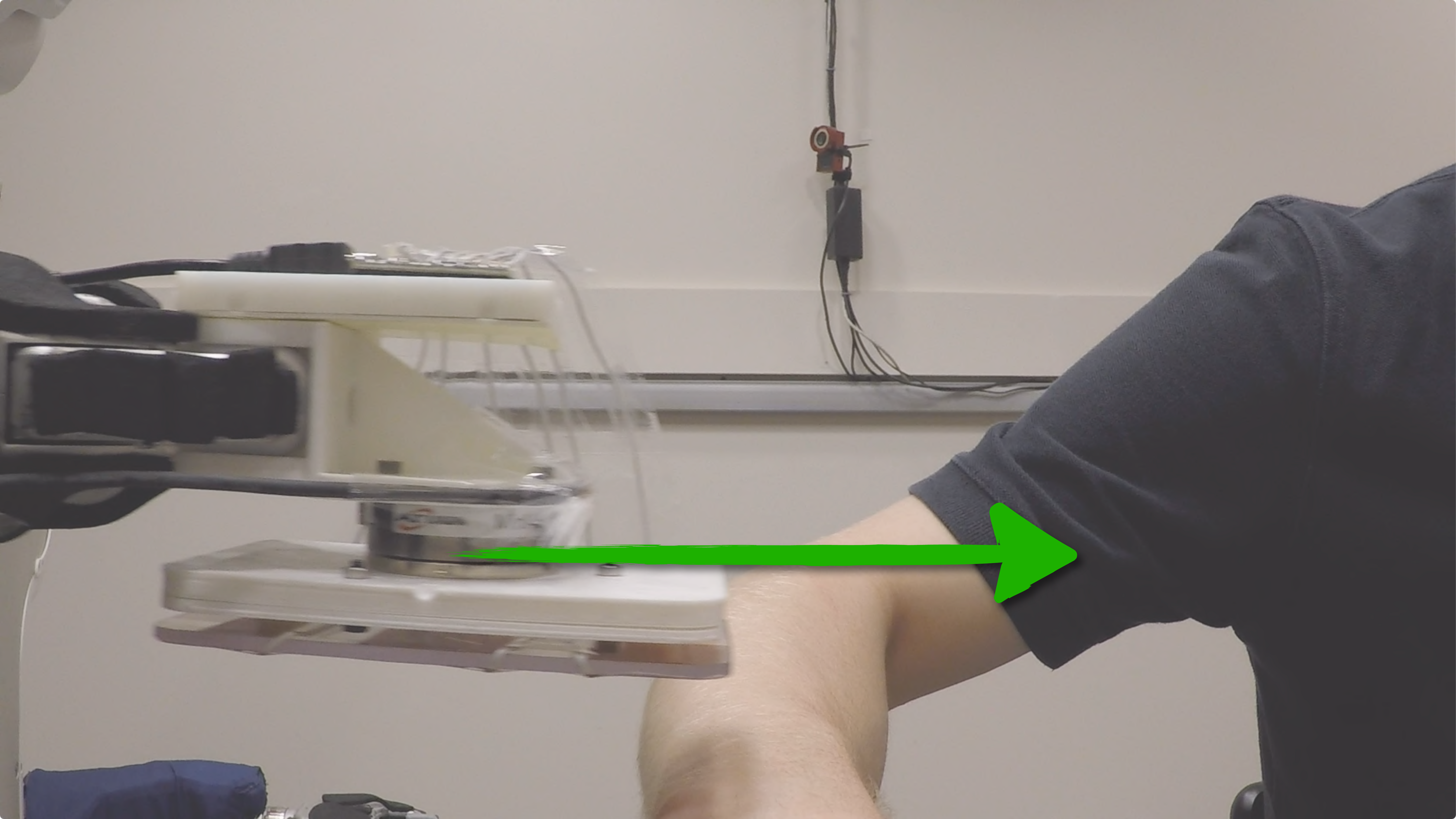}\\
\vspace{0.2cm}
\includegraphics[width=0.48\textwidth, trim={1cm 18.5cm 1cm 2.5cm}, clip]{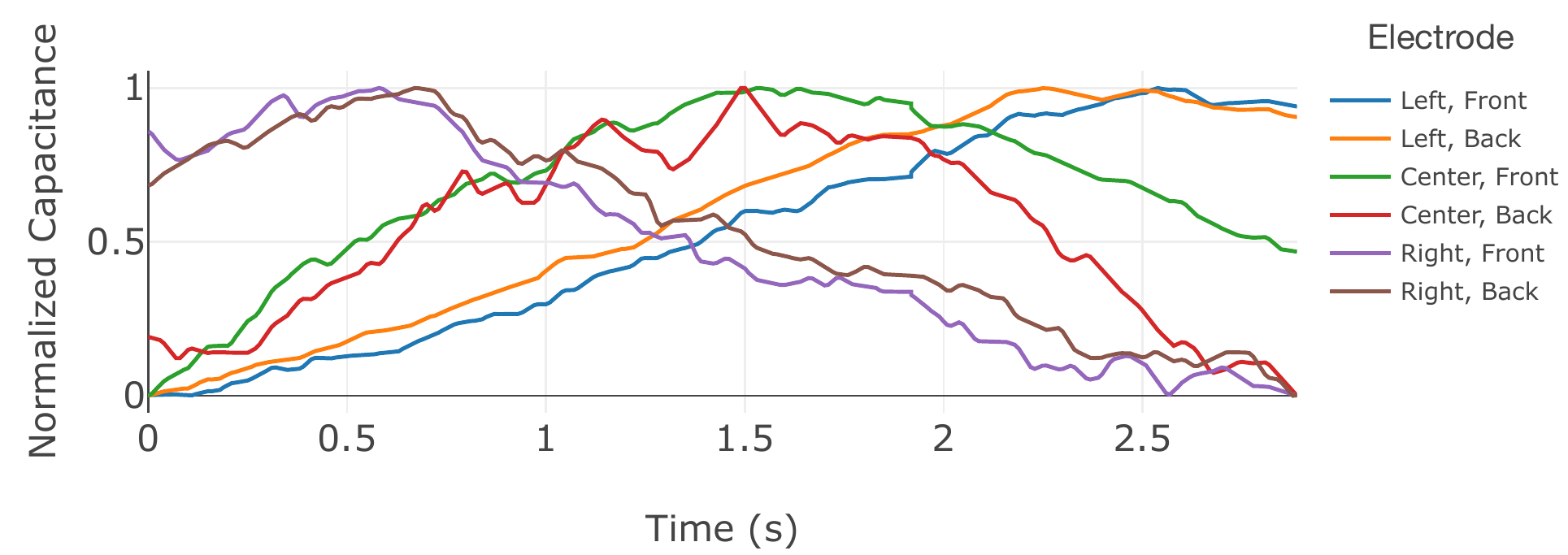}
\vspace{-0.3cm}
\caption{\label{fig:capacitance_plot}Capacitance measurements from all six capacitive electrodes as the robot moves the sensor from one side of a person's arm to the other. The sensor remains $\sim$5~cm vertically above a person's arm. We apply a finite impulse response (FIR) filter with numerator and denominator coefficients of 1 and 0.02, respectively. We normalize all signals to the range of [0, 1].}
\end{figure}

Fig.~\ref{fig:capacitance_plot} depicts an example of how capacitance measurements from this six-electrode sensor vary over time as a PR2 robot moves the sensor over a human arm. The sensor remains 5~cm vertically above the arm as it translates laterally across the arm. Since the sensor begins on the left side of the arm, the right column of capacitive electrodes initially measure a high capacitance. Once the sensor is centered above the arm (around the 1.5 second mark), the center row of electrodes record a high capacitance, while the four corner electrodes record lower capacitance readings. This trend continues as the sensor translates to the right side of the arm at which point the left column of electrodes measure the greatest capacitance.

\subsection{Estimating Relative Pose of a Human Limb}
\label{sec:learning}

Given measurements from this robot-mounted six-electrode capacitive sensor, we then consider how to estimate the geometric transpose between this capacitive sensor and a nearby human limb.

Using a single electrode capacitive sensor, \cite{erickson2018tracking} and \cite{kapusta2019personalized} demonstrated how non-linear least squares optimization can be used to fit a model for estimating the one-dimensional vertical distance between a robot's end effector and a human arm. Inspired by the classical capacitance equation for a parallel plate capacitor, this prior one-dimensional model, $d(C) = \frac{\alpha}{C + \beta}$, estimates the shortest distance between a capacitive sensor and a human limb given a single capacitance measurement, $C$, and optimized constants, $\{\alpha, \beta\}$. Yet, this straightforward model faces several challenges with respect to estimating both position and orientation of a human limb. The model is designed around a single capacitance measurement and lacks an evident generalization to multiple sensory inputs and pose outputs. In addition, this one-dimensional approach does not model variable electromagnetic interference (EMI) or crosstalk between adjacent electrodes, which are present in multi-electrode sensor configurations. 

To overcome these challenges, we consider a model $f: \mathbb{R}^{6h} \rightarrow \mathbb{R}^4$, that maps a temporal window of prior capacitance measurements over $h$ previous time steps from all six electrodes to a 4-dimensional relative pose estimate of a human limb. This model, $f(\bm{c}_{t-h+1:t})$, outputs an estimated relative pose $\bm{\hat{p}}^{}_t$ of a nearby human limb given a window of prior capacitance measurements, $\bm{c}_{t-h+1:t}$, from time step $t-h+1$ to the current time step $t$. Each capacitance measurement $\bm{c}_t=(c_1, \ldots, c_6)$ is a vector of measurements from all six capacitive electrodes. 

\begin{figure*}
\centering
\subfloat[\label{subfig:a}]{\includegraphics[width=0.23\textwidth, trim={24cm 7cm 14cm 13cm}, clip]{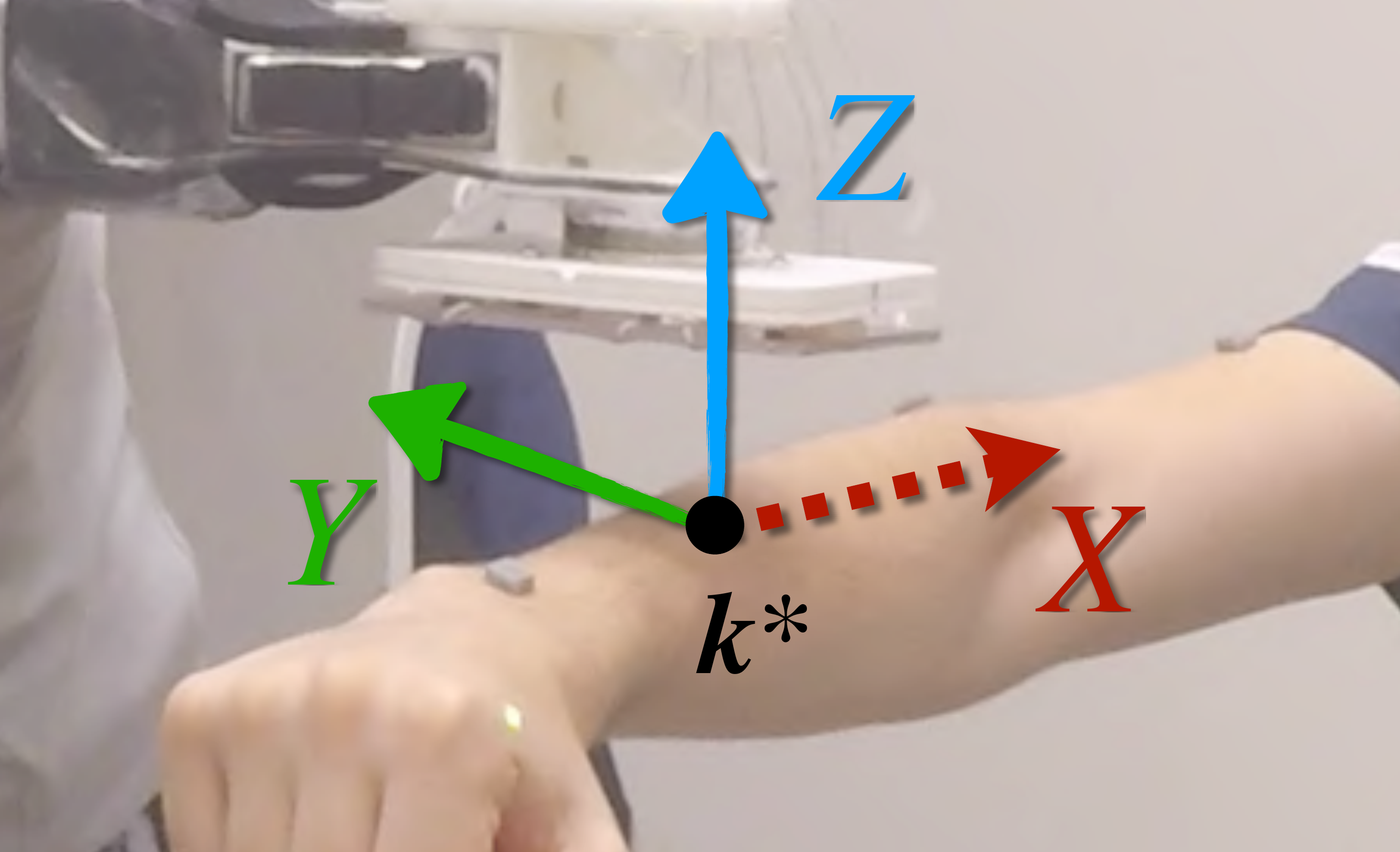}}\ 
\subfloat[\label{subfig:b}]{\includegraphics[width=0.23\textwidth, trim={12cm 8cm 12cm 3.5cm}, clip]{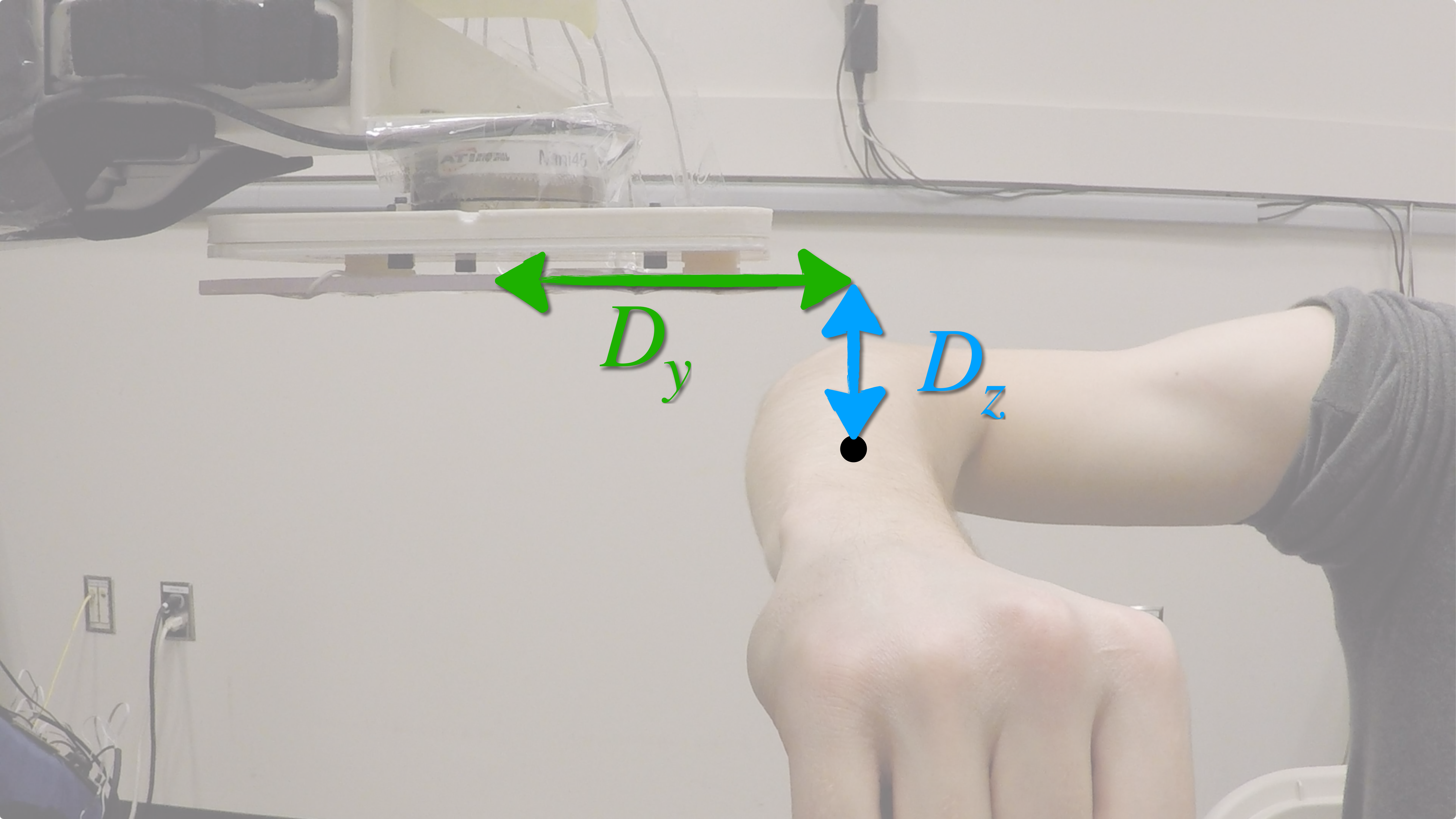}}\ 
\subfloat[\label{subfig:c}]{\includegraphics[width=0.23\textwidth, trim={8cm 3cm 6cm 2.5cm}, clip]{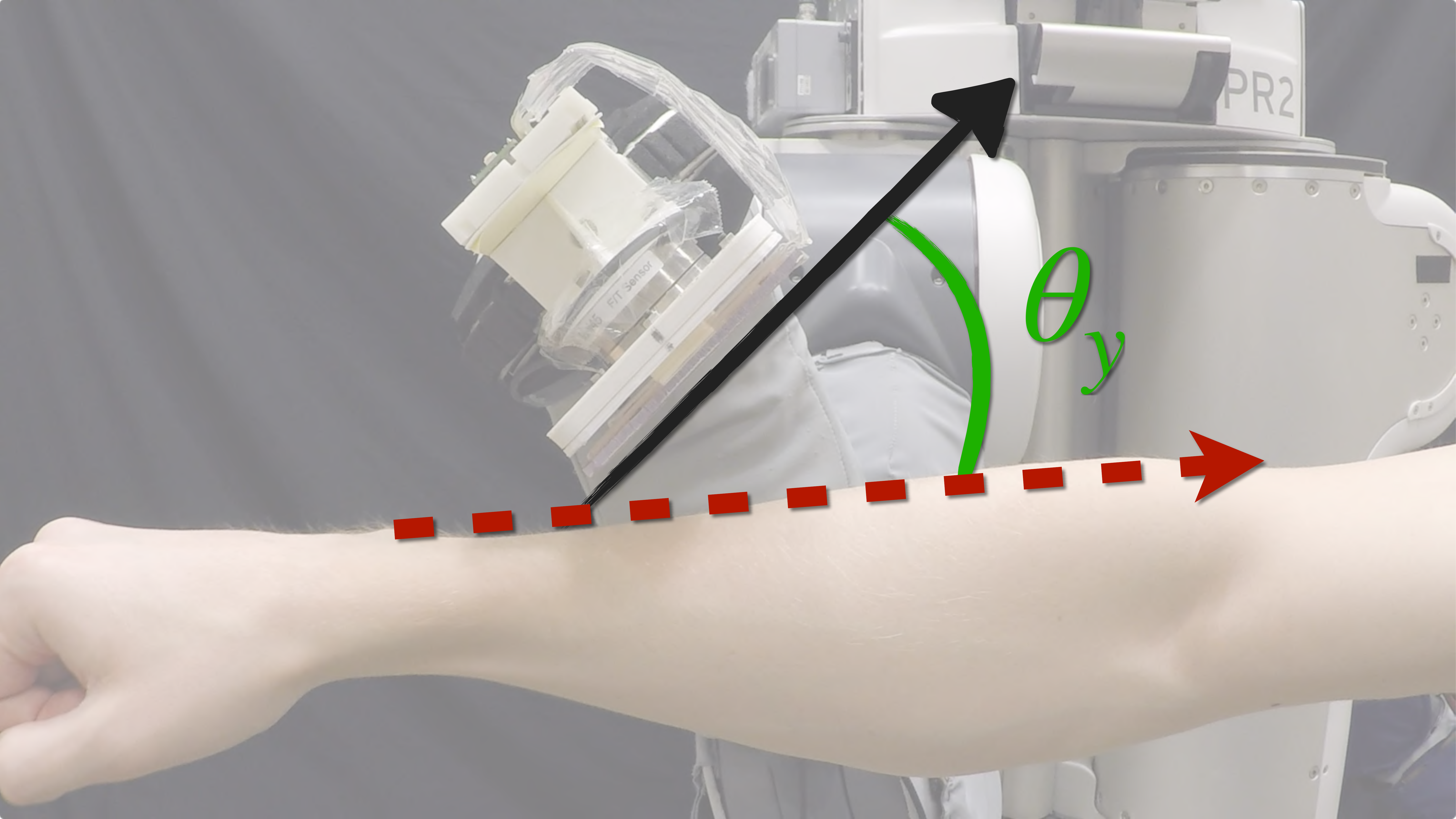}}\ 
\subfloat[\label{subfig:d}]{\includegraphics[width=0.23\textwidth, trim={4cm 1.5cm 20cm 10cm}, clip]{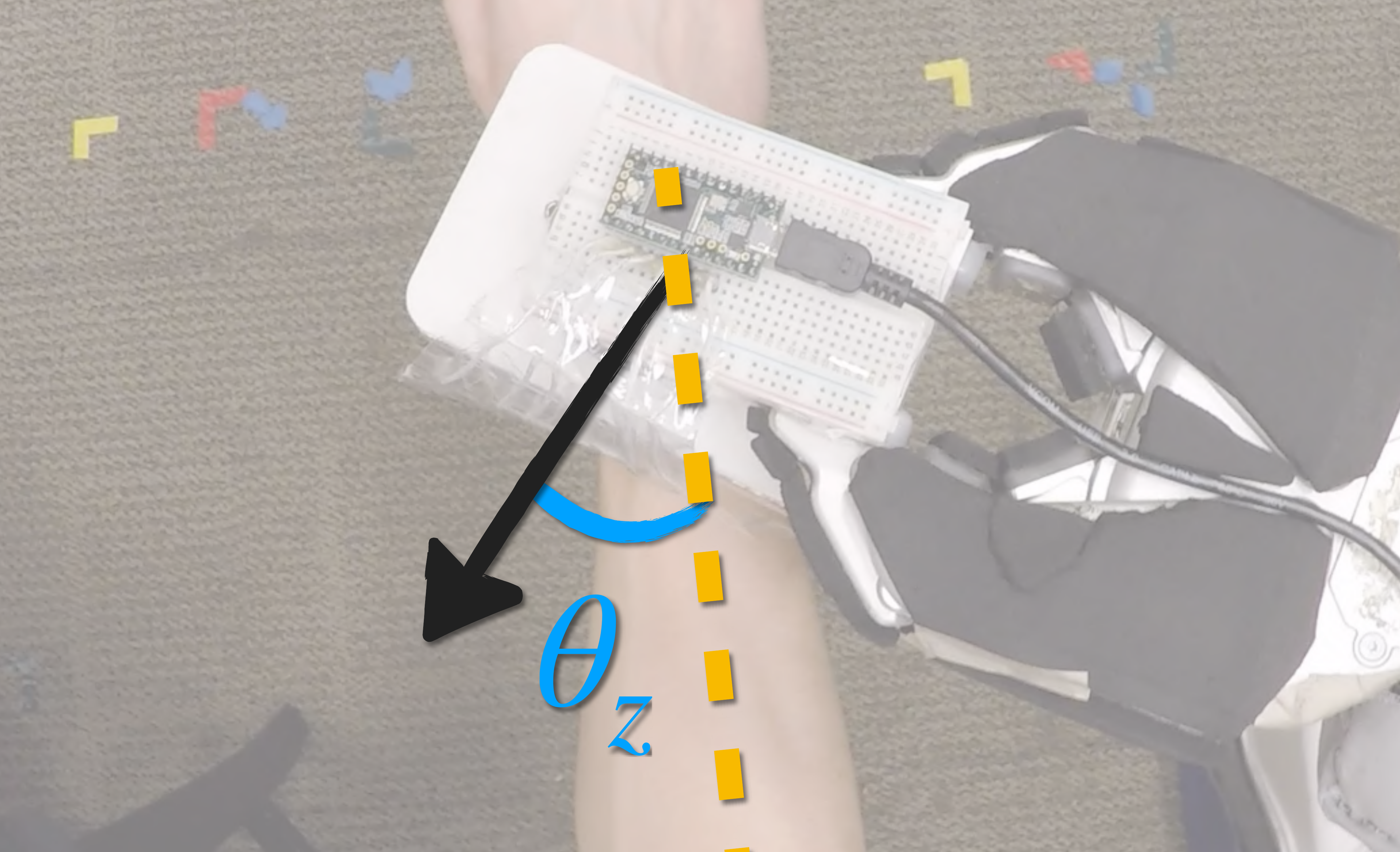}}\ 
\caption{\label{fig:axes}(a) The local coordinate frame at the top most vertical point $\bm{k}^*$ along the nearest perpendicular cross section of a human limb, as defined in (\ref{eq:k}). (b) The lateral $D^{}_y$ and vertical $D^{}_z$ displacement between the center of the capacitive sensor array and the point on the limb. (c) Pitch rotation $\theta^{}_y$ between the capacitive sensing plane and the local limb surface. (d) Relative yaw rotation $\theta^{}_z$ difference between the capacitive sensing array and the central axis of a human limb.} 
\end{figure*}

The pose estimate $\bm{\hat{p}}^{}_t = (\hat{D}^{}_{t,y}, \hat{D}^{}_{t,z}, \hat{\theta}^{}_{t,y}, \hat{\theta}^{}_{t,z})$ includes the relative position $\bm{D}^{}_t = (D^{}_{t,y}, D^{}_{t,z})$ and orientation of the limb $\bm{\theta}^{}_t = (\theta^{}_{t,y}, \theta^{}_{t,z})$, as shown in Fig.~\ref{fig:axes}. 
The relative 2-dimensional translations are represented in relation to a point on the surface of a human limb.
Consider a vector $\bm{n}$ along the central axis of a human limb that is near the capacitive sensor array and consider a point along the central axis $\bm{k}_0$ which is closest to the center of the capacitive sensor array.
We define a set of all points $\bm{k}$ such that $\bm{n} \cdot (\bm{k} - \bm{k}_0) = 0$ forms a plane perpendicular to the central axis and a cross section of the limb.
2-dimensional translations $\bm{D}^{}_t$ between the capacitive sensor and human limb are then defined with respect to the top most point in this plane along the surface of the limb,
\begin{align}
    &\bm{k}^* = \argmax_{\bm{k} \in \mathcal{H}}\;\; \bm{k} \cdot \bm{e}_z \nonumber \\
    &\mathrm{subject\;to\;\;\;} \bm{n} \cdot (\bm{k} - \bm{k}_0) = 0,
    \label{eq:k}
\end{align}
where $\bm{e}_z = (0, 0, 1)^T$ is a constant vector and $\mathcal{H}$ is the surface of the human limb.

Fig.~\ref{fig:axes} (a) visually depicts this point $\bm{k}^*$ along a human arm near the capacitive sensor array. Pitch orientation $\theta^{}_{t,y}$ is defined according to the Y-axis of a coordinate frame at the point $\bm{k}^*$ on this surface of the limb, as shown in Fig.~\ref{fig:axes} (a) and (c), whereas yaw orientation $\theta^{}_{t,z}$ is defined with respect to the central axis of the limb, depicted in Fig.~\ref{fig:axes} (d).

As discussed in Section~\ref{sec:multicap}, this sensing approach operates with respect to the Y- and Z-axes of a human limb, as it is difficult to sense relative position and orientation changes around the X-axis along the surface of the limb, due to the approximate cylindrical shape of a limb (Fig.~\ref{fig:axes} (a)).

\begin{figure}
\centering
\includegraphics[width=0.48\textwidth, trim={0.1cm 0cm 0.1cm 0cm}, clip]{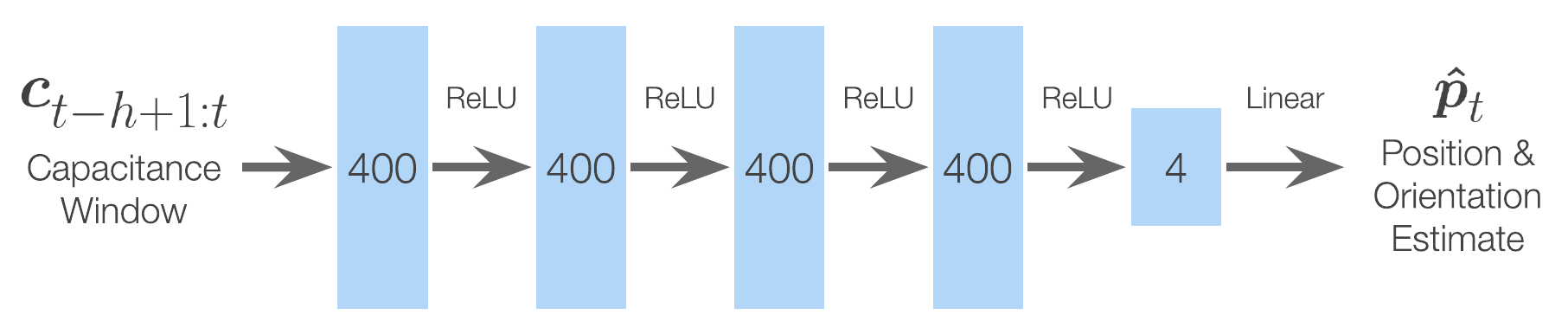}
\vspace{-0.4cm}
\caption{\label{fig:nn}The human pose estimation model used in capacitive servoing, implemented as a fully-connected neural network. This model estimates the current relative human limb pose $\bm{\hat{p}}^{}_t$ given a window of measurements $\bm{c}^{}_{t-h+1:t}$ from a capacitive sensor.}
\end{figure}

To estimate the relative pose of a human limb near the capacitive sensor, we leveraged a data-driven implementation of $f(\bm{c}_{t-h+1:t})$ using a fully-connected neural network architecture, as depicted in Fig.~\ref{fig:nn}. The network has four 400 node layers with a ReLU activation after each layer and a final output layer of four nodes with a linear activation. This final layer outputs an estimate, $\bm{\hat{p}_t}$, of the relative pose of a person's limb with respect to the capacitive sensor. We trained this model over 100 epochs using the Adam optimizer with $\beta_1=0.9, \beta_2=0.999$, a batch size of 128, and a learning rate of 0.001.

At any given time step $t$, our implementation of this model takes as input a window of the 50 most recent measurements from the six capacitive sensor electrodes, $\bm{c}^{}_{t-49:t}\in\mathbb{R}^{50\times 6}$. When measured at a frequency of 100~Hz, this results in a 0.5 second window of prior capacitance data. We then vectorize these measurements for the network, resulting in a model of the form $f: \mathbb{R}^{300} \rightarrow \mathbb{R}^4$. By using a window over prior measurements, this model is able to learn the temporal aspects of how capacitance measurements vary over time as the sensor translates and rotates around human limbs. This approach also provides an opportunity to implicitly model and account for crosstalk (electromagnetic interference) between the co-located sensors during human-robot interaction.

\subsection{Data Collection}
\label{sec:data_collection}

We trained our human pose estimation model for capacitive servoing using approximately 630,000 data pairs $(\bm{c}^{}_{t-h+1:t}, \bm{p}^{}_t)$ of time-varying capacitance measurements and relative limb poses. These data pairs were captured as a PR2 robot moved the capacitive sensor around a human participant's stationary arm and leg. Data Collection~\ref{alg:datacollect} in the Appendix outlines the capacitance data collection process between the robot and a human participant.

\begin{figure}
\centering
\includegraphics[width=0.23\textwidth, trim={12cm 6cm 14cm 0cm}, clip]{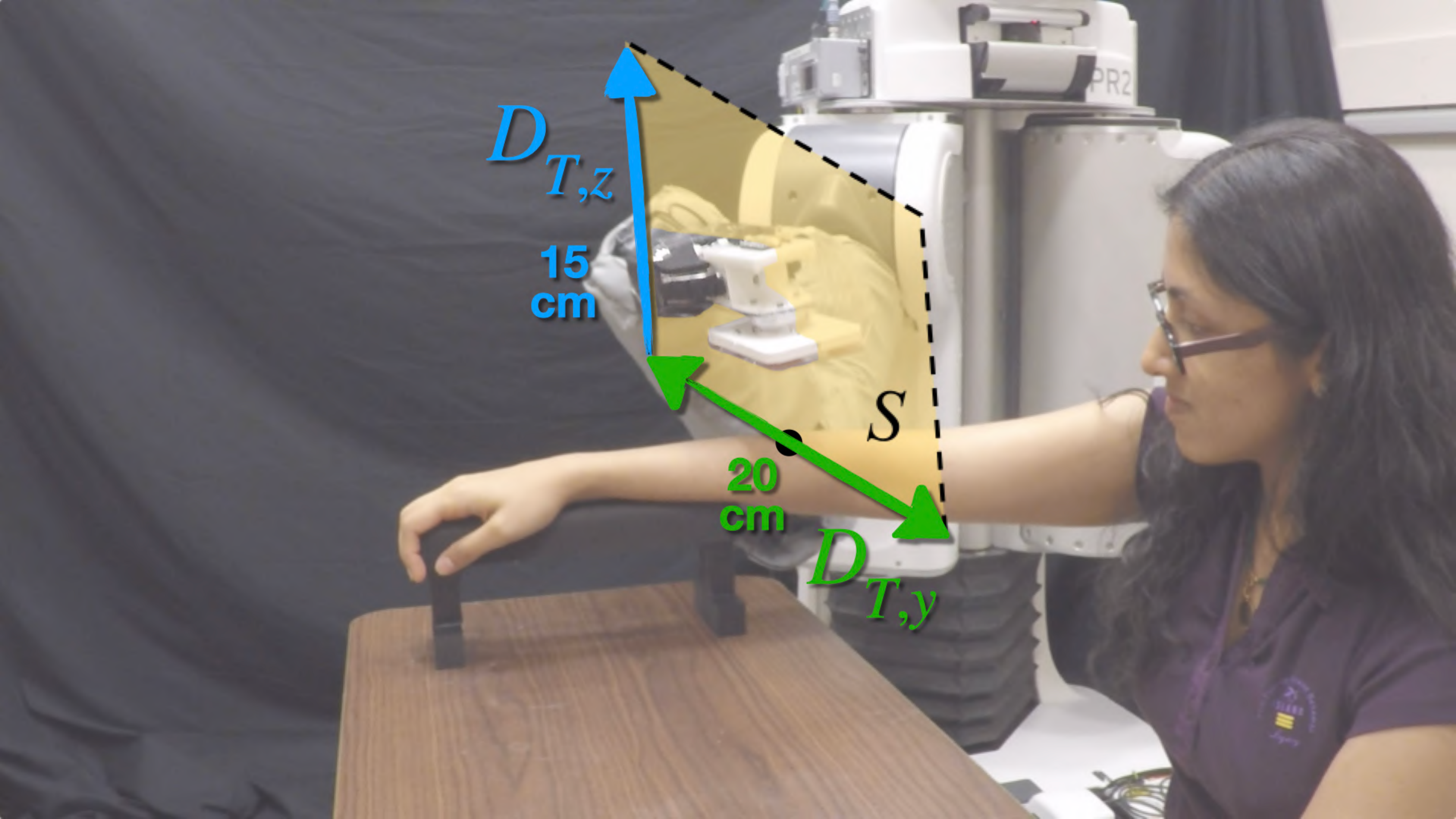}
\includegraphics[width=0.23\textwidth, trim={6cm 6cm 20cm 0cm}, clip]{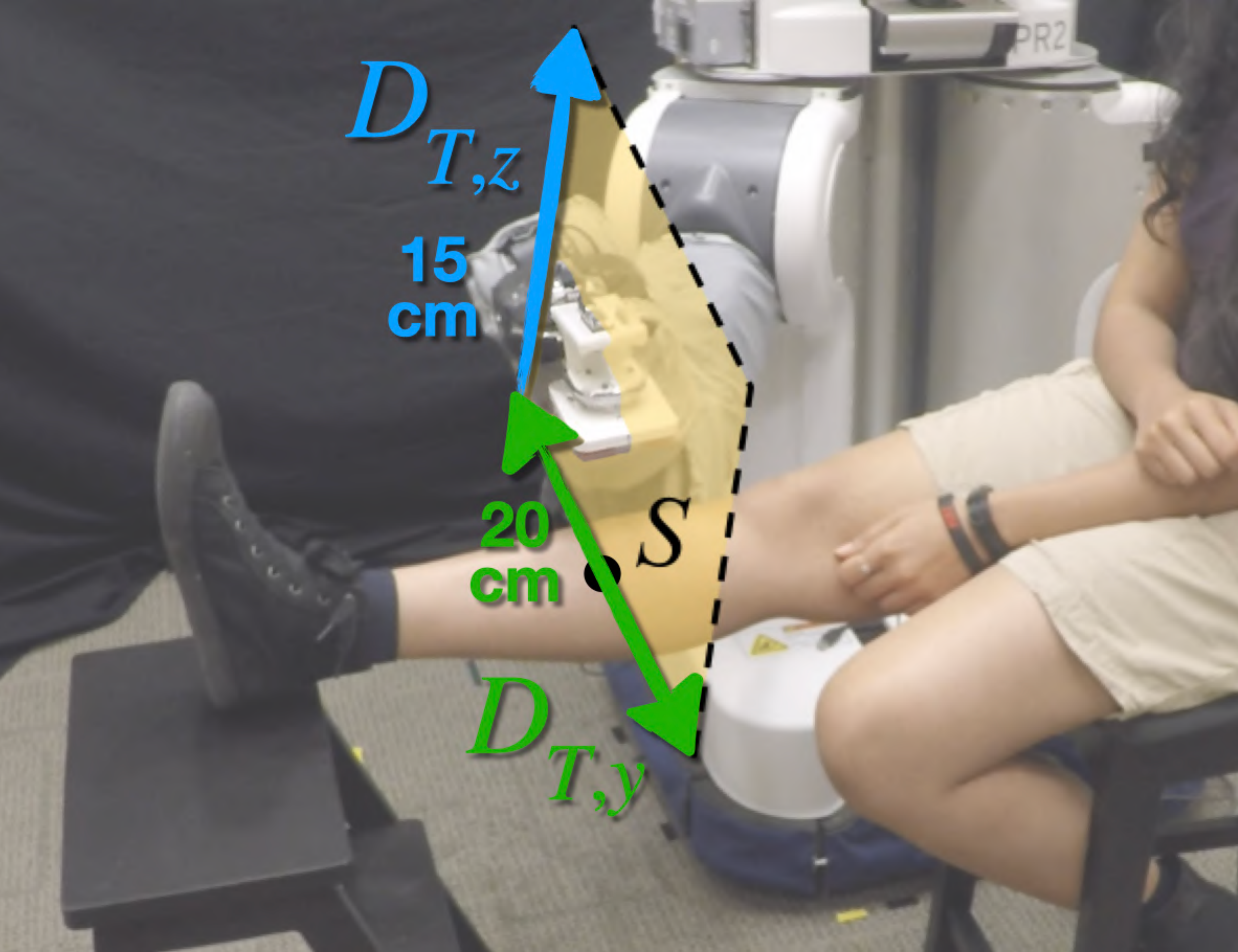}
\caption{\label{fig:datacollection}During data collection, a participant elevated their arm and leg at a stationary pose parallel to the ground using an armrest and footstool. The capacitive sensor then started at a relative pose of $\bm{p}^{}_0 = (0, 0, 0, 0)$ above the limb. The yellow highlighted region represents the space of target end effector positions $(D^{}_{T,y}, D^{}_{T,z})$ relative to the point $\bm{k}^*$ on the surface of the limb.}
\end{figure}

As shown in Fig.~\ref{fig:datacollection}, the participant began with their arm or leg elevated outward parallel to the ground with the support of an armrest and footstool. We position the robot's end effector and capacitive sensor above the participant's limb with an initial relative pose of $\bm{p}^{}_0 = (0, 0, 0, 0)$. The robot then selects a target end effector pose above the limb $\bm{p}^{}_T$ from a uniform distribution, along with translational and rotational velocities for its end effector to move towards the target. The bounded space $S$ of end effector poses above a person's static limb is depicted in Fig.~\ref{fig:datacollection}. At each time step $t$ while moving its end effector to the target, the robot records capacitance measurements $\bm{c}^{}_t$ from the six capacitive sensor electrodes coupled with the current pose of the end effector $\bm{p}^{}_t$ from forward kinematics. 
Once the robot reaches the target state $\bm{p}^{}_{t} \approx \bm{p}^{}_T$, the robot iteratively selects and moves towards a new target pose $\bm{p}^{}_T$ for a total of $N=500$ end effector trajectories above the limb. The detailed data collection process is provided in the Appendix.

We repeat the data collection process for three locations along the arm (wrist, forearm, upper arm) and three locations along the leg (ankle, shin, knee). Conducting $N=500$ iterations resulted in $\sim$17.5 minutes of data collection for each location along the arm and leg. We recorded measurements at a frequency of 100~Hz ($\tau=100$) for a total of 630,000 samples $(\bm{c}^{}_t, \bm{p}^{}_t)$. We then windowed all of the observed data into pairs $(\bm{c}^{}_{t-h+1:t}, \bm{p}^{}_t)$ for model training and evaluation. We collected these measurements with a single human participant, and in Section~\ref{sec:servoing_eval} we demonstrate how a trained pose estimation model can generalize to multiple participants.

\subsection{Capacitive Servoing Feedback Control}
\label{sec:control}

At any given time step $t$ when a robot is operating near the human body (e.g. during the evaluation Section~\ref{sec:servoing_eval}), the robot observes a window of the most recent capacitance measurements $\bm{c}^{}_{t-h+1:t}$ from the capacitive sensor. As discussed in Section~\ref{sec:learning}, we can feed these measurements through our trained pose estimation model $f(\bm{c}_{t-h+1:t})$, which produces a relative pose estimate $\bm{\hat{p}}^{}_t$ of the nearby person's limb. In this section, we discuss the general control scheme used in capacitive servoing to adapt to human motion and follow the contours of a human limb. In addition, we detail our specific implementation of this controller, which our robot used to interact with human participants.

Algorithm~\ref{alg:control} outlines the capacitive servoing control scheme for robots, which uses a high-level Cartesian controller to define end effector trajectories. Before control begins, the robot fills the temporal window of capacitance measurements used for pose estimation (lines~\ref{line:2_2}-\ref{line:2_3}). At each time step during execution, the robot observes the current capacitance readings from all capacitive electrodes, $\bm{c}_t$ (line~\ref{line:2_6}). Just as during data collection (Section~\ref{sec:data_collection}), with our implementation the robot collects these measurements at a frequency of 100~Hz ($\tau_d=100$).

\begin{algorithm}[t]
\caption{Capacitive Servoing for Human Interaction}\label{alg:control}
\begin{algorithmic}[1]
\State \textbf{Given:} $f$: pose estimation model,\newline
$u$: control policy,\newline
$\bm{p}^{}_{desired}$: target end effector pose,\newline
$h$: capacitance window size,\newline
$v_x$: forward velocity,\newline
$\tau_d$: data capture frequency,\newline
$\tau_u$: control frequency.
\For{$t=1,\ldots, h-1$}\label{line:2_2}
    \State $\bm{c}^{}_t \gets$ GetCapacitanceMeasurements().\label{line:2_3}
\EndFor
\State $t \gets h$.
\While{\textit{task not completed}}\label{line:2_5}
    \State $\bm{c}^{}_t \gets$ GetCapacitanceMeasurements().\label{line:2_6}
    \If{$t \bmod \lfloor\frac{\tau_d}{\tau_u}\rfloor = 0$}
        \State $\bm{e}(t) \gets \bm{p}^{}_{desired} - f(\bm{c}^{}_{t-h+1:t})$.\label{line:2_8}
        \State Compute $(u^{}_y(t), u^{}_z(t), u^{}_{\theta_y}(t), u^{}_{\theta_z}(t))$ using (\ref{eq:controller}).\label{line:2_9}
        \State $\bm{u}_{t,x} \gets$ TransformToRobotFrame($(\frac{v_x}{\tau_u}, 0, 0)$).\label{line:2_10}
        \State $\bm{\omega}^{}_t, \bm{\phi}^{}_t \gets$ GetCurrentEndEffectorPose().\label{line:2_11}
        \State $\bm{\omega}^{}_t \gets \bm{\omega}^{}_t + (0, u^{}_y(t), u^{}_z(t)) + \bm{u}_{t,x}$.
        \State $\bm{\phi}^{}_t \gets \bm{\phi}^{}_t + (0, u^{}_{\theta_y}(t), u^{}_{\theta_z}(t))$.
        \State $\bm{\alpha}^{}_t \gets$ InverseKinematics($\bm{\omega}^{}_t, \bm{\phi}^{}_t$).
        \State SendToActuators($\bm{\alpha}^{}_t$).\label{line:2_15}
    \EndIf
    \State $t \gets t+1$.
\EndWhile
\end{algorithmic}
\end{algorithm}

In order to update its end effector trajectory, the robot begins by computing the current tracking error $\bm{e}(t)$ (line~\ref{line:2_8}). This error is calculated as the difference between the desired pose offset $\bm{p}^{}_{desired} = (D^{}_y, D^{}_z, \theta^{}_y, \theta^{}_z)$ and the estimated relative pose $\bm{\hat{p}}^{}_t$ from our trained model $f(\bm{c}^{}_{t-h+1:t})$.

\begin{figure}
\centering
\includegraphics[width=0.48\textwidth, trim={5cm 6cm 6cm 6cm}, clip]{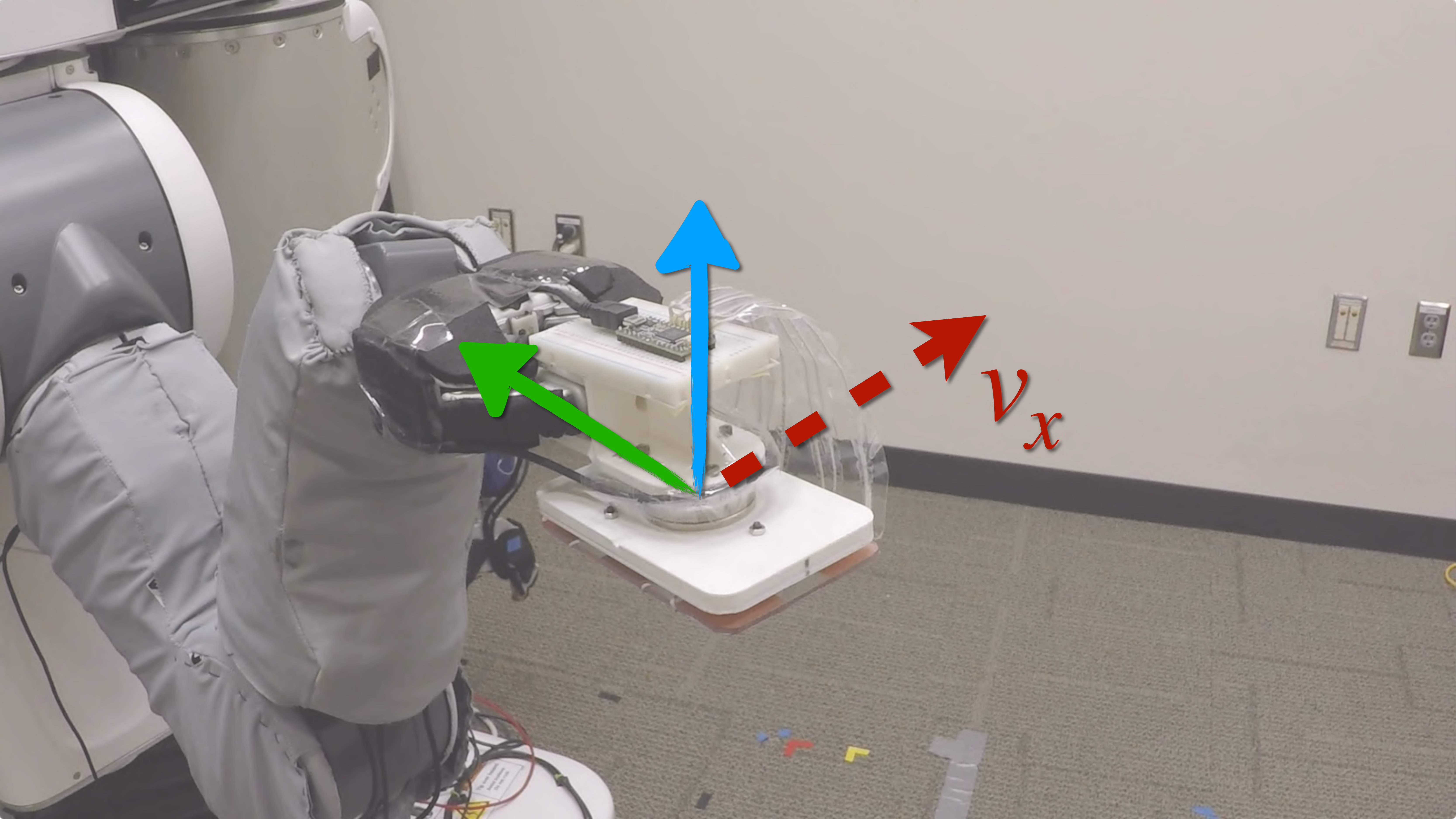}
\caption{\label{fig:endeffector}In addition to the control action $u(t)$ during capacitive servoing, the robot moves along the X-axis of its end effector at a velocity of $v_x$. Doing so ensures the robot navigates along the surface of a human limb during physical interaction.}
\end{figure}

The robot then computes a proposed action $\bm{u}(t) = (u^{}_y(t), u^{}_z(t), u^{}_{\theta_y}(t), u^{}_{\theta_z}(t))$ (line~\ref{line:2_9}) towards the target pose $\bm{p}^{}_{desired}$. This action consists of both translation $(u^{}_y(t), u^{}_z(t))$ and orientation $(u^{}_{\theta_y}(t), u^{}_{\theta_z}(t))$ steps for the pose of the end effector.
The robot also moves along the X-axis of its end effector at a fixed velocity $v_x$ (see Fig.~\ref{fig:endeffector}), which ensures that the end effector traverses either proximally or distally along a person's limb. To do so, the robot transforms this fixed velocity action (line~\ref{line:2_10}) from the end effector coordinate frame to its central base coordinate frame.
To execute this entire action, we form a new target end effector pose for the robot by adding the action to the current end effector pose $\bm{\omega}^{}_t, \bm{\phi}^{}_t$ and then use inverse kinematics to compute target angles for each robot actuator (lines~\ref{line:2_11}-\ref{line:2_15}).

In our implementation, we define a PD (proportional-derivative) feedback controller to compute actions $\bm{u}(t)$ for the robot's end effector. We define this controller as,
\begin{equation}
    \begin{aligned}
        \bm{u}(t)  =  \bm{K}^{}_p \bm{e}(t) + \bm{K}^{}_d \bm{\dot{e}}(t).
    \end{aligned} \label{eq:controller}
\end{equation}
Here, $\bm{K}^{}_p = \text{diag}(0.025, 0.025, 0.1, 0.1)$ and $\bm{K}^{}_d = \text{diag}(0.0125, 0.0125, 0.025, 0.025)$ represent diagonal matrices for the proportional and derivative gains, which we tuned to generate smooth end effector motion when tracking the contours of a person's limb. We also instruct the robot to move 2~cm/s along the X-axis of its end effector, i.e. $v_x = 2$.
In this work, we emphasize that capacitive servoing can be a successful control technique around the human body, even with straightforward feedback controllers, such as PD control.

In comparison to the measurement capture frequency, which we run at 100~Hz, this control sequence (lines~\ref{line:2_8}-\ref{line:2_15}) can be run at lower frequencies (e.g. at 10~Hz with $\tau_u = 10$). Doing so can be useful in cases where performing a forward pass through the pose estimation model $f$ is slower than the data capture frequency. The robot performs this control loop (line~\ref{line:2_5}) continuously until a given task is complete.
For the tasks explored in this article, capacitive servoing continues until the end effector has traversed a person's entire limb of length $l$ (as determined by end effector velocity along the X-axis $v_x$), such as navigating from the person's hand up to their shoulder. Additionally, we implement a force threshold monitor using the 6-axis ATI force/torque sensor mounted on the assistive tool, which halts the robot's movements if the robot makes contact with a person and applies a magnitude of force greater than 10~N.

We control 4 degrees of freedom for the end effector along the Y- and Z-axes $(D^{}_{t,y}, D^{}_{t,z}, \theta^{}_{t,y}, \theta^{}_{t,z})$ and we control translation along the X-axis via $v_x$. We do not control or constrain rotation around the X-axis, which provides the robot and inverse kinematics some flexibility as it rotates around sharply bent limb joints, as depicted in Section~\ref{sec:servo_around_limbs}. In practice, we did not observe significant drift in end effector rotation along the X-axis $\theta^{}_{t,x}$, except in cases when the human limb was farther away from the sensor than the sensor's functional 15~cm sensing range (sensing ranges are characterized in Section~\ref{sec:sensing_ranges}).

With this controller setup, $\bm{p}^{}_{desired}$ can be dynamically changed to accommodate the robot's current task. In our evaluations with human participants (Section~\ref{sec:servoing_eval}), we set $\bm{p}^{}_{desired} = (0, 5, 0, 0)$ so that the robot's end effector stays 5~cm above a participant's limb and maintains the same orientation as the limb. For other tasks, such as providing bathing assistance (see Section~\ref{sec:assistive_tasks}), we can set $\bm{p}^{}_{desired} = (0, 1, 0, 0)$ so that a washcloth attached to the capacitive sensor makes direct contact with a person's skin.

\section{Capacitive Servoing Characterization}
\label{sec:servoing_eval}

In order to characterize the proposed capacitive servoing technique, including the trained pose estimation model and capacitive sensor limits, we conducted a human-robot study with 12 able-bodied human participants (6 females and 6 males).

We conducted several experiments with each participant to evaluate sensing ranges and generalization of capacitance measurements across people. As part of our experiments, we evaluated how a PR2 robot can use multidimensional capacitive servoing to move its end effector along human limbs---a task which is valuable for robotic caregiving and several physical human-robot interaction contexts. During these studies, the robot used the six-electrode capacitive sensor array described in Section~\ref{sec:method}.

We obtained informed consent from all participants and approval from the Georgia Institute of Technology Institutional Review Board (IRB). We recruited participants to meet the following inclusion/exclusion criteria: $\geq$ 18 years of age; fluent in written and spoken English; and have not been diagnosed with ALS or other forms of motor impairments. Participant ages ranged from 18 to 27 years old. Participant arm lengths (measured from wrist to shoulder) varied between 56~cm to 64~cm and leg lengths (measured from ankle to hipbone) varied between 75~cm to 97~cm. Video sequences of experiments and results can be found in the supplementary video.

\subsection{Characterizing Sensing Ranges}
\label{sec:sensing_ranges}

\begin{figure}
\centering
\includegraphics[width=0.48\textwidth, trim={15cm 5cm 13cm 3cm}, clip]{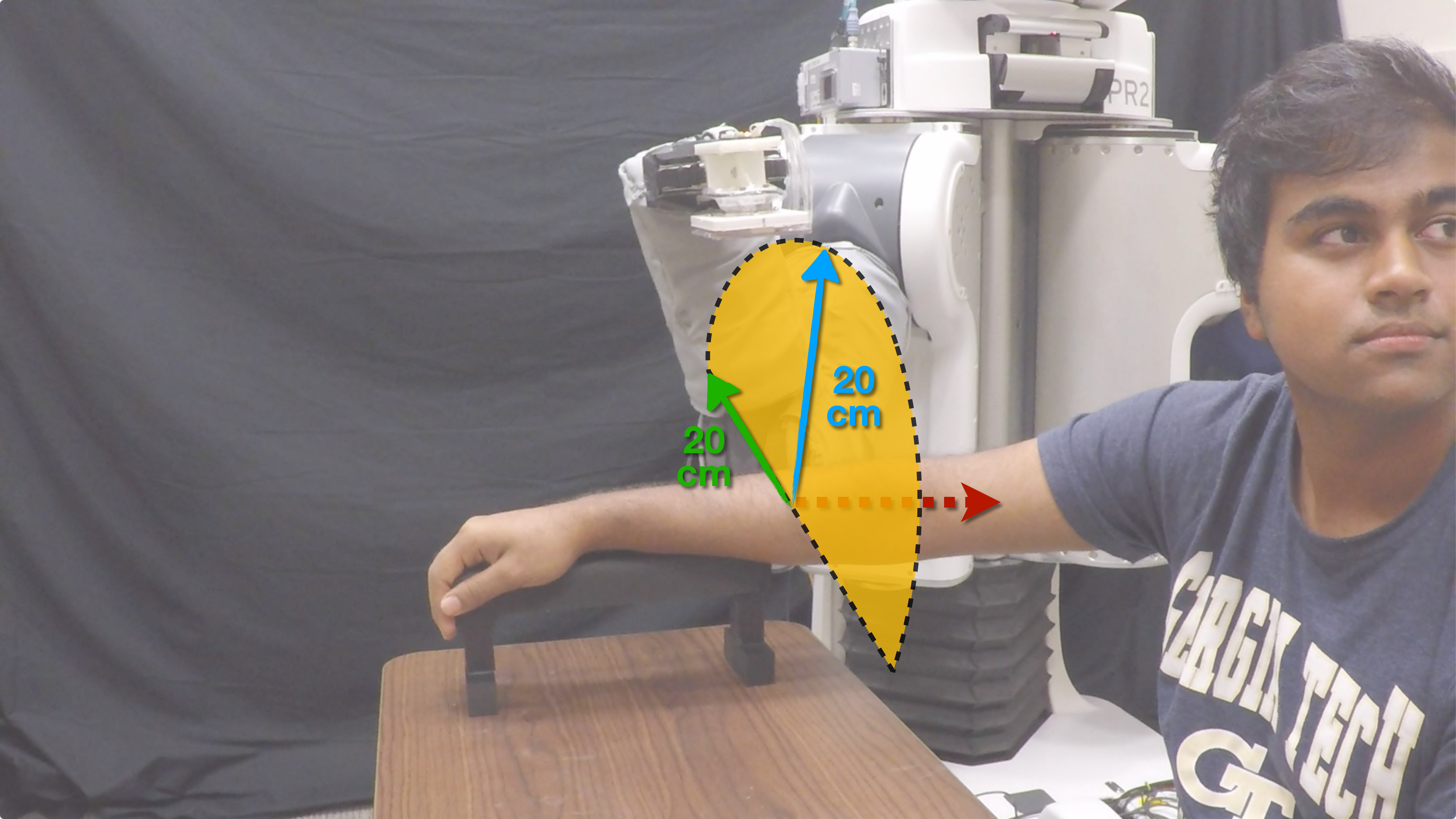}
\caption{\label{fig:evaluation_region}To evaluate the sensing ranges of the implemented six-electrode capacitive sensor, the PR2 recorded data pairs $(\bm{c}^{}_t, \bm{p}^{}_t)$ while following 20 linear trajectories above each participant's forearm. The robot would linearly translate its end effector in a random direction within the Y-Z plane until the capacitive sensor was 20~cm from the forearm. This resulted in the semicircle region of target end effector positions shown in yellow.}
\end{figure}

We began by characterizing the effective sensing range of the six-electrode capacitive sensor and trained pose estimator. We did this by computing pose estimation error as the PR2 moved the capacitive sensor away from each participant's stationary arm.

We first investigated the translational sensing range of the capacitive sensor along the two relative position axes $(D^{}_y, D^{}_z)$. The PR2 started by holding the capacitive sensor centered 3~cm above each participant's forearm. The robot's end effector then followed a linear trajectory in a random direction away from a participant's forearm within the Y-Z plane until the capacitive sensor was 20~cm away from the forearm. This resulted in the semicircle region of end effector positions shown in Fig.~\ref{fig:evaluation_region}. Throughout a linear trajectory, the robot recorded capacitance measurements at 100~Hz as the end effector maintained a velocity uniformly selected from [3~cm/s, 10~cm/s] and a relative orientation parallel to the participant's arm ($\theta_y=0$ and $\theta_z=0$). We use variable velocities to account for noise in actuation, variability in feedback control, and human motion, all which can occur in typical operation. Once a trajectory completed, the end effector returned to the initial starting point above a participant's forearm and repeated the process.
In total, the robot executed 20 trajectories above each of the 12 participant's forearms, resulting in 93,000 data pairs $(\bm{c}^{}_t, \bm{p}^{}_t)$ across 240 linear trajectories. We preprocessed these data into temporal windows of 50 capacitance measurements $\bm{c}^{}_{t-49:t}$, as discussed in Section~\ref{sec:learning}, and computed relative pose estimates using our trained model $f(\bm{c}^{}_{t-49:t}) = \bm{\hat{p}}^{}_t = (\hat{D}^{}_{t,y}, \hat{D}^{}_{t,z}, \hat{\theta}^{}_{t,y}, \hat{\theta}^{}_{t,z})$. 

These linear trajectories away from a limb intentionally differ slightly from the random end effector trajectories found in the training data distribution. These linear trajectories are closer aligned to the trajectories we expect our feedback controller to exhibit during typical operation and include end effector poses that extrapolate outside of the training distribution. In the following section we also discuss repeating the data collection process with all 12 participants and evaluating generalization across the resulting random end effector trajectories.

\begin{figure}
\centering
\includegraphics[width=0.48\textwidth, trim={0cm 15cm 0cm 1.5cm}, clip]{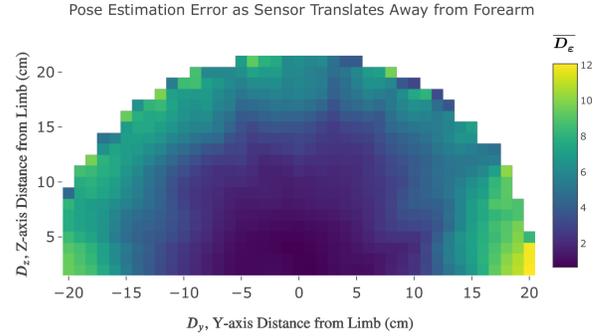}
\caption{\label{fig:distance}Mean estimation error $\overline{\bm{D^{}_\varepsilon}}$ (in centimeters) of the trained pose estimation model when the capacitive sensor was at various positions away from a human forearm. The X- and Y-axis show the ground truth lateral position $\bm{D}^{}_{y}$ and vertical position $\bm{D}^{}_{z}$ of the capacitive sensor away from the stationary human forearm, respectively. Each 1~cm $\times$ 1~cm grid cell with center $(y, z)$ is colored to represent the mean magnitude of pose estimation error present over all data pairs $(\bm{c}^{}_{t-49:t}, \bm{p}^{}_t)$ near the cell where $D^{}_y \in [y-1.5, y+1.5]$ and $D^{}_z \in [z-1.5, z+1.5]$. }
\end{figure}

We quantify position estimation errors by computing the mean absolute error $\overline{\bm{D^{}_\varepsilon}} = \frac{1}{2} || \bm{D}^{}_t - \bm{\hat{D}}^{}_t ||_1$ between the ground truth relative position $\bm{D}^{}_t = (D^{}_{t,y}, D^{}_{t,z})$ and estimated position $\bm{\hat{D}}^{}_t$ of a participant's forearm at time step $t$. Fig.~\ref{fig:distance} depicts the mean position estimation error $\overline{\bm{D^{}_\varepsilon}}$ at varying end effector locations away from a human forearm. The X-axis represents the ground truth relative lateral position $\bm{D}^{}_{y}$ between the capacitive sensor and human forearm, whereas the Y-axis enumerates the relative vertical position $\bm{D}^{}_{z}$. For each 1~cm $\times$ 1~cm grid cell with center $(y, z)$ in Fig.~\ref{fig:distance}, we computed the mean error over all data pairs $(\bm{c}^{}_{t-49:t}, \bm{p}^{}_t)$ that lie near the cell, i.e. $D^{}_y \in [y-1.5, y+1.5]$ and $D^{}_z \in [z-1.5, z+1.5]$.

When the center of the capacitive sensor array was within a 10~cm radius from a participant's forearm, the mean position estimation error $\overline{\bm{D^{}_\varepsilon}}$ was less than 1.1~cm---shown as the center dark blue region in Fig.~\ref{fig:distance}. Between a 10~cm to 15~cm radius from the arm, error increased to 2.8~cm. When the capacitive sensor was greater than 15~cm from the forearm, the mean position error from our pose estimation model was 5.6~cm, which we see by the green shading along the outer semicircle region in Fig.~\ref{fig:distance}. From this, we can postulate that this capacitive sensing technique and pose estimation model has a functional sensing range of up to 15~cm from a human arm. In Sections \ref{sec:servo_around_limbs} and \ref{sec:assistive_tasks}, we demonstrate that this sensing range is enough for a mobile manipulator to accurately follow the contours of a person's arm and assist participants with two activities of daily living.

We also quantify the rotational sensing range of the capacitive sensor along the two relative orientation axes $(\theta^{}_y, \theta^{}_z)$. The PR2 initialized its end effector above each participant's forearm such that the capacitive sensor started 5~cm above the forearm ($D^{}_y=0$~cm, $D^{}_z=5$~cm) with an initial orientation parallel to the arm ($\theta^{}_y=0$, $\theta^{}_z=0$). The robot then rotated its end effector to a random target orientation around the Y- and Z-axes until the capacitive sensor was oriented 45 degrees away from its starting orientation (while remaining 5~cm above the forearm), i.e. $||\bm{\theta}||_2=45$ degrees. We again repeated this process 20 times for each participant, for a total of 86,000 data pairs $(\bm{c}^{}_{t-49:t}, \bm{p}^{}_t)$ and associated pose estimates $f(\bm{c}^{}_{t-49:t})$ across 240 rotation trajectories. The supplemental video visually depicts these trajectories above a participant's forearm.

\begin{figure}
\centering
\includegraphics[width=0.48\textwidth, trim={0cm 8cm 0cm 1.5cm}, clip]{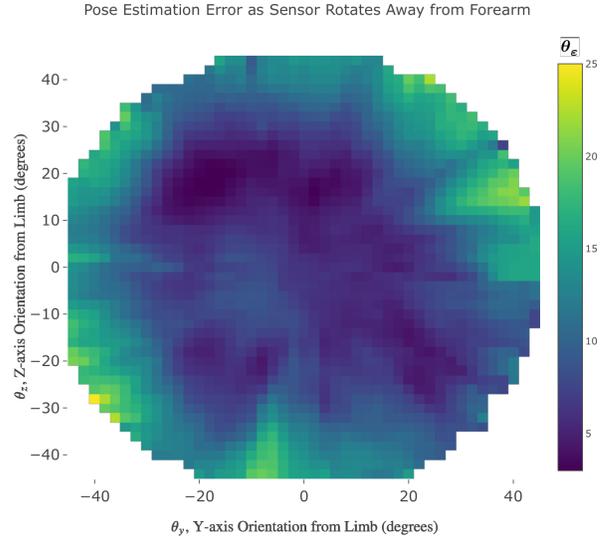}
\caption{\label{fig:angle}Mean estimation error $\overline{\bm{\theta^{}_\varepsilon}}$ (in degrees) of the trained pose estimation model when the capacitive sensor was at various orientations facing away from a human forearm. The pose estimation error in each 2$^\circ$ $\times$ 2$^\circ$ grid cell with center $(y, z)$ is averaged over all data pairs $(\bm{c}^{}_{t-49:t}, \bm{p}^{}_t)$ near the cell where $\theta^{}_y \in [y-3, y+3]$ and $\theta^{}_z \in [z-3, z+3]$. This circular region has a radius of 45 degrees, wherein the robot would rotate the capacitive sensor around the Y- and Z-axes until it was oriented 45 degrees from the sensor's starting orientation above a participant's forearm, i.e. $||\bm{\theta}||_2=45$ degrees.}
\end{figure}

We quantify orientation estimation errors via the mean absolute error $\overline{\bm{\theta^{}_\varepsilon}} = \frac{1}{2} || \bm{\theta}^{}_t - \bm{\hat{\theta}}^{}_t ||_1$ between the true relative orientation $\bm{\theta}^{}_t = (\theta^{}_{t,y}, \theta^{}_{t,z})$ and estimated orientation $\bm{\hat{\theta}}^{}_t$ of a participant's forearm. Fig.~\ref{fig:angle} visualizes the mean orientation estimation error $\overline{\bm{\theta^{}_\varepsilon}}$ as the capacitive sensor rotates above a human forearm in different directions. For all points in which the capacitive sensor had a less than 30 degree orientation from its starting orientation above a participant's forearm ($||\bm{\theta}_t||_2 < 30$ degrees), the mean orientation error was $\overline{\bm{\theta^{}_\varepsilon}} = 6.2$ degrees, depicted as the blue center region in Fig.~\ref{fig:angle}. Yet, when the sensor was between 30 to 45 degrees away from its starting orientation, the mean orientation estimation error increased to 11.0 degrees, as shown by the outer green region in Fig.~\ref{fig:angle}. From this, we conclude that the implemented capacitive sensor and pose estimator are most accurate when the relative orientation offset between the sensor and human limb is less than 30 degrees around the rotational axes. In Section~\ref{sec:servo_around_limbs}, we show that this orientation estimation is accurate enough to enable a PR2 to sense and adapt to local changes in a person's limb orientation, such as around an elbow or knee joint, or due to human motion.

\begin{figure*}
\centering
\includegraphics[width=0.19\textwidth, trim={6cm 2cm 7cm 1cm}, clip]{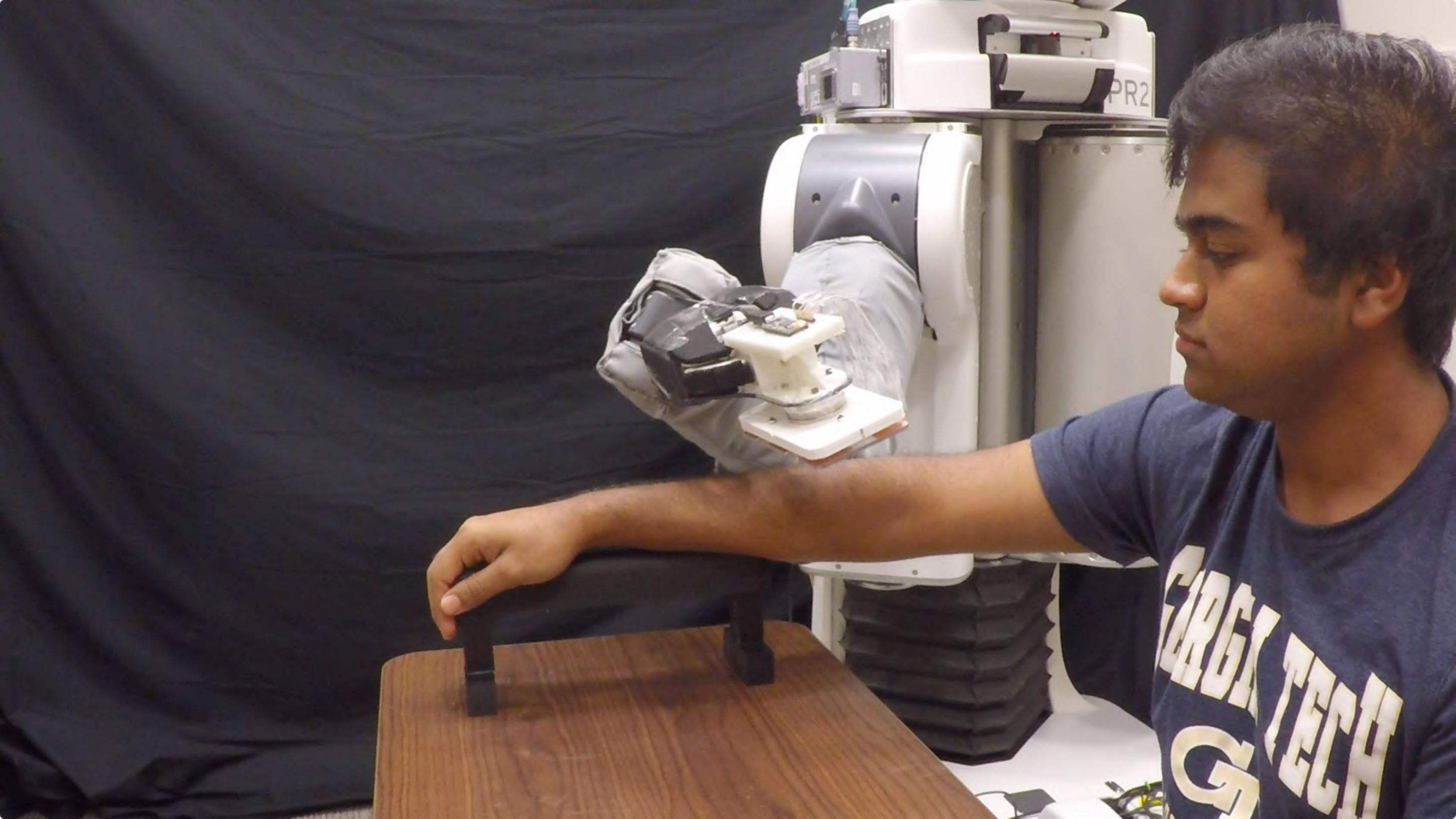}
\includegraphics[width=0.19\textwidth, trim={6cm 2cm 7cm 1cm}, clip]{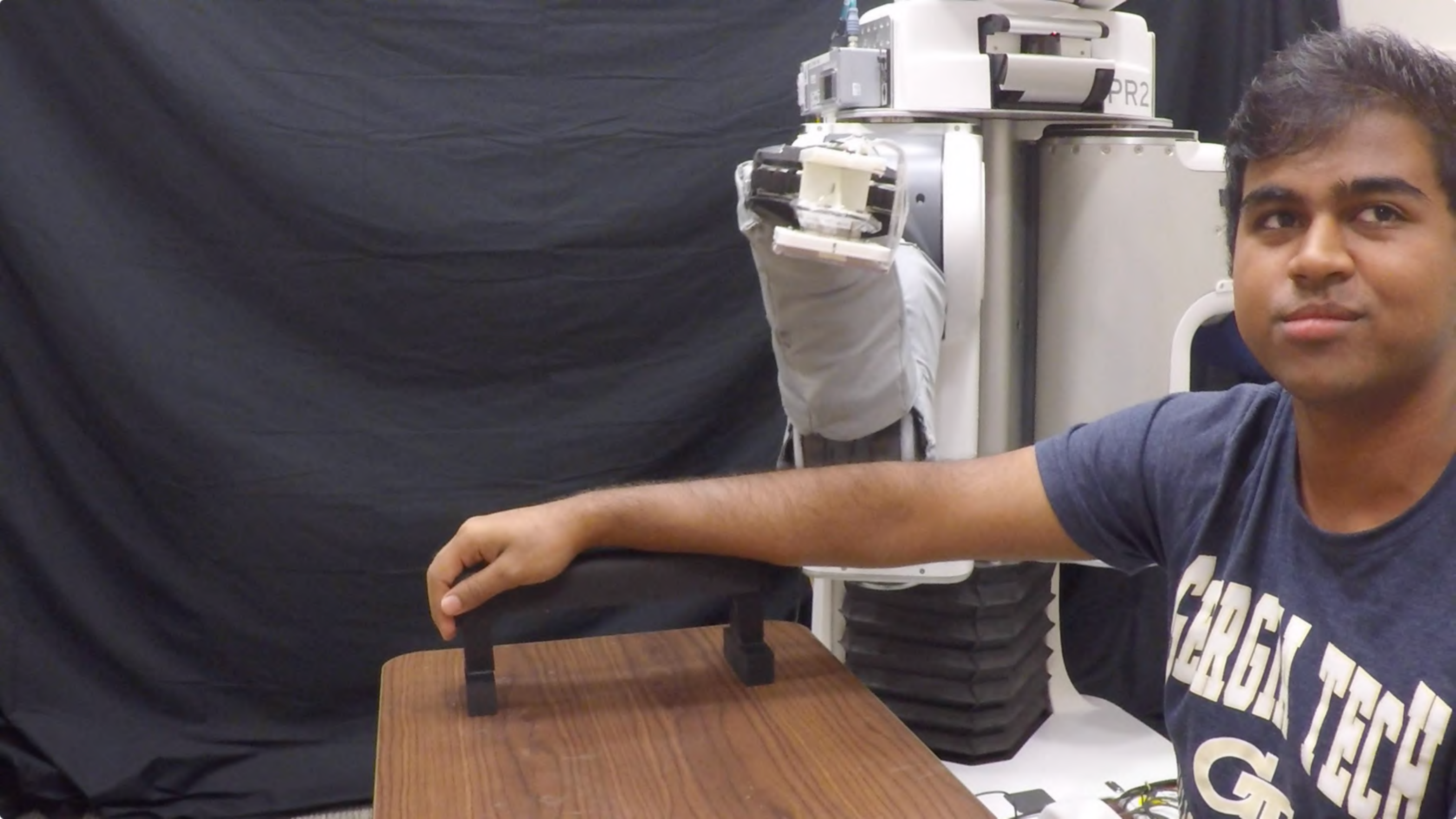}
\includegraphics[width=0.19\textwidth, trim={6cm 2cm 7cm 1cm}, clip]{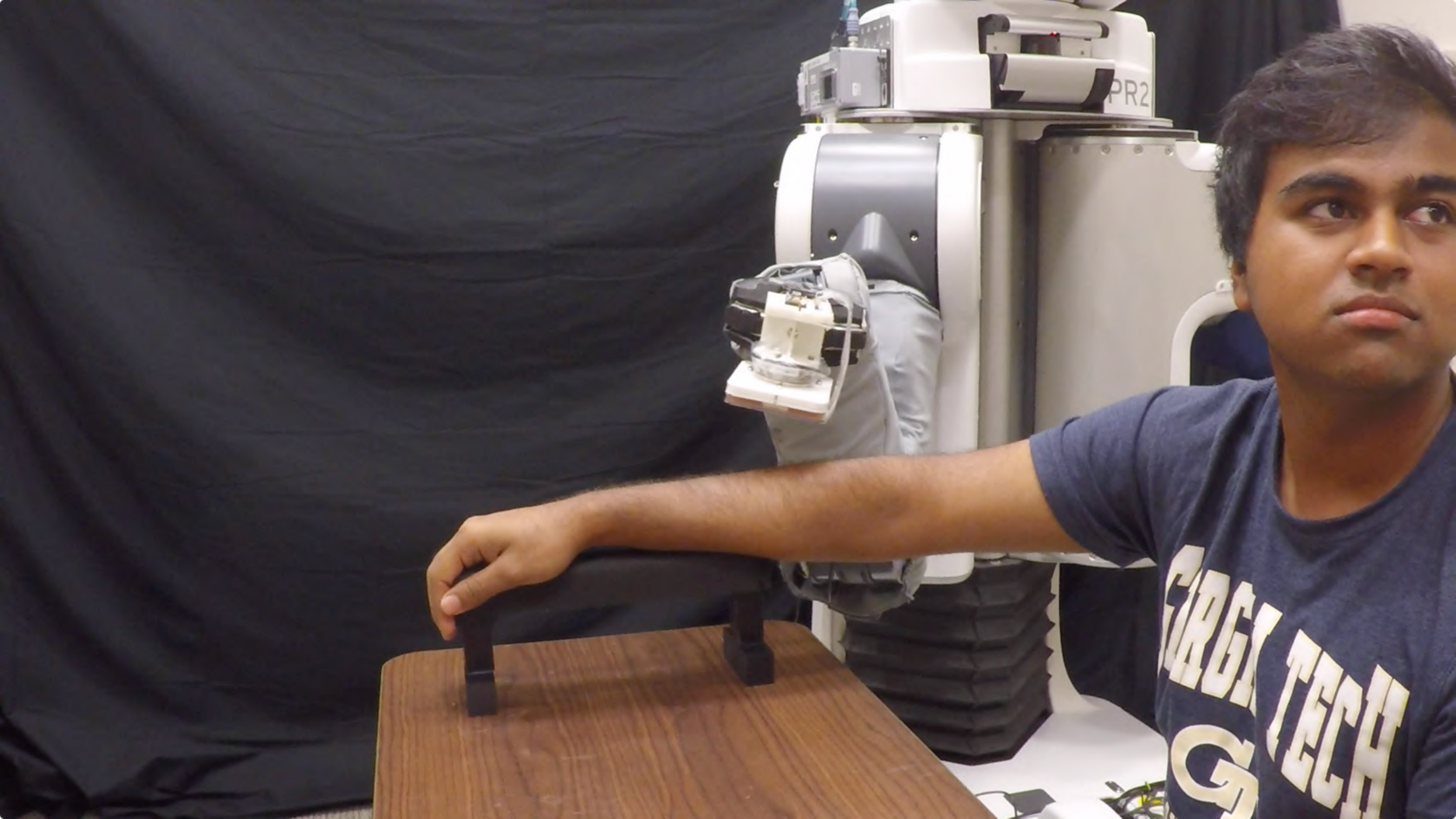}
\includegraphics[width=0.19\textwidth, trim={6cm 2cm 7cm 1cm}, clip]{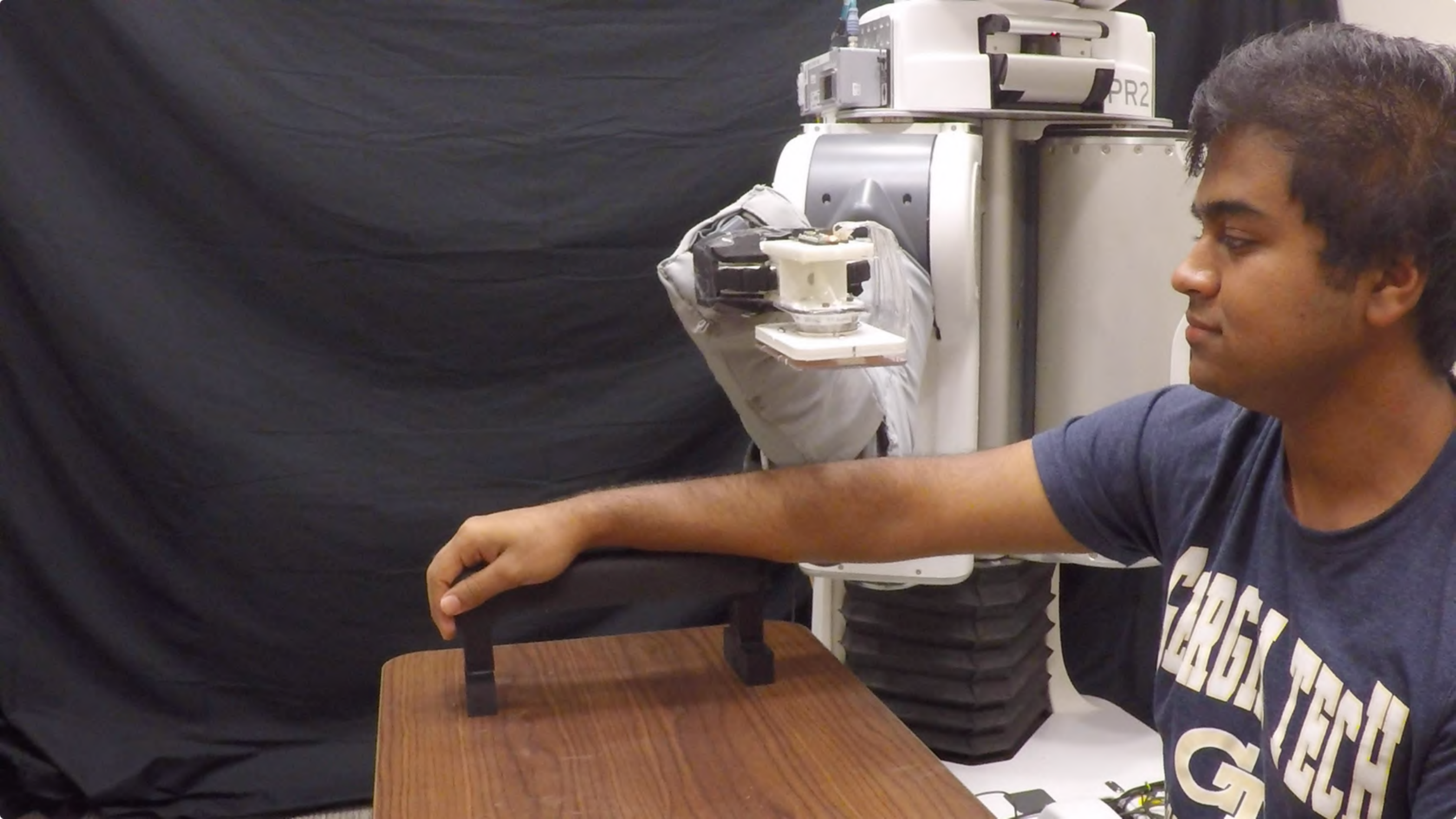}
\includegraphics[width=0.19\textwidth, trim={6cm 2cm 7cm 1cm}, clip]{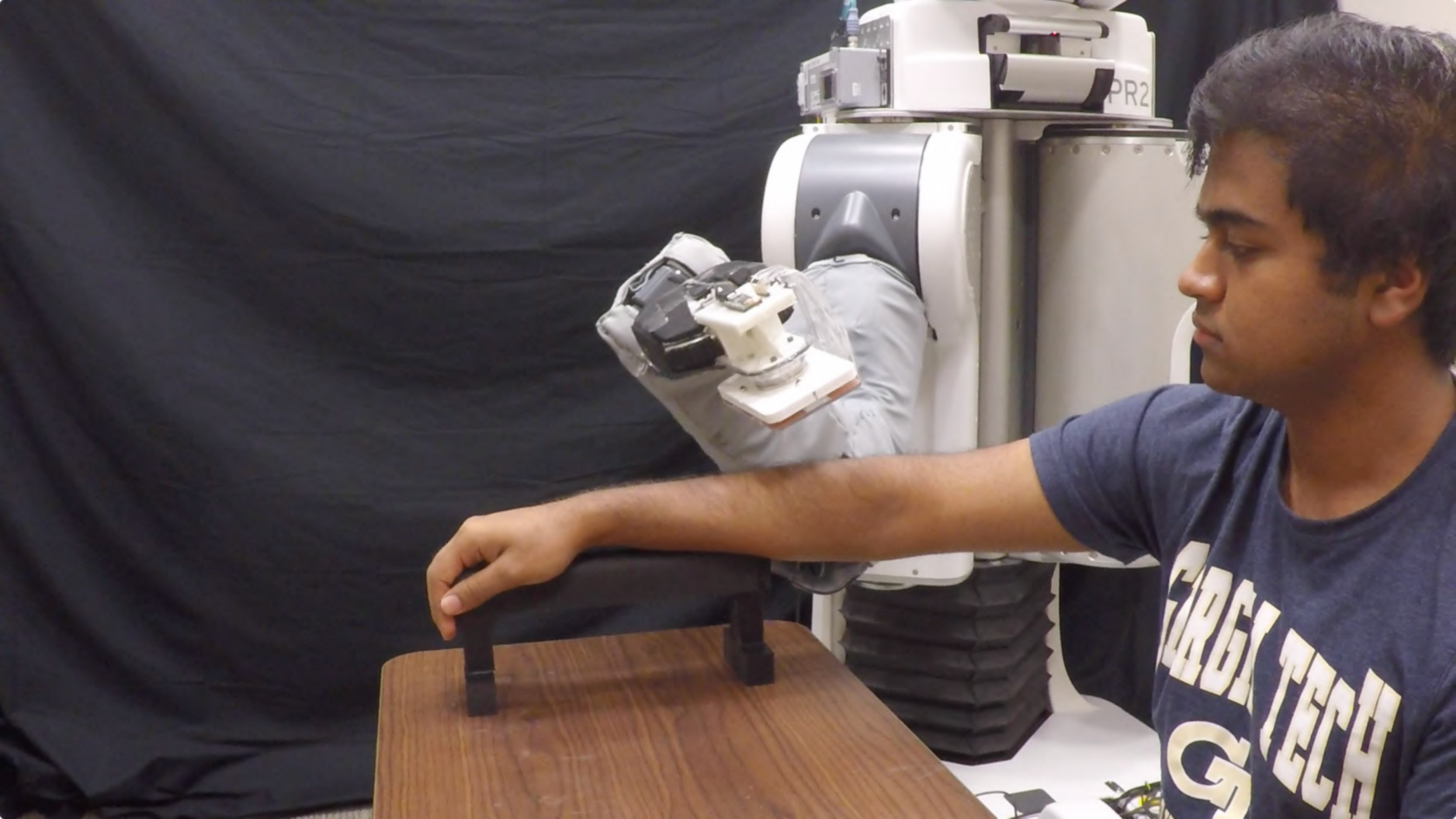}
\caption{\label{fig:data_collection_arm_generalization}The training data collection process visualized for one of the 12 participants. The PR2 collected capacitance and pose data pairs $(\bm{c}^{}_t, \bm{p}^{}_t)$ while performing 50 randomized trajectories above the participant's forearm. We used these data to evaluate generalization performance of capacitive sensing across participants of different body size.}
\end{figure*}

\begin{figure*}
\centering
\includegraphics[width=0.19\textwidth, trim={5cm 5cm 10cm 2cm}, clip]{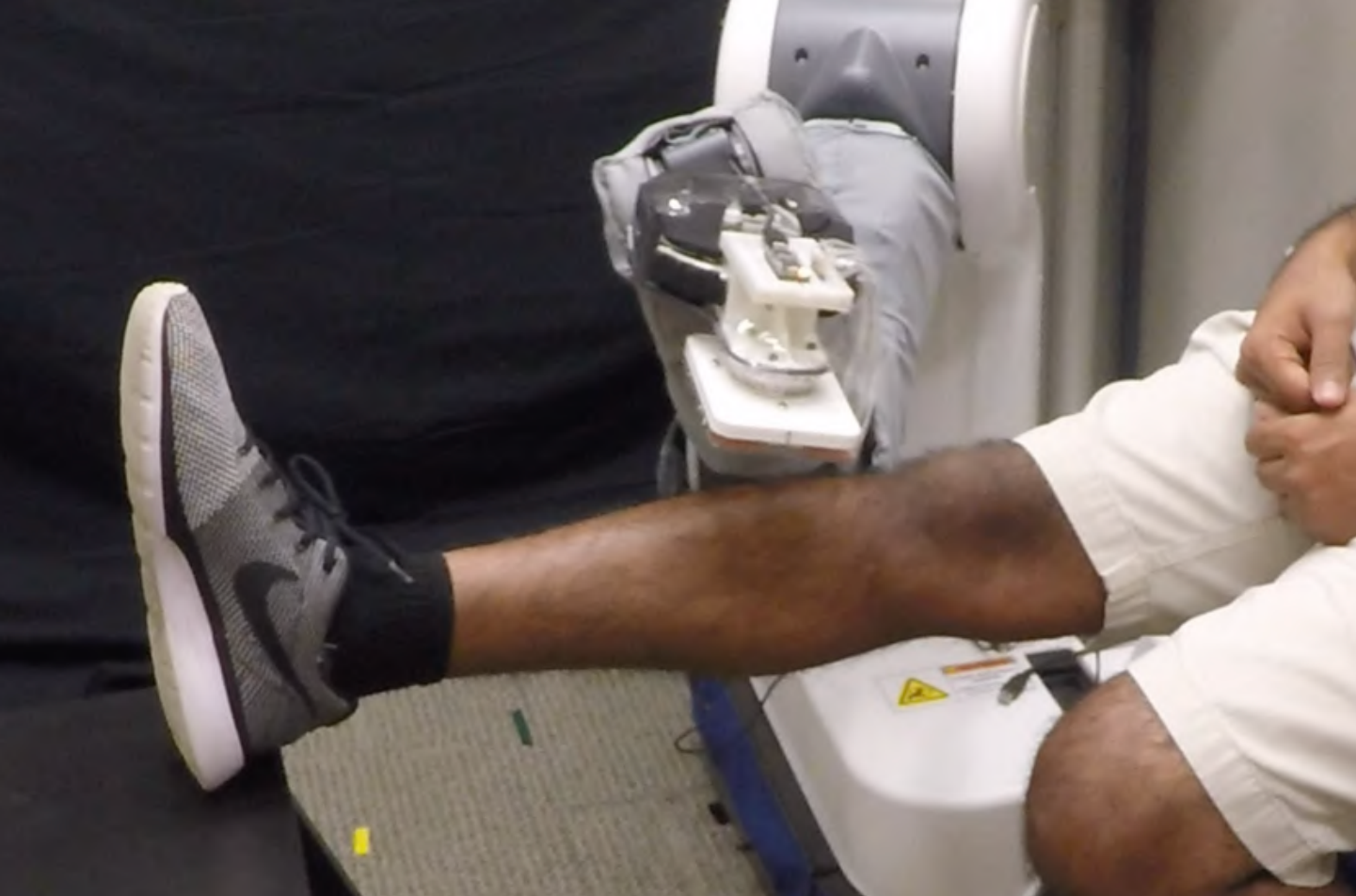}
\includegraphics[width=0.19\textwidth, trim={5cm 5cm 10cm 2cm}, clip]{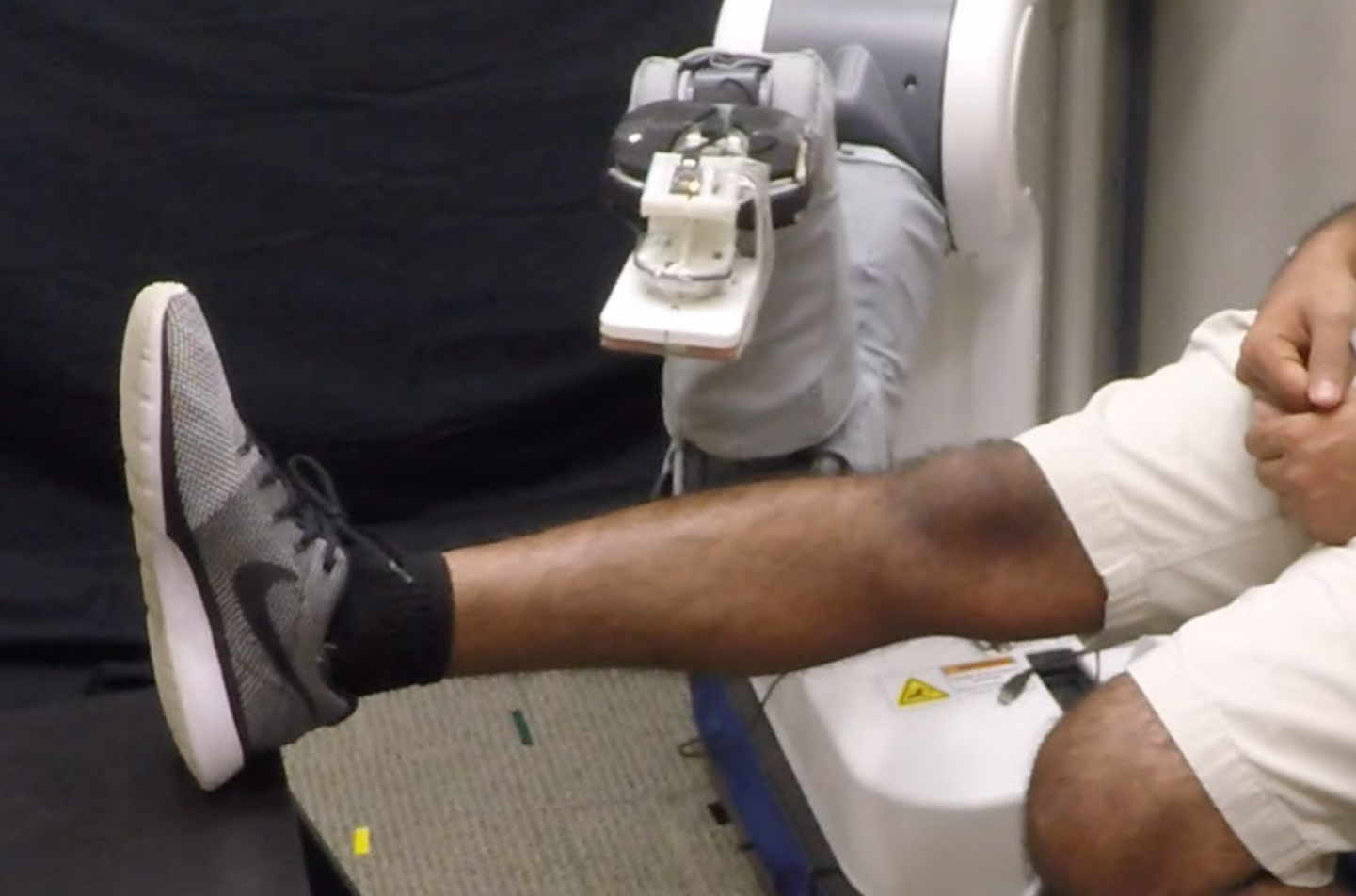}
\includegraphics[width=0.19\textwidth, trim={5cm 5cm 10cm 2cm}, clip]{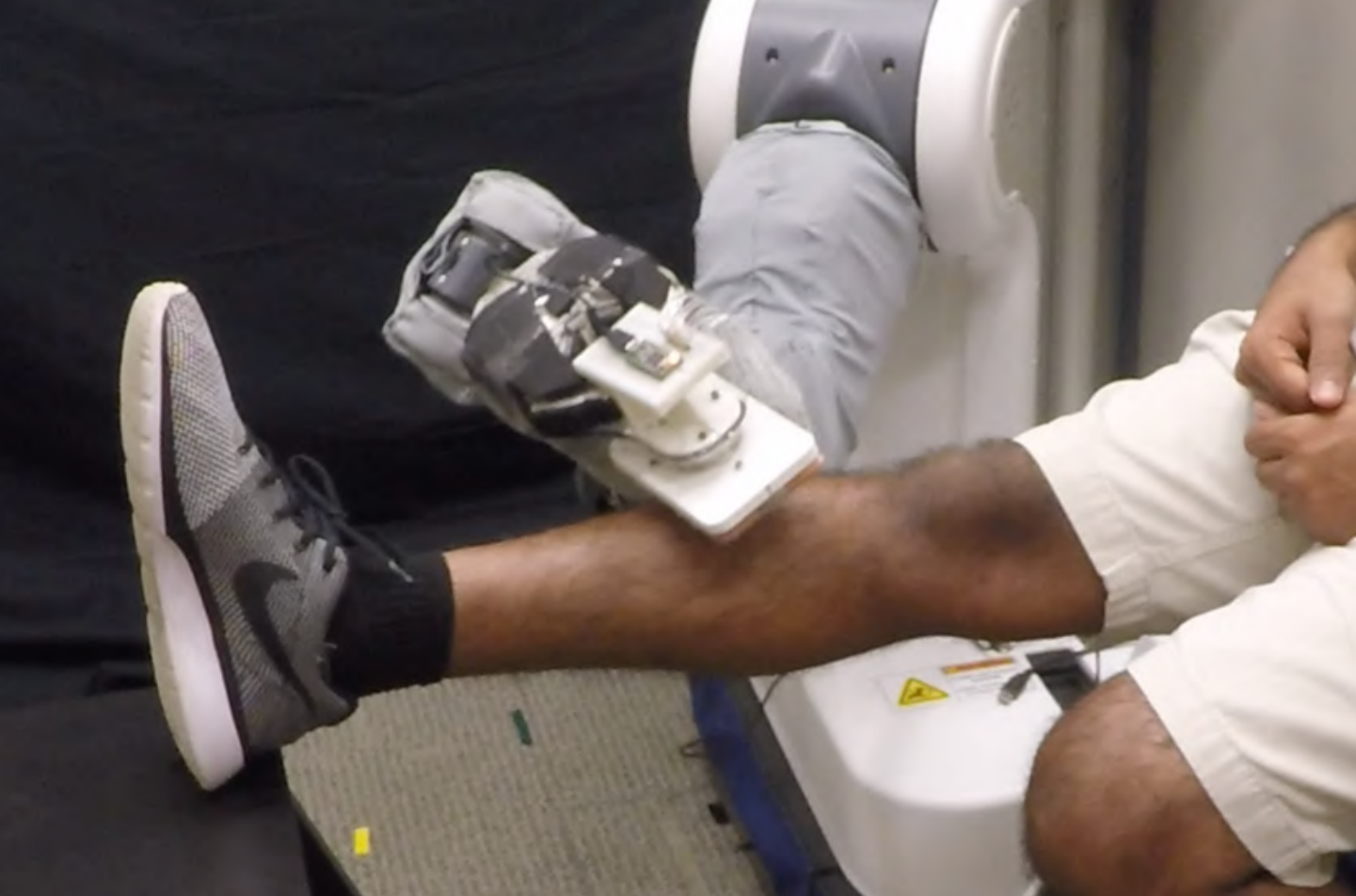}
\includegraphics[width=0.19\textwidth, trim={5cm 5cm 10cm 2cm}, clip]{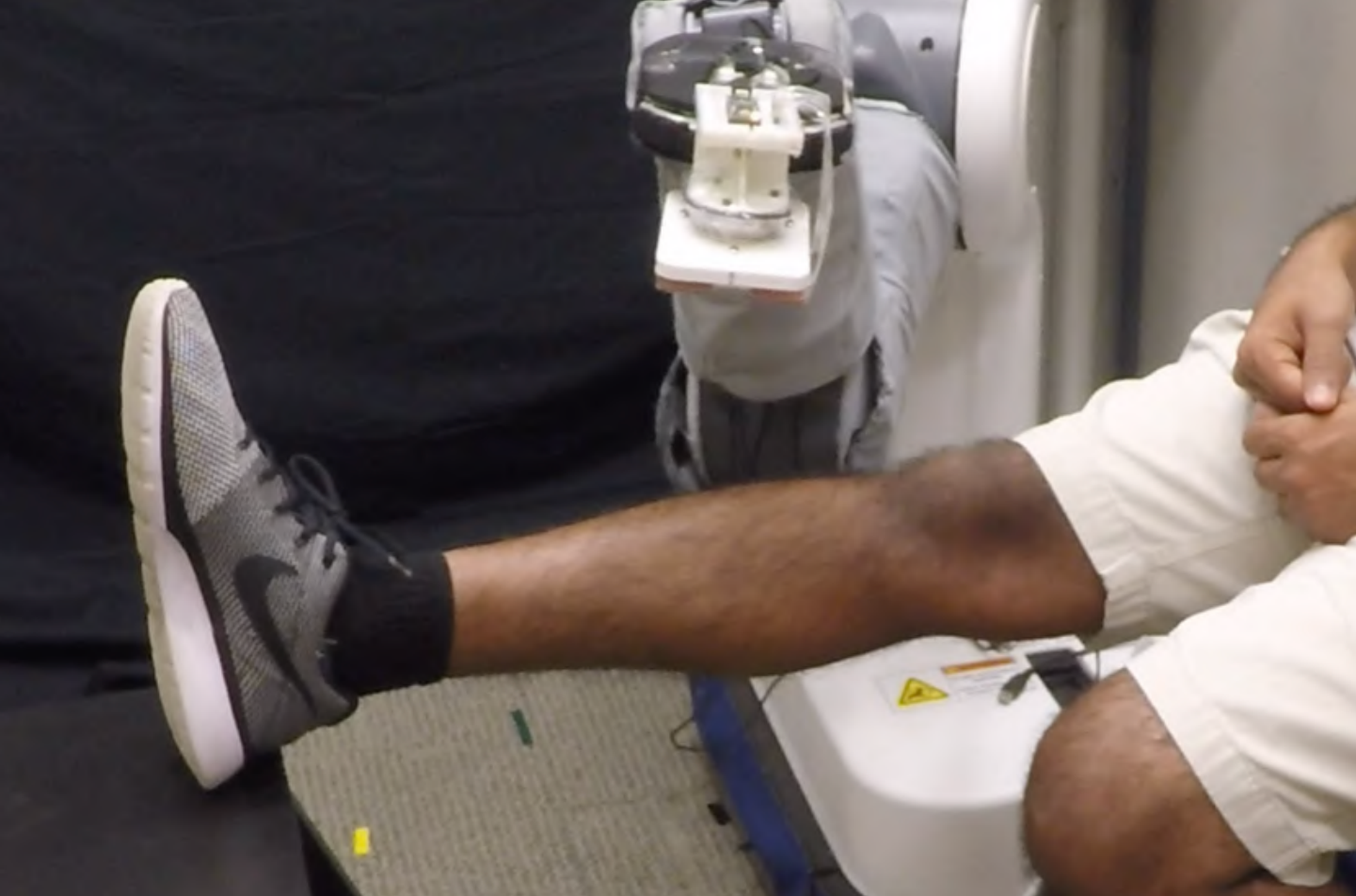}
\includegraphics[width=0.19\textwidth, trim={5cm 5cm 10cm 2cm}, clip]{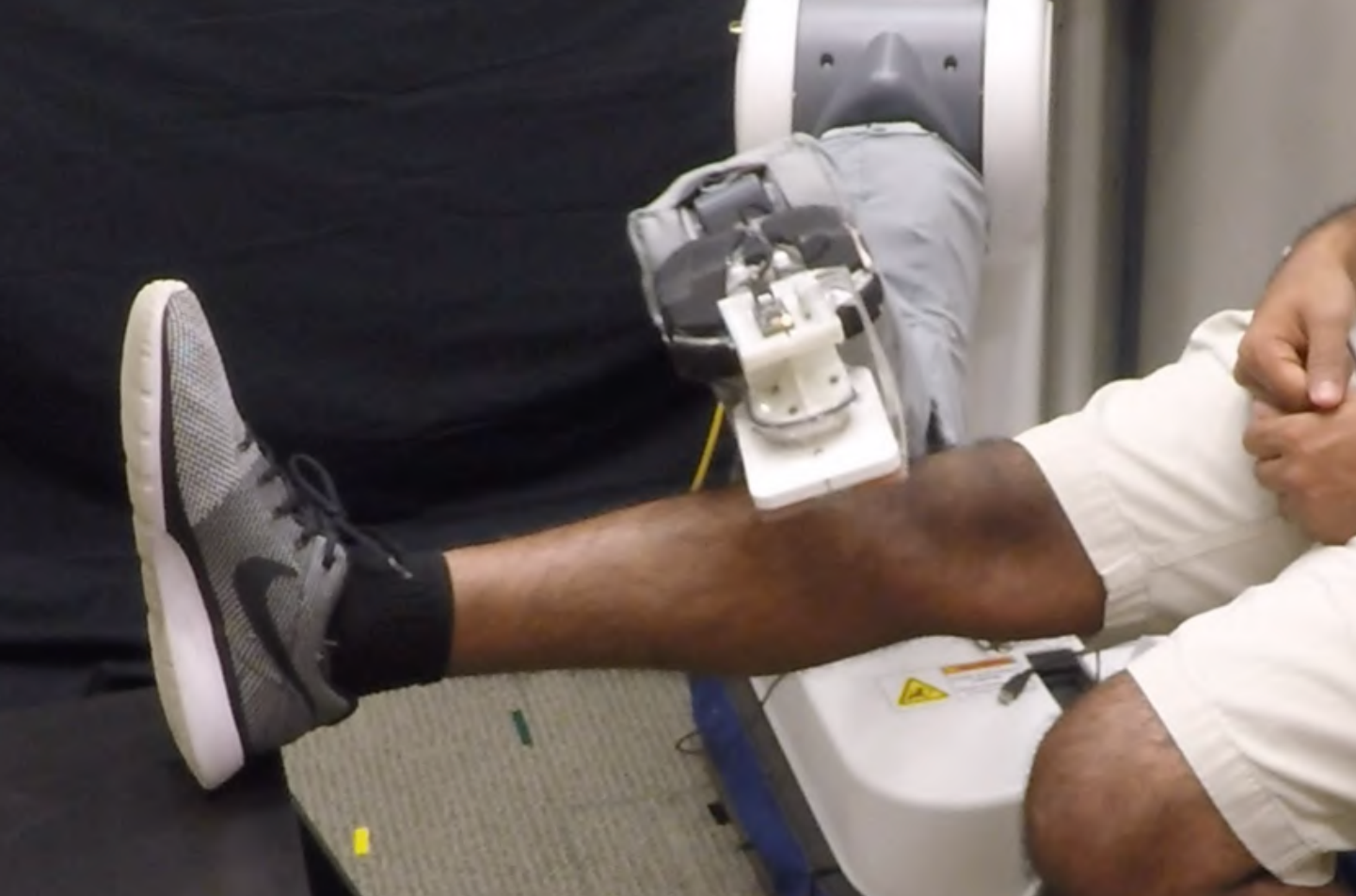}
\caption{\label{fig:data_collection_leg_generalization}The training data collection process (detailed in the Appendix) as the PR2 moved the capacitive sensor over a participant's shin.}
\end{figure*}

\subsection{Generalization Across People}

In this section, we evaluate the generalization performance of capacitive sensing for pose estimation. We do so by evaluating the accuracy of the proposed pose estimation model (see Section~\ref{sec:data_collection}) for estimating the limb poses of all 12 participants. We then contrast this performance to models trained on capacitance measurements from multiple people with different body sizes.

For this experiment, we repeat the training data collection process, shown in Data Collection~\ref{alg:datacollect} (Appendix), with all participants. For each participant, the PR2 collected data pairs $(\bm{c}^{}_t, \bm{p}^{}_t)$ while performing $N = 50$ uniformly random linear end effector trajectories above each of the six locations along the arm and leg (wrist, forearm, upper arm, ankle, shin, and knee). In total, we collected over 590,000 capacitance measurements across all 12 participants, with approximately 99,000 capacitance measurements above each of the six limb locations.
Fig.~\ref{fig:data_collection_arm_generalization} depicts several frames of this data collection process from one of the 12 participants as the PR2 moved the capacitive sensor above the participant's forearm.
Fig.~\ref{fig:data_collection_leg_generalization} visualizes data collection above the participant's shin.

We temporally window the capacitance data into pairs $(\bm{c}^{}_{t-49:t}, \bm{p}^{}_t)$, which we feed through our trained pose estimation model $f$. Table~\ref{table:generalization} depicts the pose estimation error of our model averaged over all 12 human participants for each of the sensing axes $(D^{}_y, D^{}_z, \theta^{}_y, \theta^{}_z)$. Although the pose estimation model was trained on capacitance data from only a single human participant, the model exhibits on average less than 2.5~cm of position estimation error and less than 10$^{\circ}$ of orientation error per axis. In practice, this model generalizes well enough to be used in a capacitive servoing control loop for performing assistive tasks around the human body (Section~\ref{sec:assistive_tasks}).

Similarly, we investigate the benefit of training capacitive pose estimation models on data from more than one human participant using leave-one-participant-out cross-validation. We retrain the pose estimation network (see Section~\ref{sec:learning}) on capacitance measurements from 11 participants and evaluate pose estimation performance on data from the one heldout participant. We repeat this process by holding out each participant, resulting in 12 trained pose estimation models and corresponding heldout test sets. In Table~\ref{table:lopo}, we report leave-one-participant-out results as the mean estimation error over all 12 models.

\begin{table}
\centering
\caption{\label{table:generalization}Pose estimation error per limb location averaged over all 12 participants. $D^{}_y$, $D^{}_z$ are in (cm). $\theta^{}_y$, $\theta^{}_z$ are in (degrees).}
\begin{tabular}{ccccc} \toprule
    & \multicolumn{4}{c}{Average error per sensing axis} \\ \cmidrule{2-5}
     & $D^{}_y$ (cm) & $D^{}_z$ (cm) & $\theta^{}_y$ (deg) & $\theta^{}_z$ (deg) \\ \midrule\midrule
    Wrist & 2.5~($\pm$2.6) & 1.8~($\pm$2.0) & 8.6~($\pm$6.9) & 9.9~($\pm$7.6) \\
    Forearm & 2.1~($\pm$2.2) & 1.6~($\pm$1.6) & 6.5~($\pm$5.7) & 9.2~($\pm$7.1) \\
    Upper arm & 2.6~($\pm$2.3) & 1.9~($\pm$1.6) & 7.7~($\pm$6.4) & 10.3~($\pm$7.7) \\
    Ankle & 2.0~($\pm$1.8) & 1.8~($\pm$1.6) & 8.1~($\pm$6.6) & 9.6~($\pm$7.2) \\
    Shin & 2.2~($\pm$2.1) & 1.7~($\pm$1.6) & 6.8~($\pm$5.9) & 9.1~($\pm$7.1) \\
    Knee & 2.2~($\pm$2.1) & 1.7~($\pm$1.4) & 8.2~($\pm$6.5) & 9.3~($\pm$7.1) \\ \midrule
    Average & \textbf{2.27}~cm & \textbf{1.75}~cm & \textbf{7.65}$^{\circ}$ & \textbf{9.57}$^{\circ}$ \\
	\bottomrule
\end{tabular}
\end{table}

\begin{table}
\centering
\caption{\label{table:lopo}Average error when training a pose estimation model on $n-1$ participants and evaluating on the heldout participant.}
\begin{tabular}{ccccc} \toprule
     & $D^{}_y$ & $D^{}_z$ & $\theta^{}_y$ & $\theta^{}_z$ \\ \midrule\midrule
    Average & 2.06~cm & 1.39~cm & 6.97$^{\circ}$ & 8.85$^{\circ}$ \\
	\bottomrule
\end{tabular}
\end{table}


\begin{table}
\centering
\caption{\label{table:minmax_errors}Pose estimation error for the participants with the minimum and maximum measured limb circumference.}
\begin{tabular}{ccccc} \toprule
    & \multicolumn{4}{c}{Average error per sensing axis} \\ \cmidrule{2-5}
    Limb (Circumference) & $D^{}_y$ & $D^{}_z$ & $\theta^{}_y$ & $\theta^{}_z$ \\ \midrule\midrule
    Wrist (13~cm) & 2.1~cm & 1.6~cm & 7.2$^{\circ}$ & 10.0$^{\circ}$ \\
    Wrist (18~cm) & 2.2~cm & 1.4~cm & 10.1$^{\circ}$ & 12.4$^{\circ}$ \\
    \\ 
    Forearm (22~cm) & 1.7~cm & 1.4~cm & 6.6$^{\circ}$ & 10.6$^{\circ}$ \\
    Forearm (29~cm) & 2.0~cm & 1.4~cm & 6.5$^{\circ}$ & 10.4$^{\circ}$ \\
    \\ 
    Upper arm (24~cm) & 2.1~cm & 1.6~cm & 8.4$^{\circ}$ & 11.7$^{\circ}$ \\
    Upper arm (30~cm) & 2.4~cm & 1.9~cm & 12.3$^{\circ}$ & 11.5$^{\circ}$ \\
    \\ 
    Ankle (20~cm) & 2.2~cm & 2.0~cm & 8.3$^{\circ}$ & 11.9$^{\circ}$ \\
    Ankle (26~cm) & 1.6~cm & 1.6~cm & 7.1$^{\circ}$ & 10.0$^{\circ}$ \\
    \\ 
    Shin (25~cm) & 3.0~cm & 2.4~cm & 7.6$^{\circ}$ & 9.1$^{\circ}$ \\
    Shin (40~cm) & 2.0~cm & 1.7~cm & 7.9$^{\circ}$ & 8.7$^{\circ}$ \\
    \\ 
    Knee (25~cm) & 2.0~cm & 1.5~cm & 7.7$^{\circ}$ & 9.2$^{\circ}$ \\
    Knee (37~cm) & 2.1~cm & 1.4~cm & 6.0$^{\circ}$ & 8.6$^{\circ}$ \\
	\bottomrule
\end{tabular}
\end{table}

\begin{figure*}
\centering
\includegraphics[width=0.24\textwidth, trim={15cm 8cm 6cm 1cm}, clip]{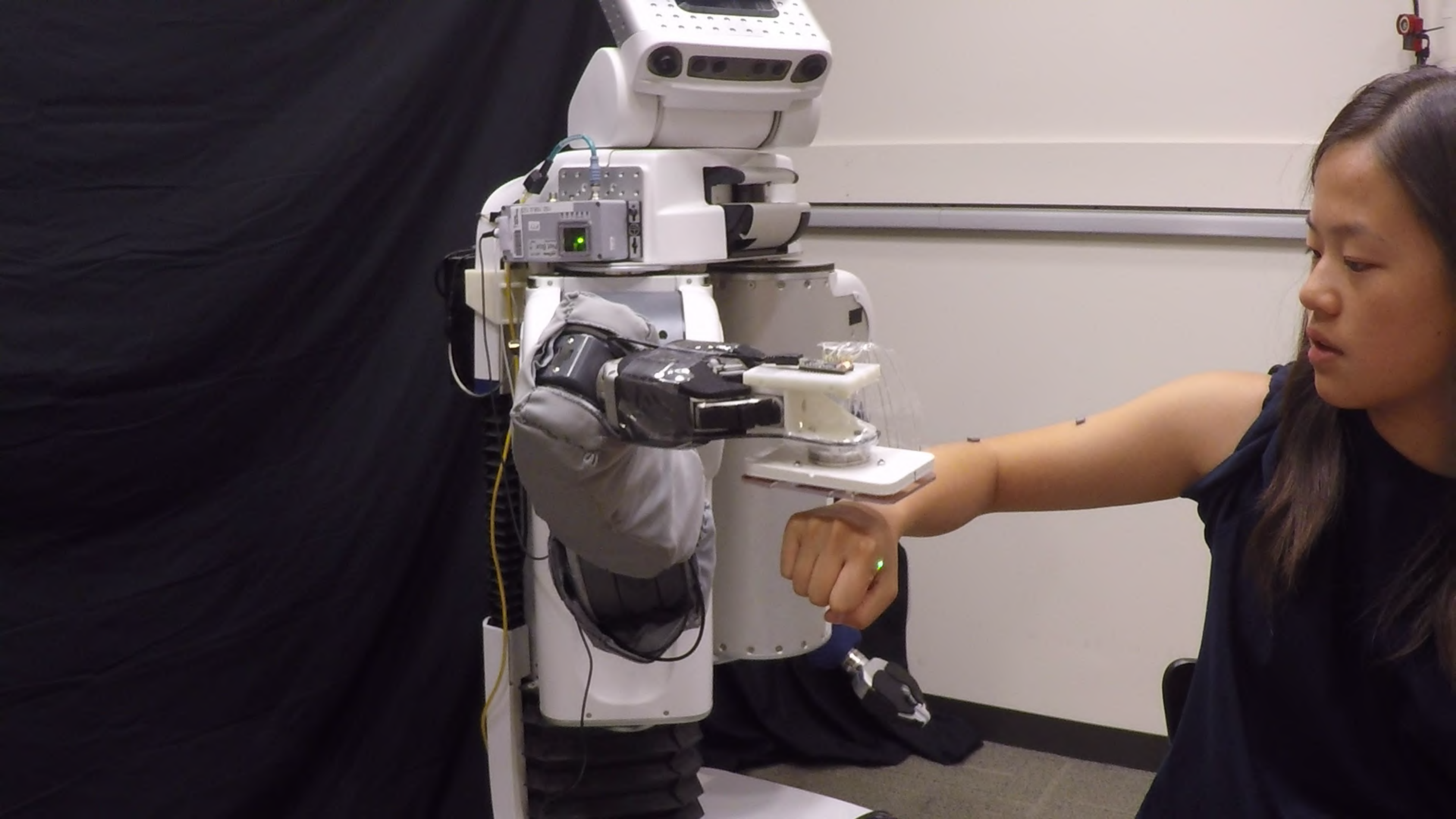}
\includegraphics[width=0.24\textwidth, trim={15cm 8cm 6cm 1cm}, clip]{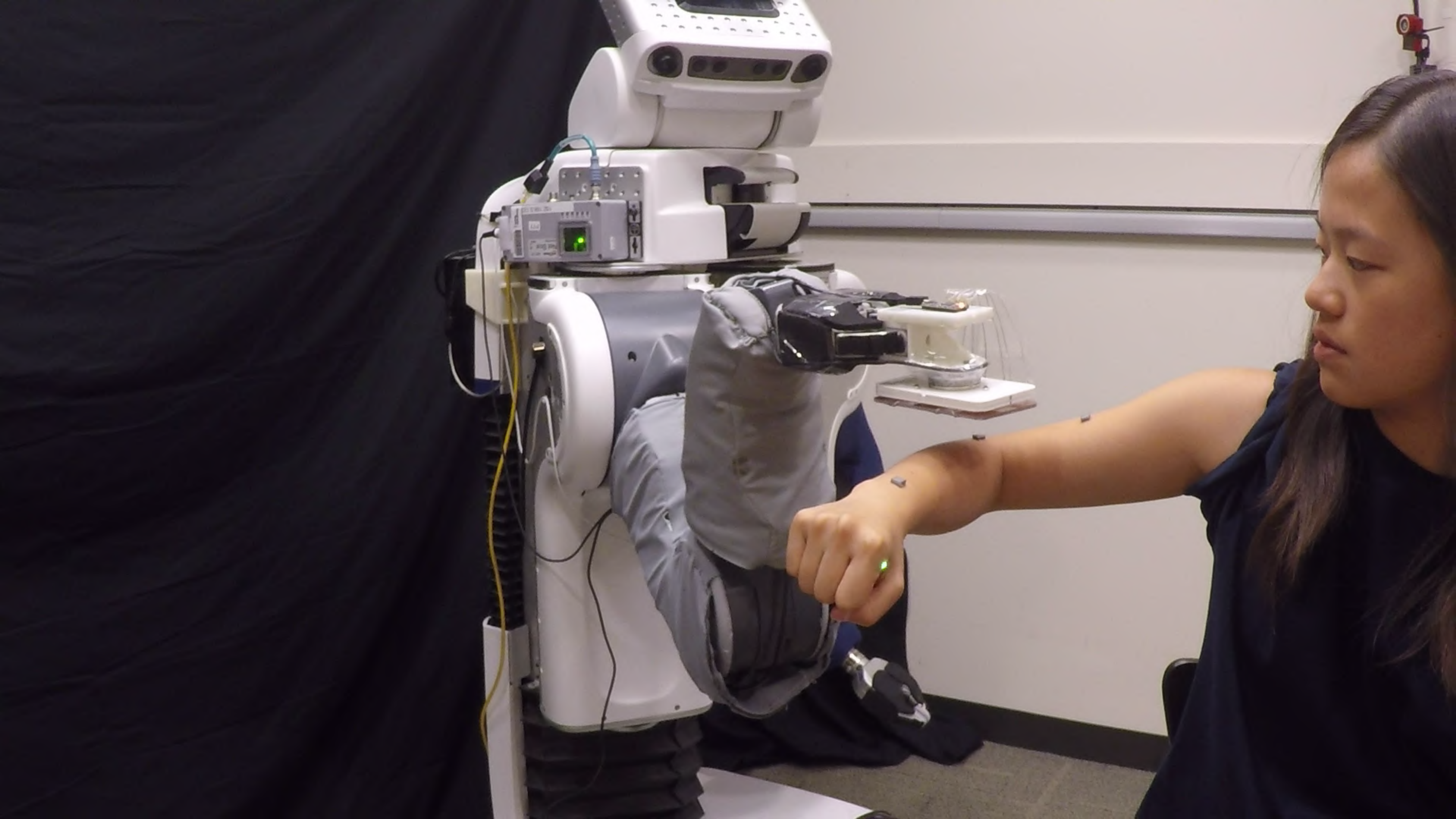}
\includegraphics[width=0.24\textwidth, trim={15cm 8cm 6cm 1cm}, clip]{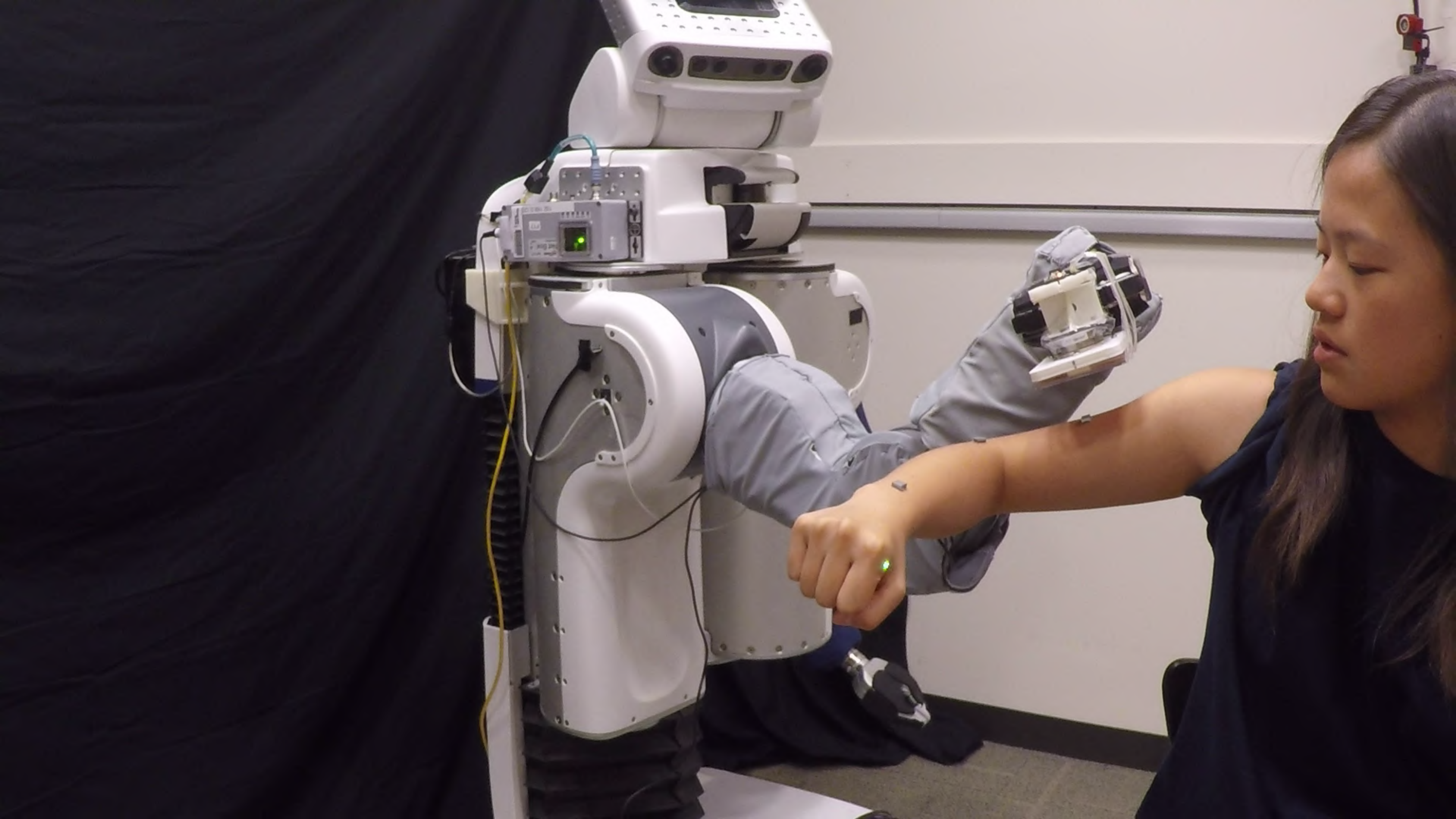}
\includegraphics[width=0.24\textwidth, trim={1cm 17.25cm 8.5cm 3.25cm}, clip]{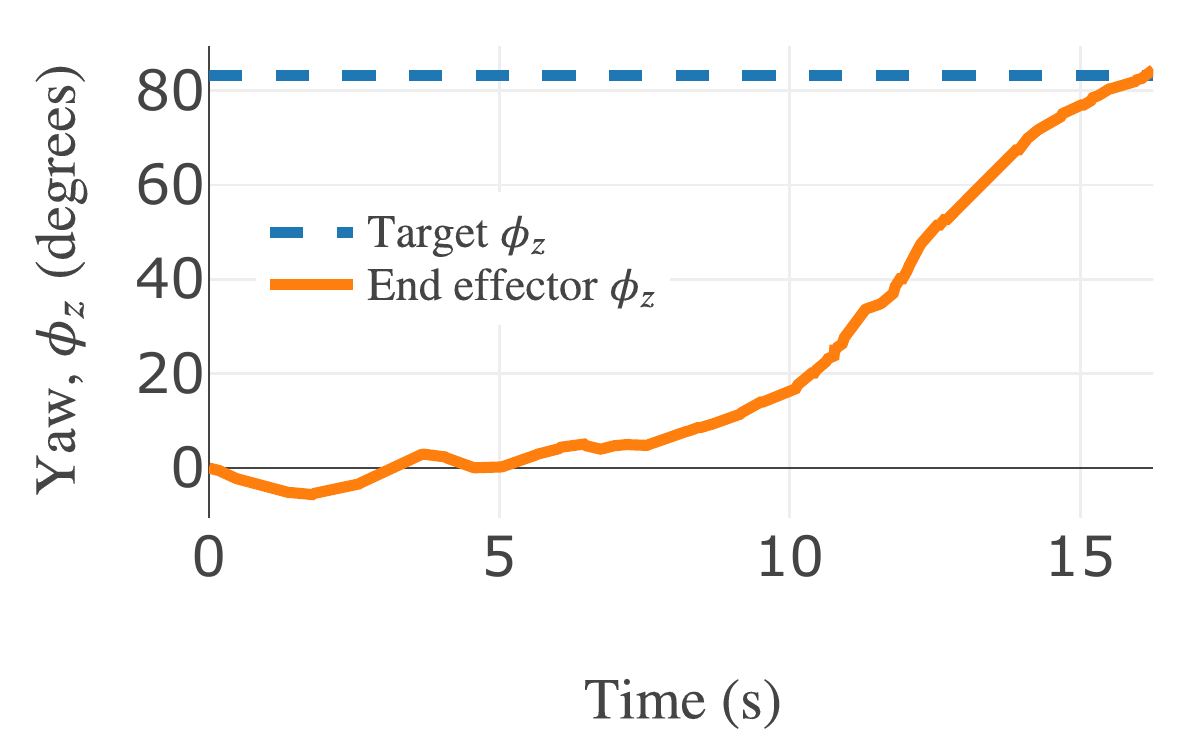}
\caption{\label{fig:elbow}The PR2 used capacitive servoing to move its end effector proximally from the participant's hand to shoulder. The participant bent their elbow at an $\sim$90 degree angle. The right plot displays the yaw orientation $\phi^{}_z$ of the capacitive sensor over time (determined by forward kinematics) and the target yaw orientation of the participant's upper arm (visually estimated using reflective markers on the arm). 3D positions of the infrared reflective markers on a participant's wrist, elbow, and upper arm \emph{were not} provided to the robot during trials.}
\end{figure*}

\begin{figure*}
\centering
\includegraphics[width=0.24\textwidth, trim={15cm 8cm 6cm 1cm}, clip]{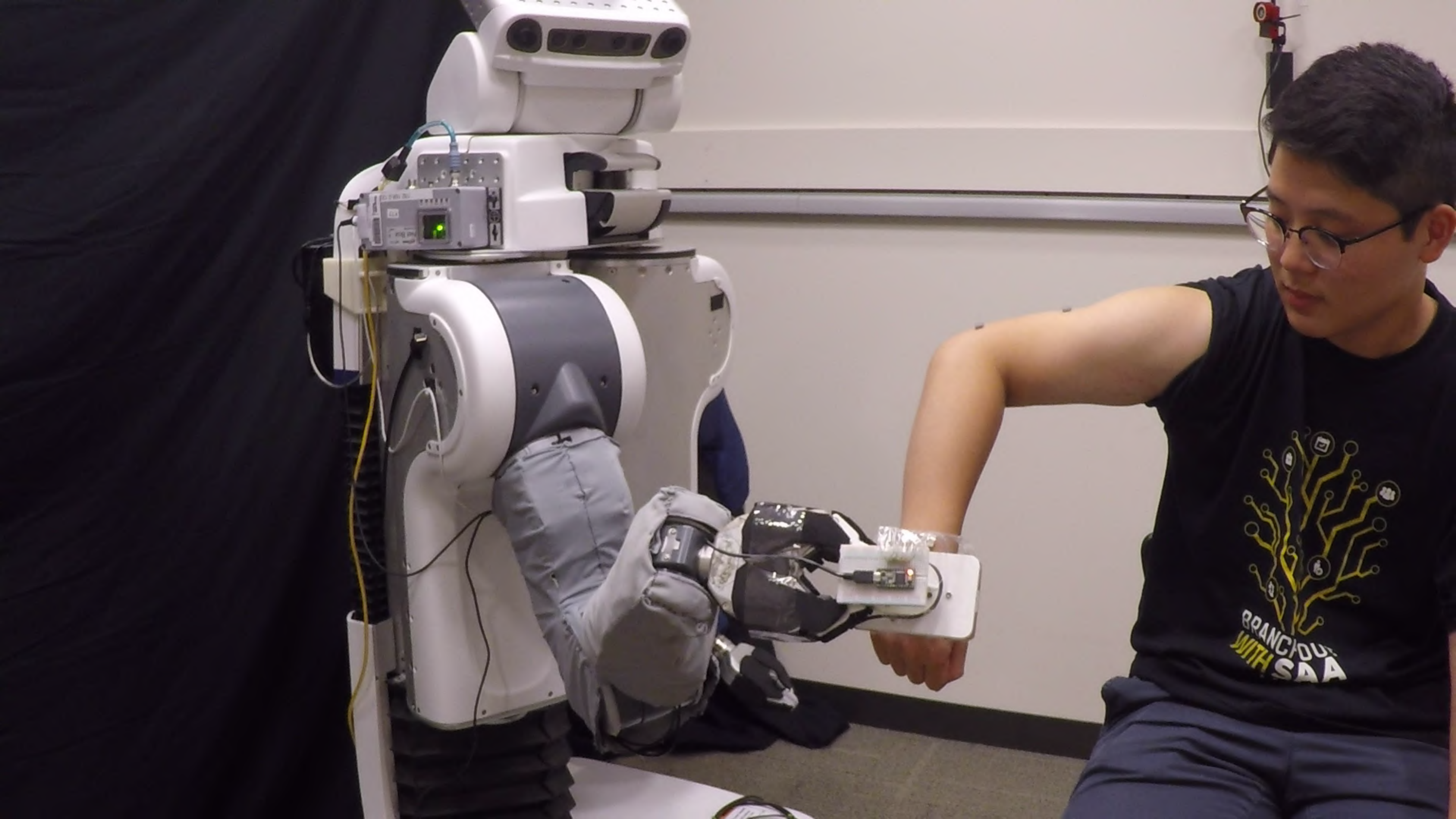}
\includegraphics[width=0.24\textwidth, trim={15cm 8cm 6cm 1cm}, clip]{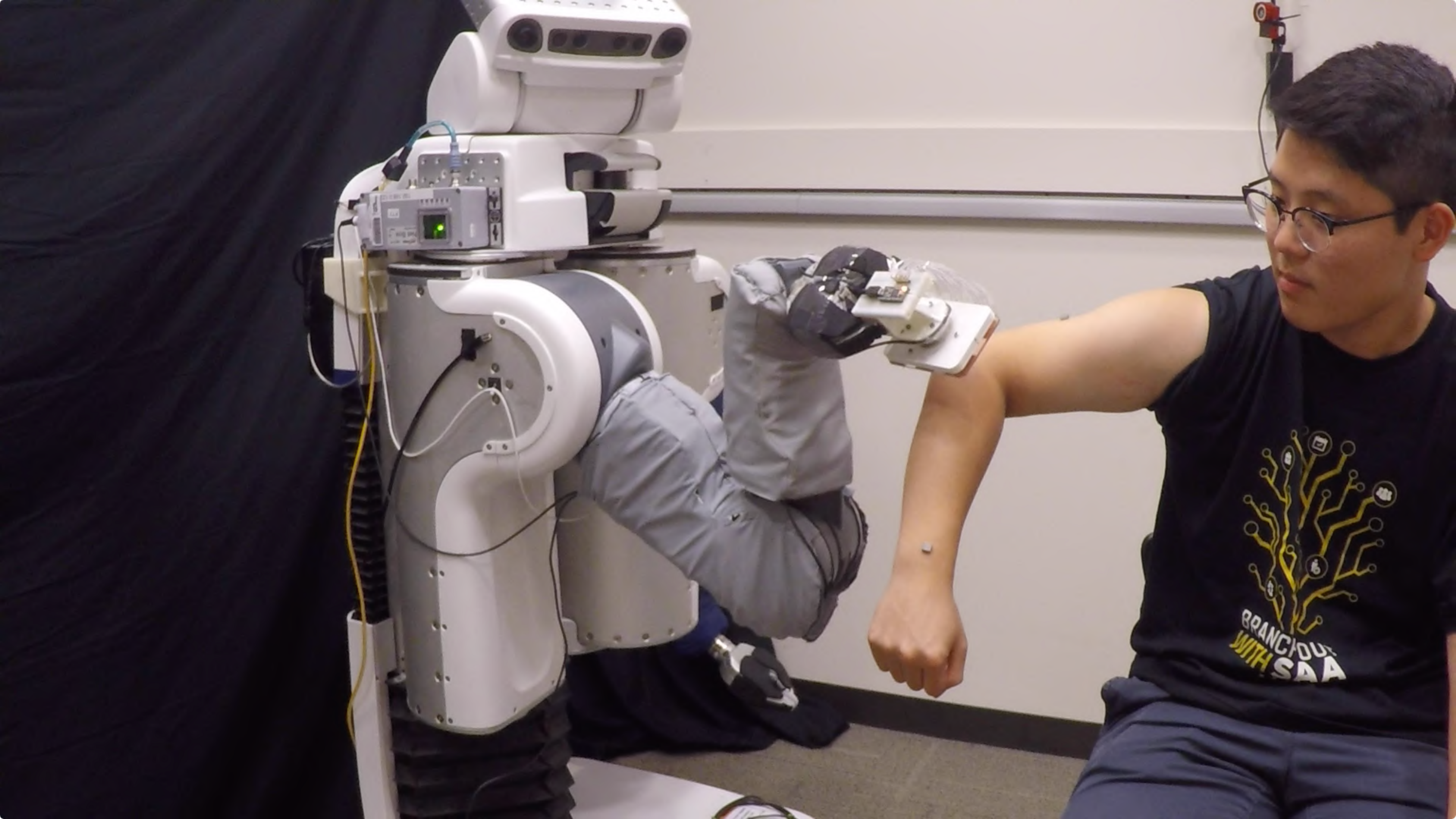}
\includegraphics[width=0.24\textwidth, trim={15cm 8cm 6cm 1cm}, clip]{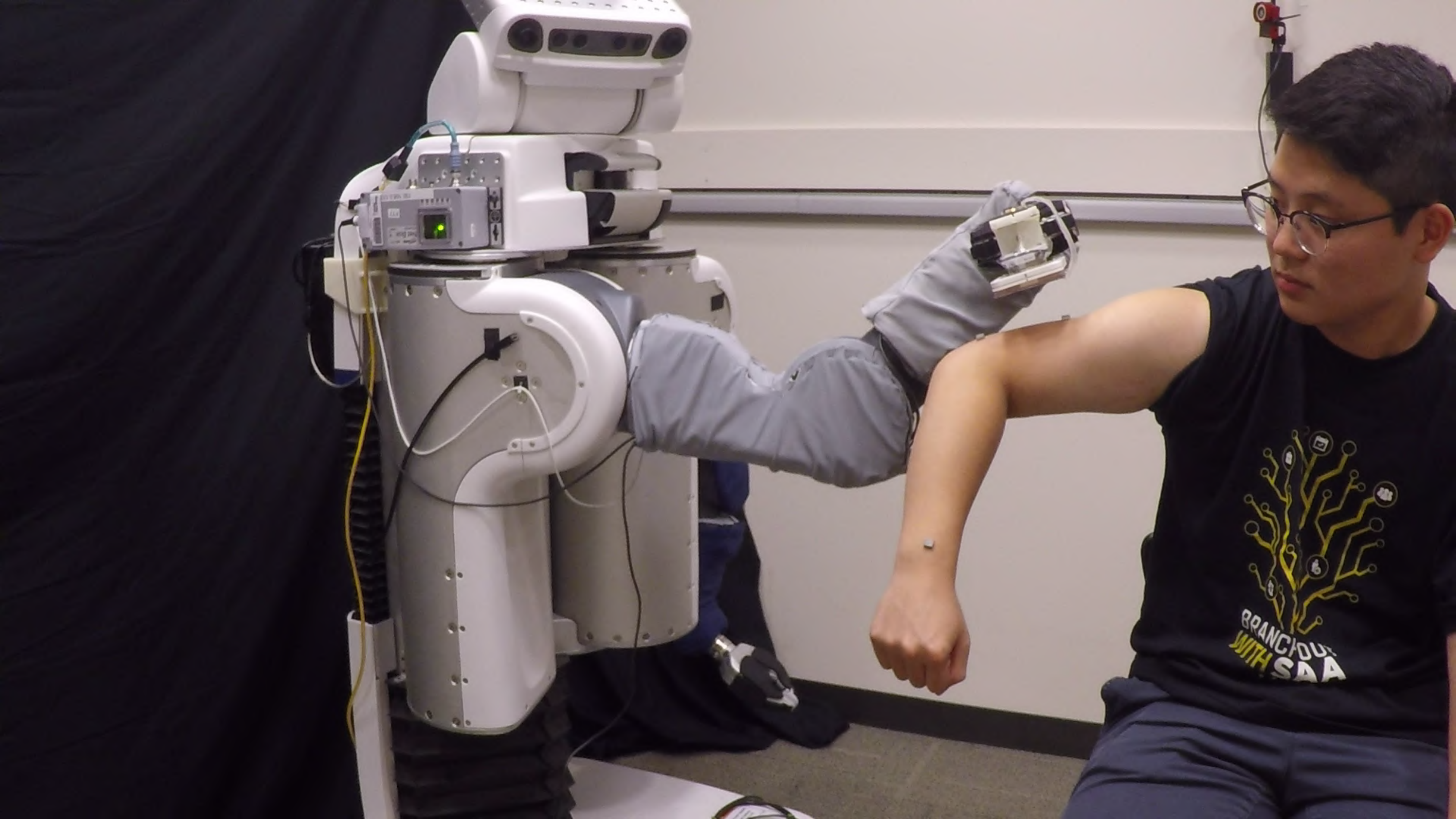}
\includegraphics[width=0.24\textwidth, trim={1cm 17.25cm 8.5cm 2.5cm}, clip]{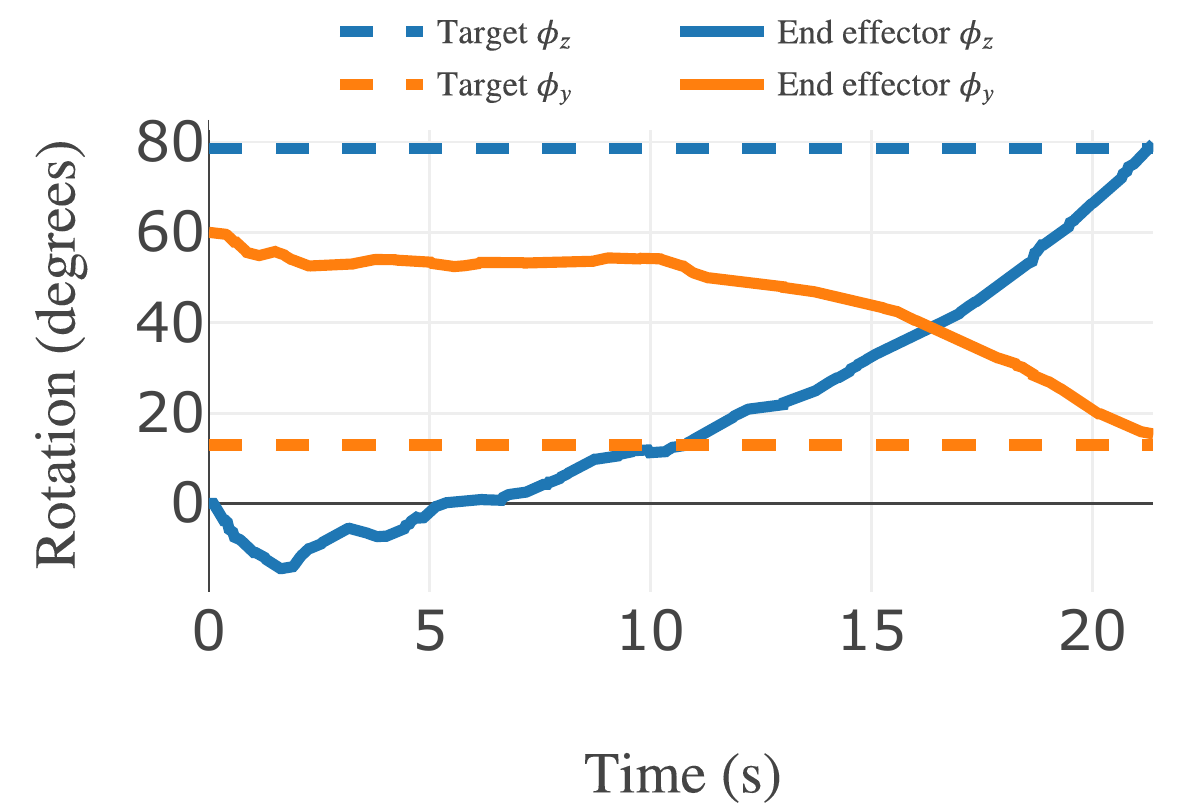}
\caption{\label{fig:shoulder}The PR2 used capacitive servoing to move proximally to a participant's shoulder around a tilted forearm and bent elbow. The participant bent their elbow at a right angle and tilted their forearm $\sim$60 degrees towards the ground. The plot displays the yaw orientation $\phi^{}_z$ and pitch orientation $\phi^{}_y$ (tilt) of both the robot's end effector and participant's upper arm (target).}
\end{figure*}

\begin{figure*}
\centering
\includegraphics[width=0.24\textwidth, trim={15cm 8cm 6cm 1cm}, clip]{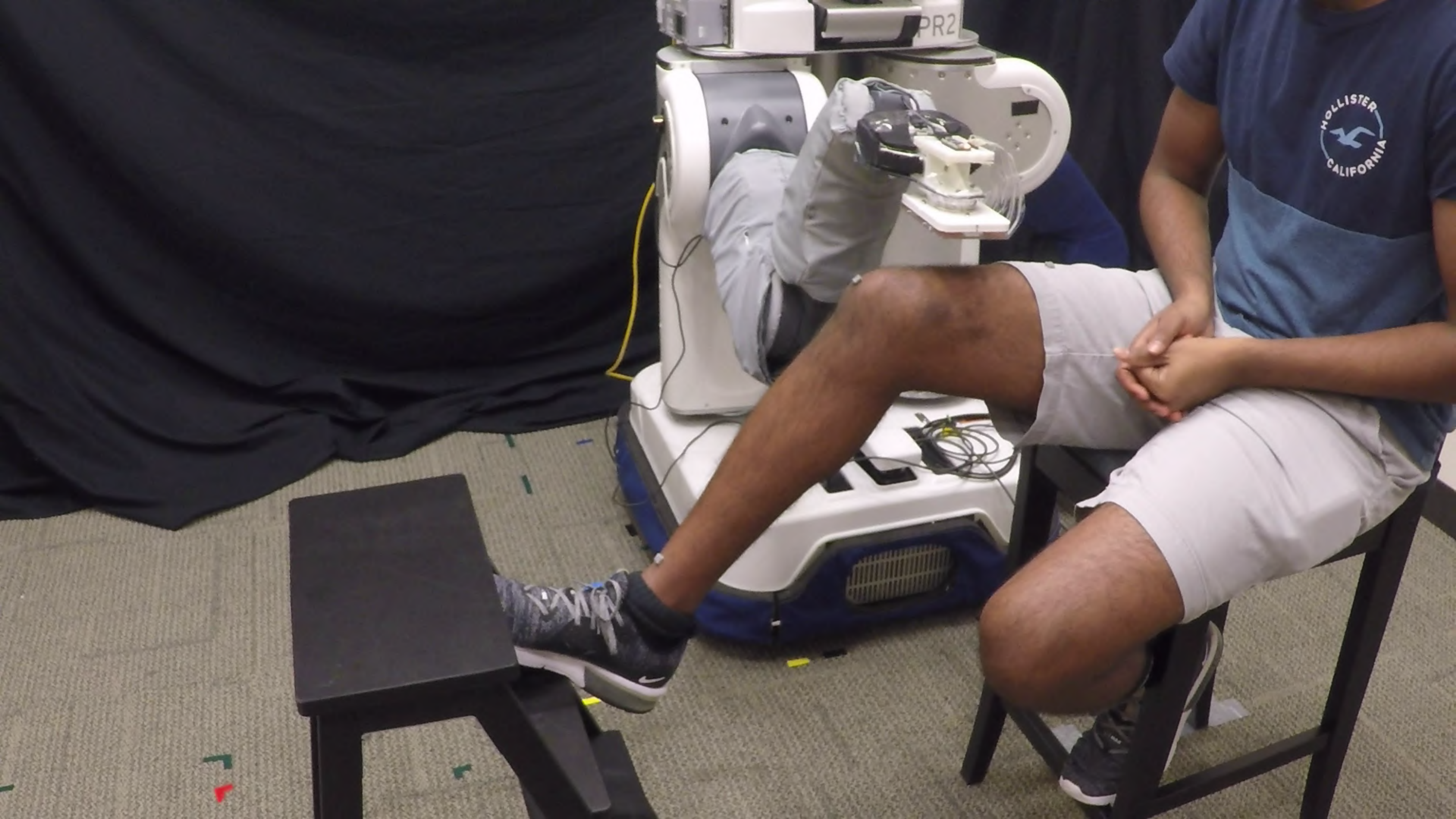}
\includegraphics[width=0.24\textwidth, trim={15cm 8cm 6cm 1cm}, clip]{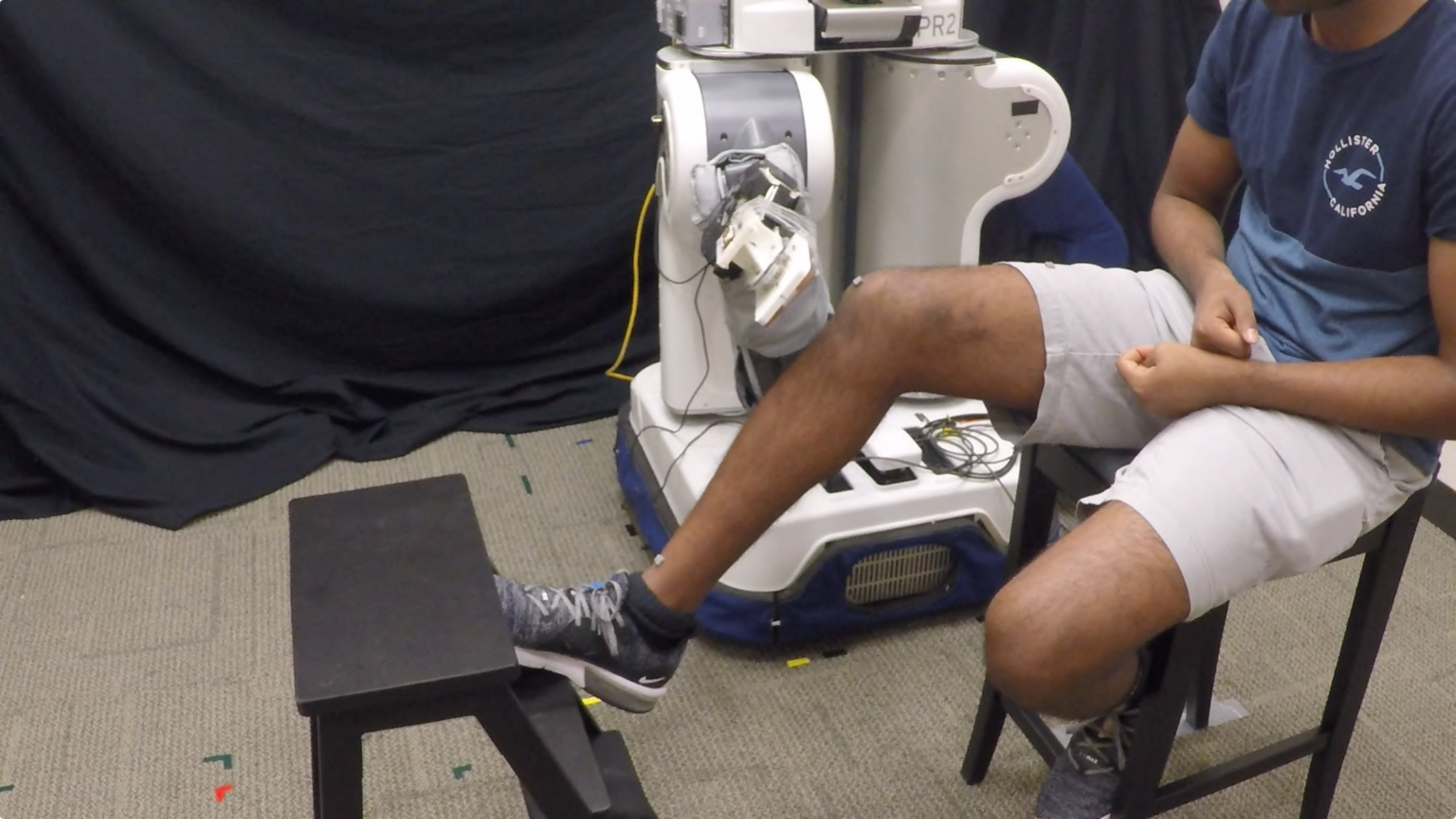}
\includegraphics[width=0.24\textwidth, trim={15cm 8cm 6cm 1cm}, clip]{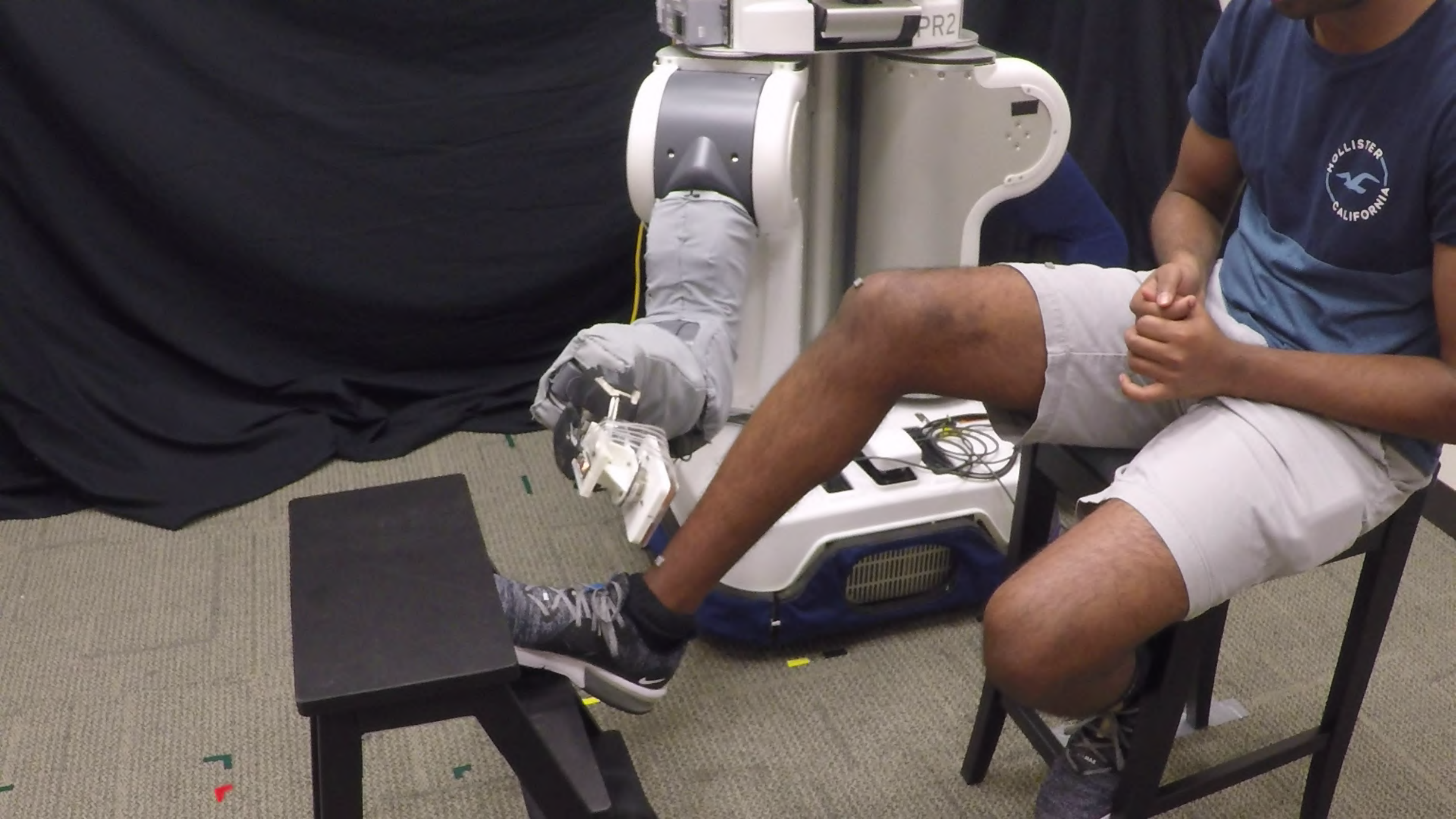}
\includegraphics[width=0.24\textwidth, trim={1cm 17.25cm 8.5cm 3.25cm}, clip]{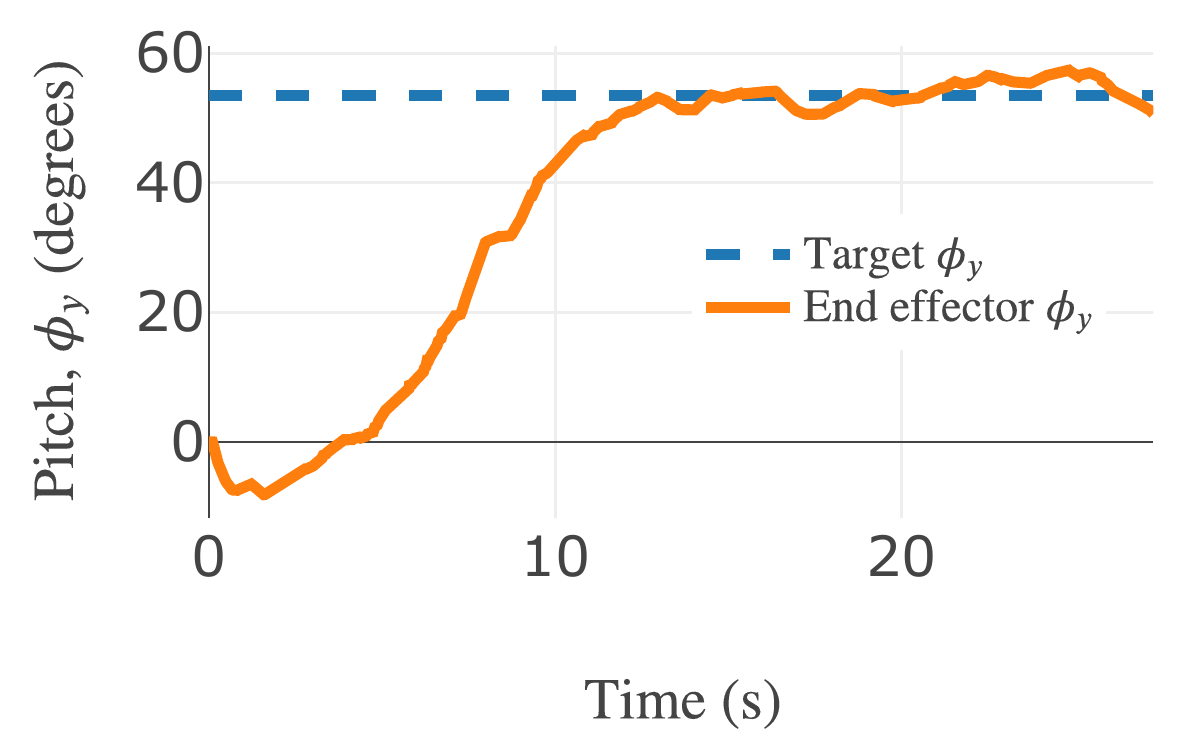}
\caption{\label{fig:knee}The PR2 used capacitive servoing to control its end effector distally from a participant's thigh to ankle. The participant bent their knee $\sim$60 degrees by resting their foot on a stool. The plot shows the pitch orientation $\phi^{}_y$ of the robot's end effector and the participant's shin (target) over time.}
\end{figure*}

We observed similar pose estimation performance between a model trained on one participant (Table~\ref{table:generalization}) and models trained on multiple participants (Table~\ref{table:lopo}). With less than 0.5~cm difference in translational $(D^{}_y, D^{}_z)$ error (difference between Table~\ref{table:generalization} and Table~\ref{table:lopo}) and less than 1$^{\circ}$ difference in rotational $(\theta^{}_y, \theta^{}_z)$ error, these results indicate that pose estimation models trained on capacitance measurements from one participant are likely to generalize to a larger population of people. 

Another question that arises is how well estimation via capacitive sensing performs across both small and large sized limbs.
Table~\ref{table:minmax_errors} provides the pose estimation error of our model for individual participants with either the smallest or largest circumference for a given limb. Error values in Table~\ref{table:minmax_errors} are averaged over all $N=50$ data collection trials for a single participant and limb location. Limb circumferences ranged from 13~cm (wrist of the smallest participant) to 40~cm (shin of the largest participant).

Despite not providing the pose estimation model with explicit information about limb circumference, results were relatively stable across the various limb sizes. Average error in lateral translation $D^{}_y$ ranged from 1.6~cm to 3~cm, whereas estimation error in vertical translation $D^{}_z$ ranged from 1.4~cm to 2.4~cm. Error in pitch orientation estimation $\theta^{}_y$ ranged from 6.0$^{\circ}$ to 12.3$^{\circ}$ and yaw orientation error $\theta^{}_z$ varied from 8.6$^{\circ}$ to 12.4$^{\circ}$.


\subsection{Capacitive Servoing Around Human Limbs}
\label{sec:servo_around_limbs}

One strength of capacitive sensing is its ability to sense the contours along a human limb. For example, the process of sensing and moving a robot's end effector along a person's limb is the fundamental task required for several caregiving scenarios, including dressing and bathing assistance. In this section, we evaluate how a robot can use multidimensional capacitive servoing (Algorithm~\ref{alg:control}) to navigate around the contours of human limbs.

We design three tasks in which the robot must navigate around a static human limb at various poses: (1) \emph{Bent Elbow}, participant arm is held parallel to the ground with a bent elbow (Fig.~\ref{fig:elbow}), (2) \emph{Forearm Tilt}, forearm and hand are tilted towards the ground while the elbow is bent (Fig.~\ref{fig:shoulder}), (3) \emph{Bent Knee}, participant bends their leg while sitting on a chair (Fig.~\ref{fig:knee}).

In the first task, each participant held his or her arm in a static pose parallel to the ground with their elbow bent at either 0, 30, 60, 90, or 120 degrees. The robot positioned the capacitive sensor starting 5~cm above a participant's hand and used capacitive servoing to navigate the end effector and sensor around the bent elbow and up to a participant's shoulder. Fig.~\ref{fig:elbow} depicts a representative image sequence of the robot using capacitive servoing to follow the contours of a participant's arm while they hold a 90 degree elbow bend. The plot in Fig.~\ref{fig:elbow} shows the yaw orientation $\phi^{}_z$ of the capacitive sensor over time (determined by the robot's forward kinematics) and the target yaw orientation of the participant's upper arm (determined by visually reflective markers on the arm). We observe that the robot was able to sense and rotate around the bent elbow to match the target orientation of the participant's upper arm by the end of the trial.

\begin{figure*}
\centering
\includegraphics[width=0.24\textwidth, trim={15cm 5cm 10cm 6cm}, clip]{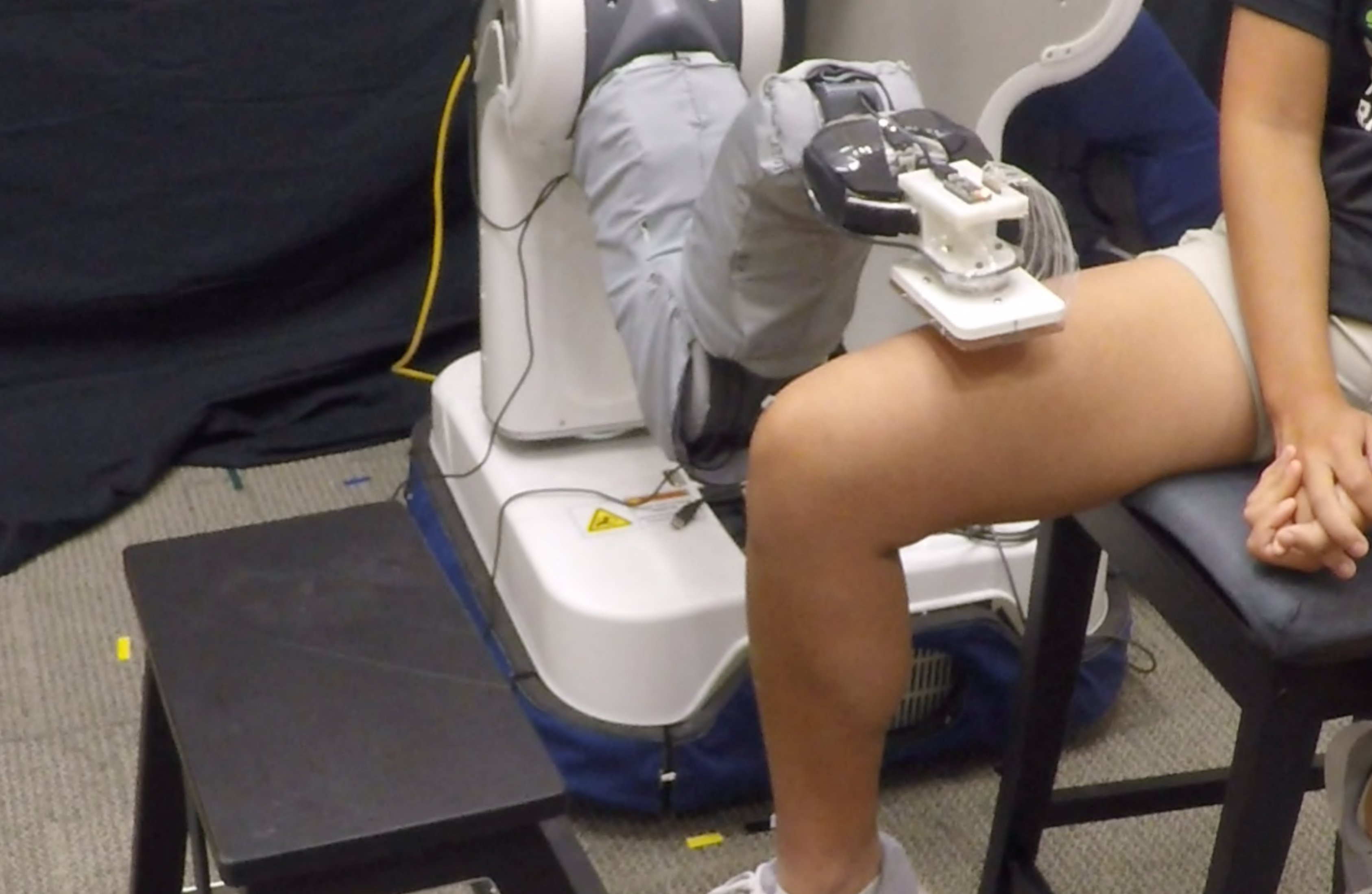}
\includegraphics[width=0.24\textwidth, trim={15cm 5cm 10cm 6cm}, clip]{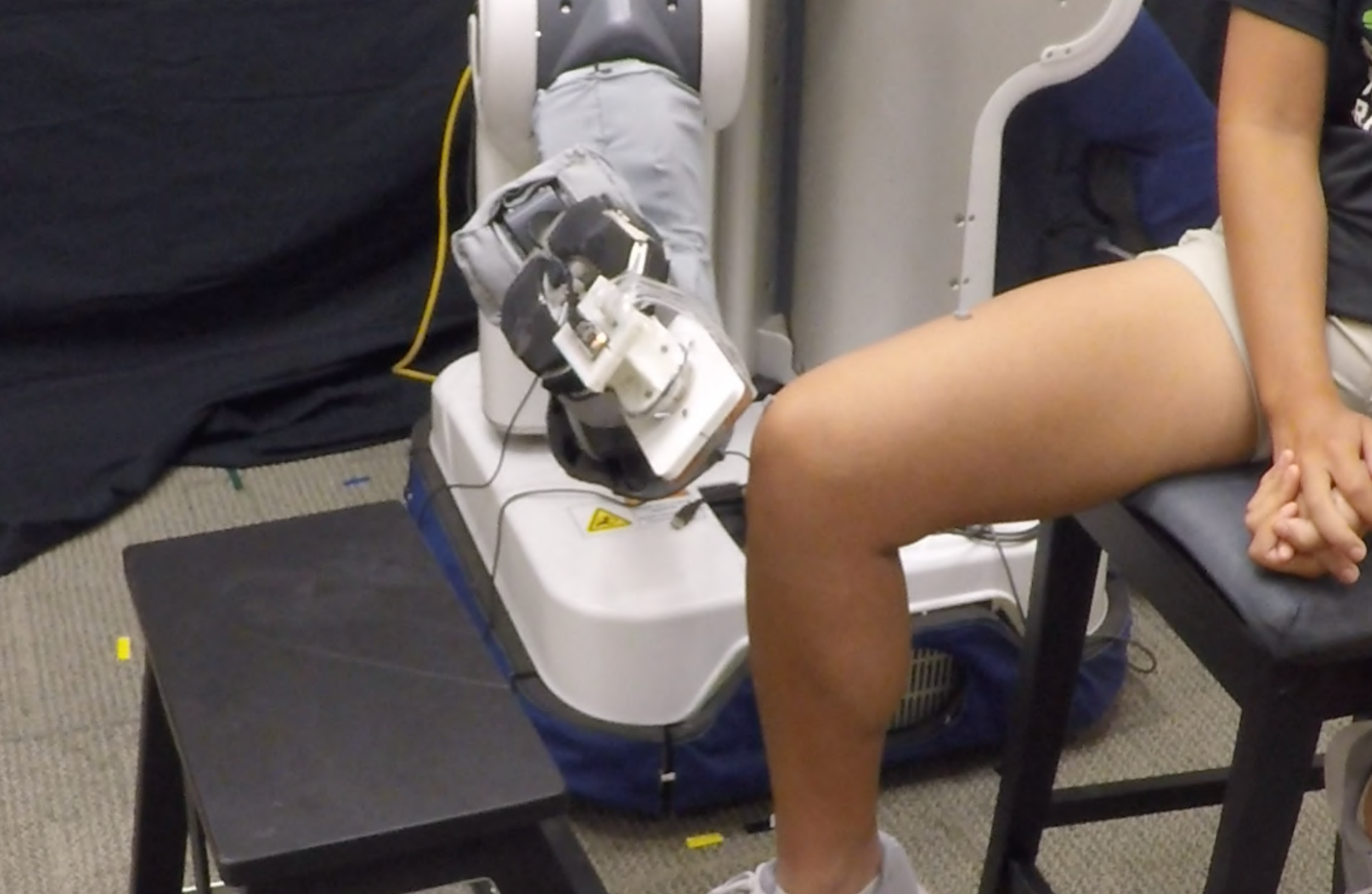}
\includegraphics[width=0.24\textwidth, trim={15cm 5cm 10cm 6cm}, clip]{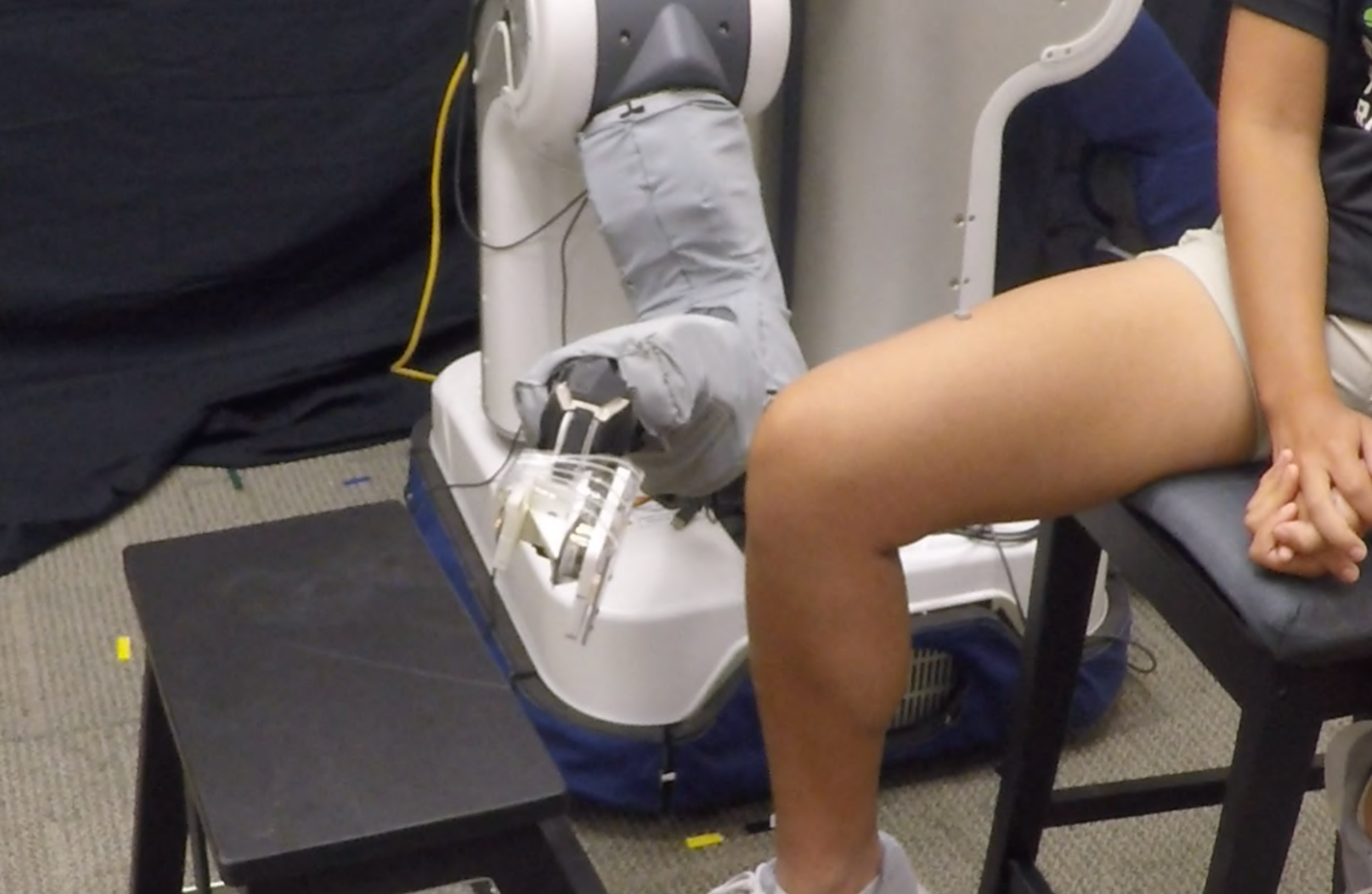}
\includegraphics[width=0.24\textwidth, trim={15cm 5cm 10cm 6cm}, clip]{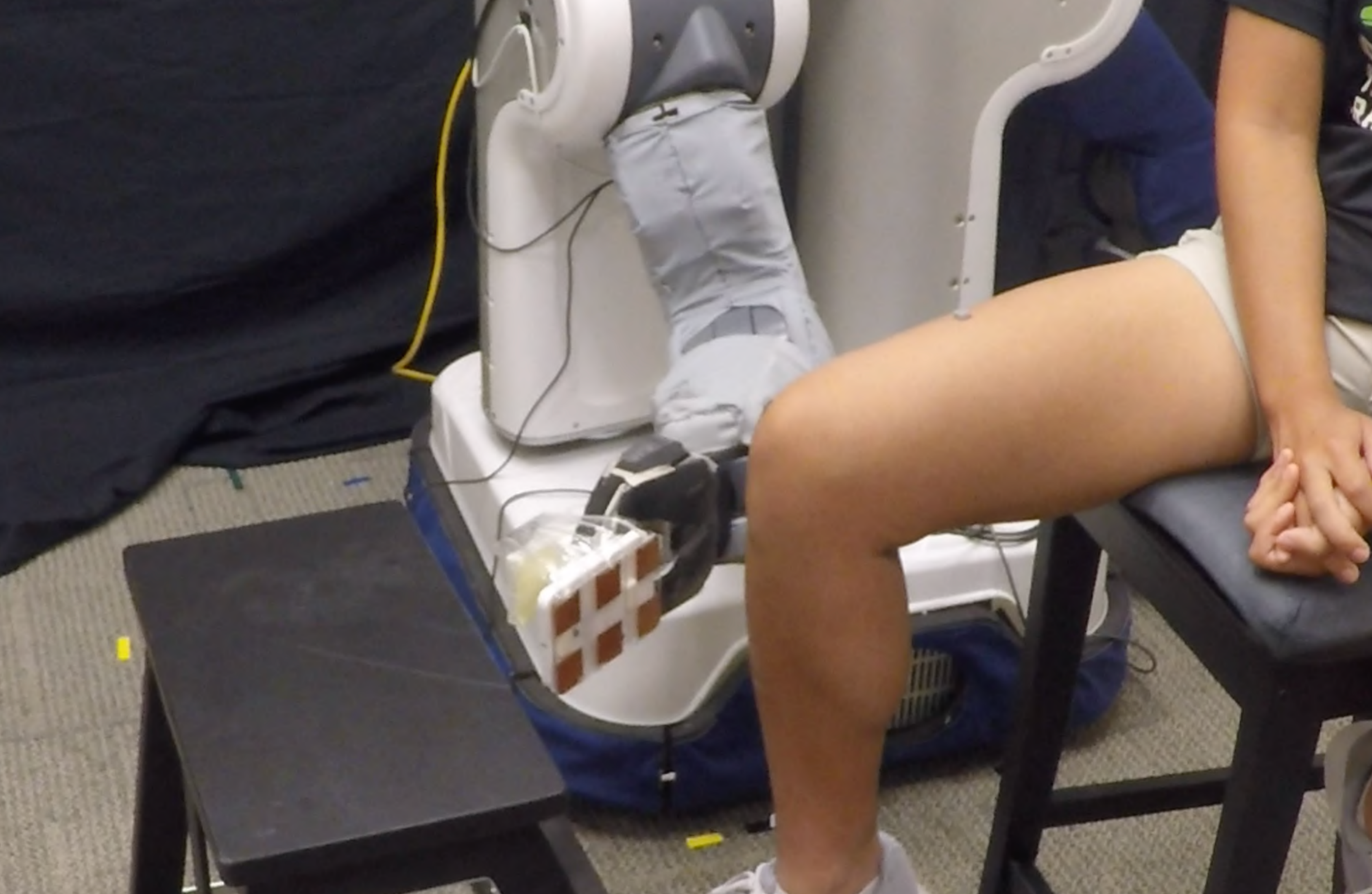}
\caption{\label{fig:knee_failure}An example in which capacitive servoing lost track of a participant's leg when attempting to rotate around a 90 degree bent knee.}
\end{figure*}

\begin{figure*}
\centering
\includegraphics[width=0.3\textwidth, trim={1cm 17.5cm 8.5cm 2cm}, clip]{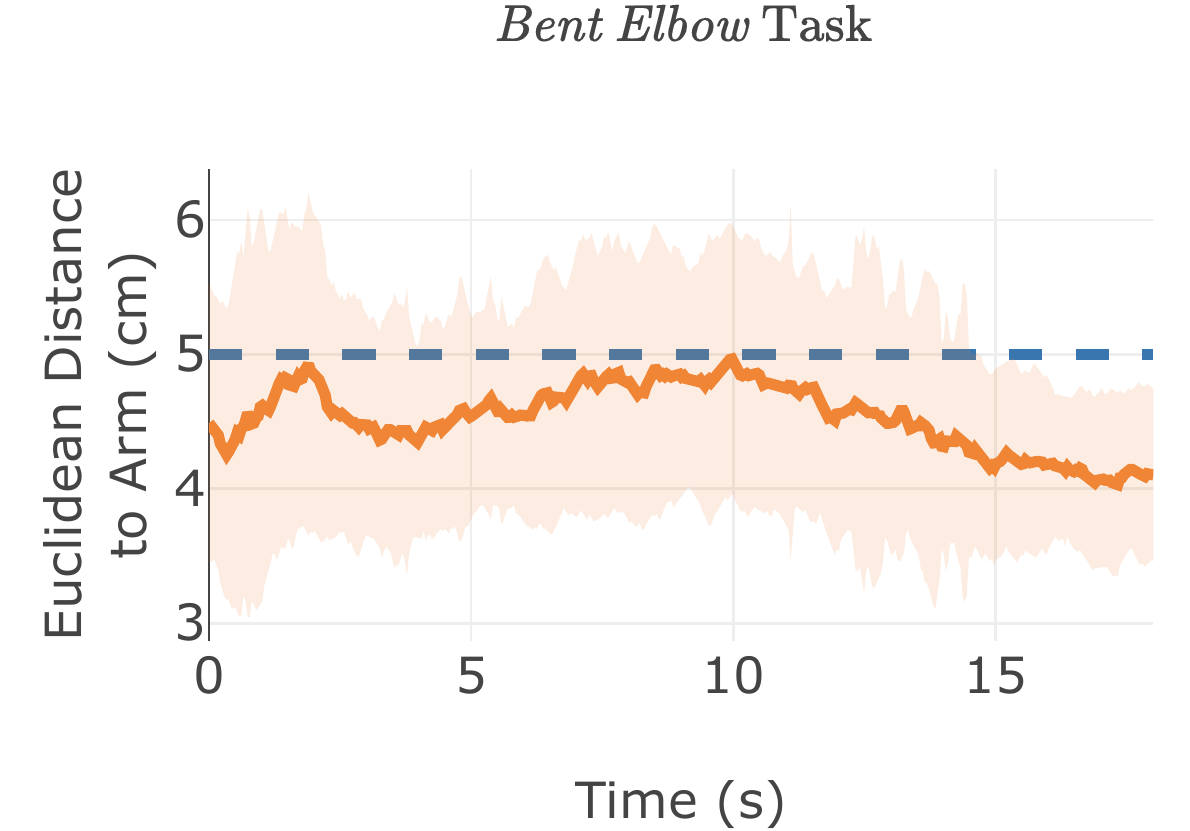}
\includegraphics[width=0.3\textwidth, trim={1cm 17.5cm 8.5cm 2cm}, clip]{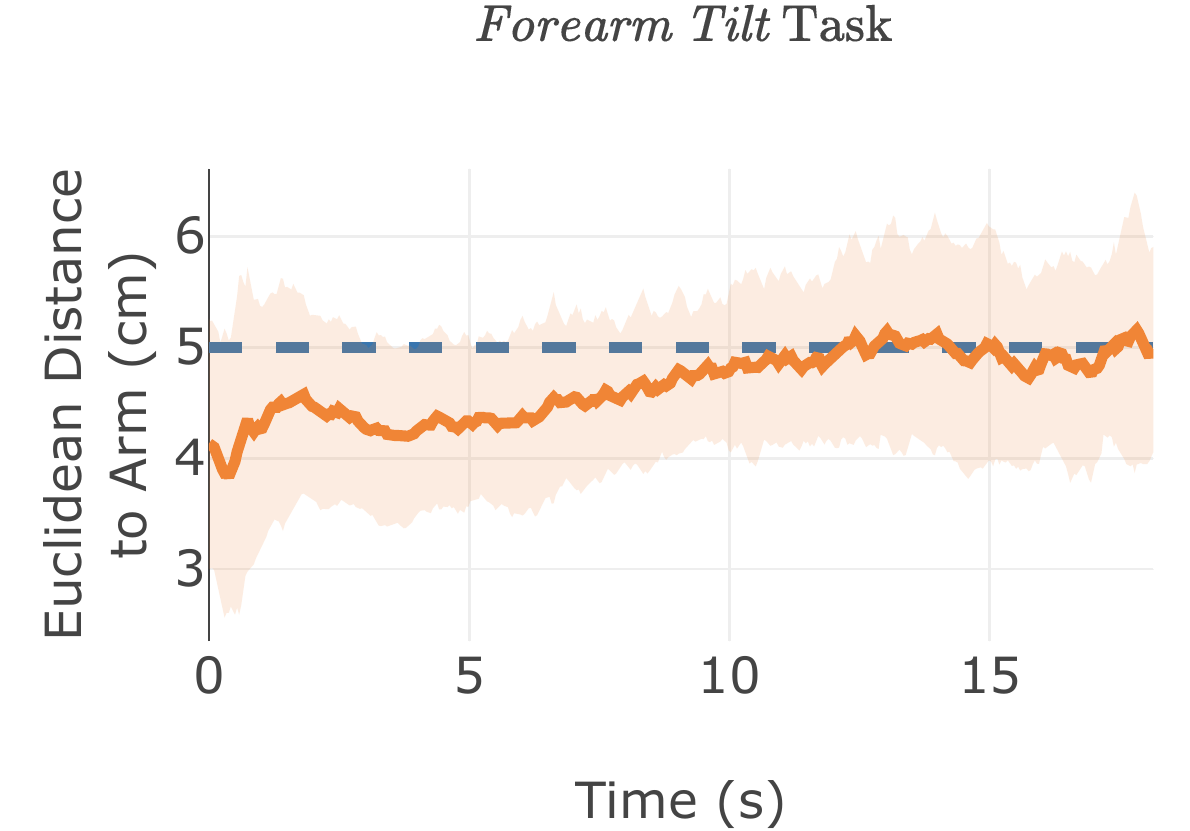}
\includegraphics[width=0.36\textwidth, trim={1cm 17.5cm 6.5cm 2cm}, clip]{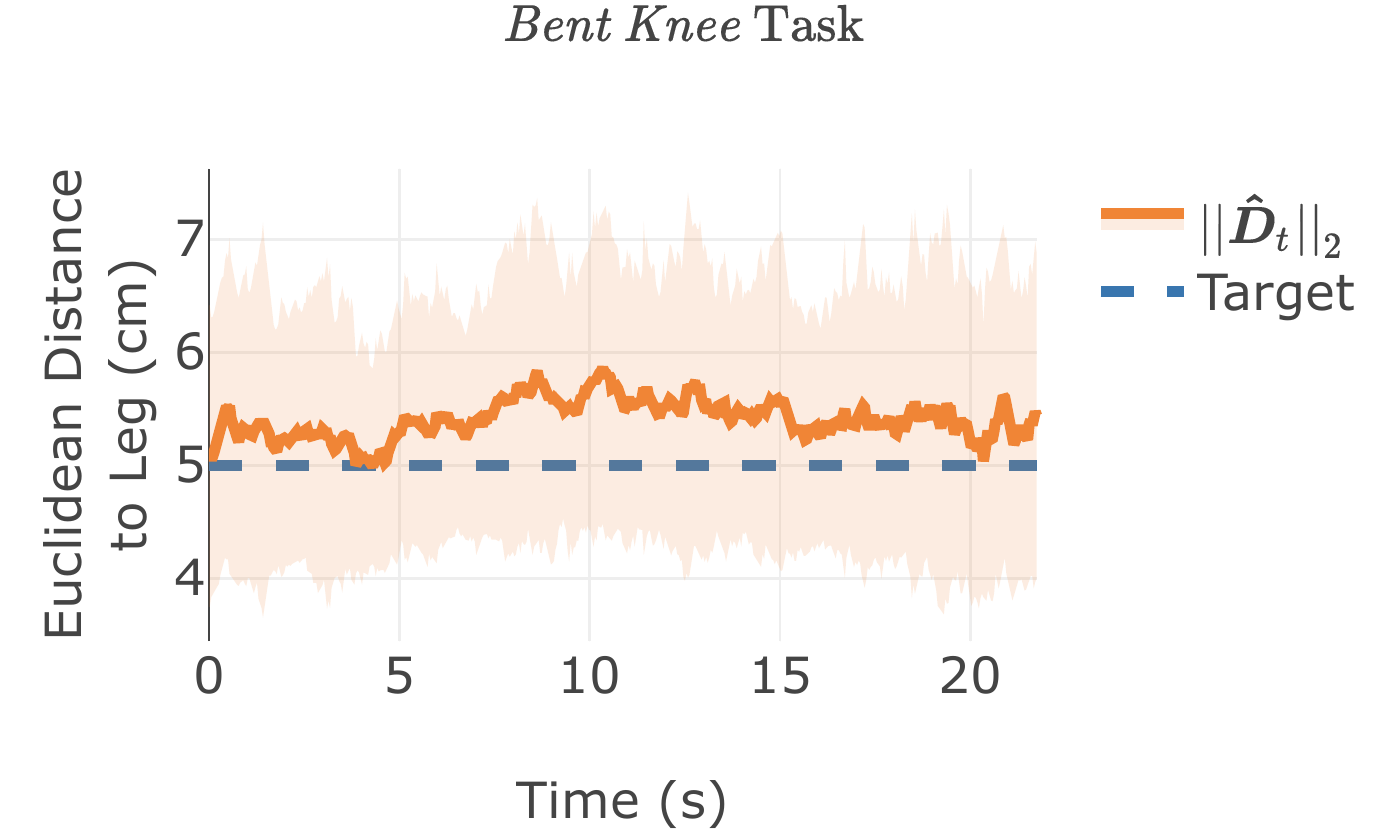}
\caption{\label{fig:estimated_distances}Estimated distance $||\bm{\hat{D}}^{}_t||^{}_2$ between the capacitive sensor and human limb, averaged over all participant trials and joint rotations for each task. During each trial, the robot is instructed to remain 5~cm above the participant's limb given $\bm{p}^{}_{desired} = (0, 5, 0, 0)$ (see Section~\ref{sec:control}).}
\end{figure*}

For the second task, each participant held their arm parallel to the ground with a 90 degree elbow bend, and then tilted their forearm and hand towards the ground at degree intervals of 0, 30, 60, and 90 degrees. Fig.~\ref{fig:shoulder} shows an image sequence for capacitive servoing around a participant's arm as their forearm is tilted $\sim$60 degrees towards the ground. The corresponding plot in Fig.~\ref{fig:shoulder} depicts the yaw $\phi^{}_z$ and pitch $\phi^{}_y$ orientation of the capacitive sensor along with the target orientations determined by the pose of the participant's upper arm. We again observe that the robot's end effector was able to traverse the participant's arm and match the orientation of the upper arm by the end of the capacitive servoing trial.

During the final task, participants sat on a chair and rested their foot on a stool such that their knee was bent at 0, 30, 60, or 90 degrees. The robot began above a participant's thigh and then used capacitive servoing to navigate down towards their ankle. A sequence of this task is shown in Fig.~\ref{fig:knee} where the participant had an approximately 60 degree bend at their knee. The associated plot shows the pitch $\phi^{}_y$ of the sensor as it rotates around the knee to match the target orientation of the shin (visually estimated by reflective markers on the knee and ankle).

\begin{table}
\centering
\caption{\label{table:accuracies}Task success averaged over all trials.}
\begin{tabular}{cccccc} \toprule
    & \multicolumn{5}{c}{Rotation of limb joint (degrees)} \\ \cmidrule{2-6}
    Task & 0 & 30 & 60 & 90 & 120 \\ \midrule\midrule
    \emph{Bent Elbow} & 100\% & 100\% & 100\% & 100\% & 91.6\% \\
    \emph{Forearm Tilt} & 100\% & 100\% & 100\% & 100\% & --- \\
    \emph{Bent Knee} & 100\% & 100\% & 100\% & 62.5\% & --- \\
	\bottomrule
\end{tabular}
\end{table}

\begin{figure*}
\centering
\includegraphics[width=0.19\textwidth, trim={4cm 2cm 4cm 0cm}, clip]{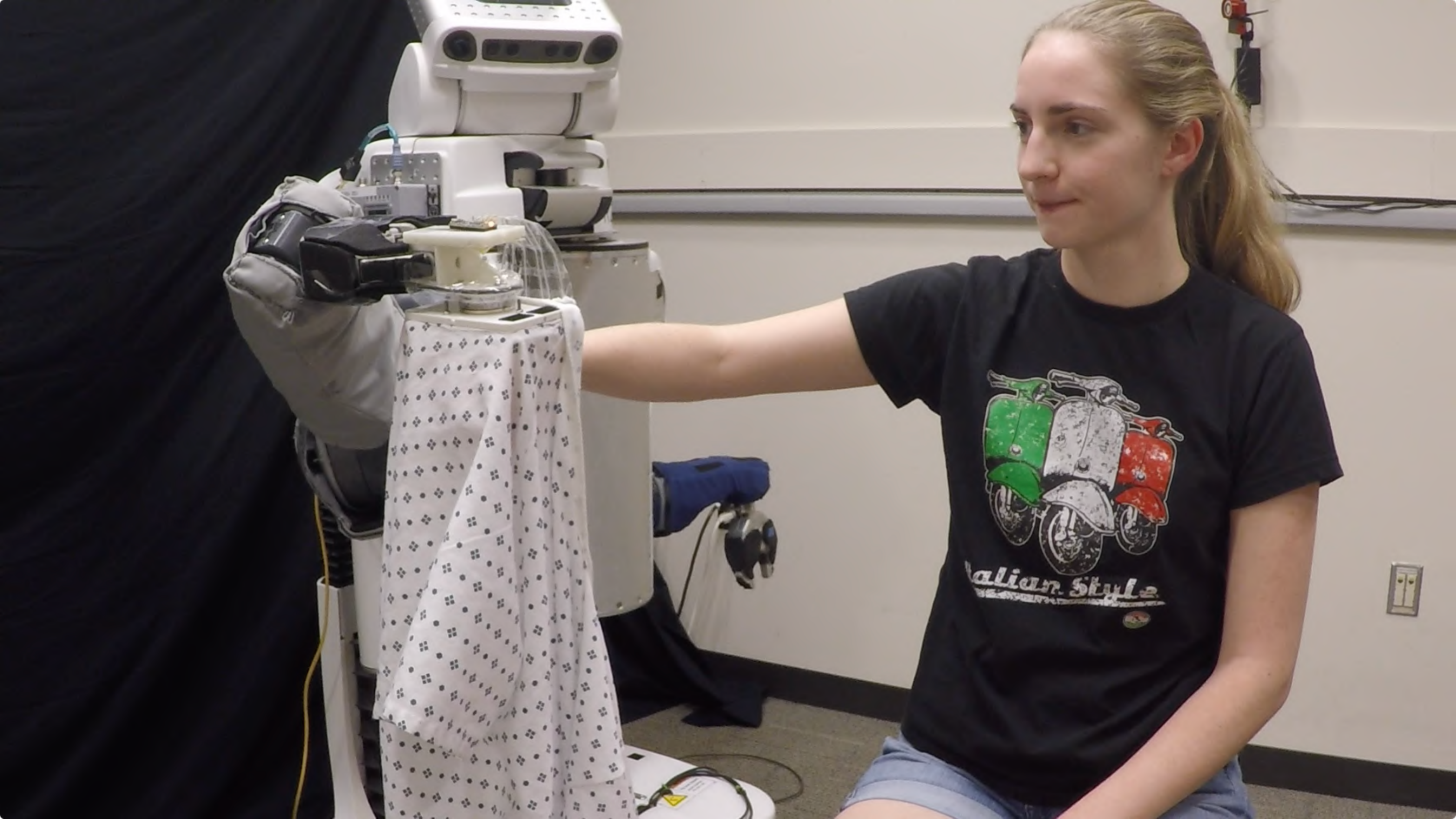}
\includegraphics[width=0.19\textwidth, trim={4cm 2cm 4cm 0cm}, clip]{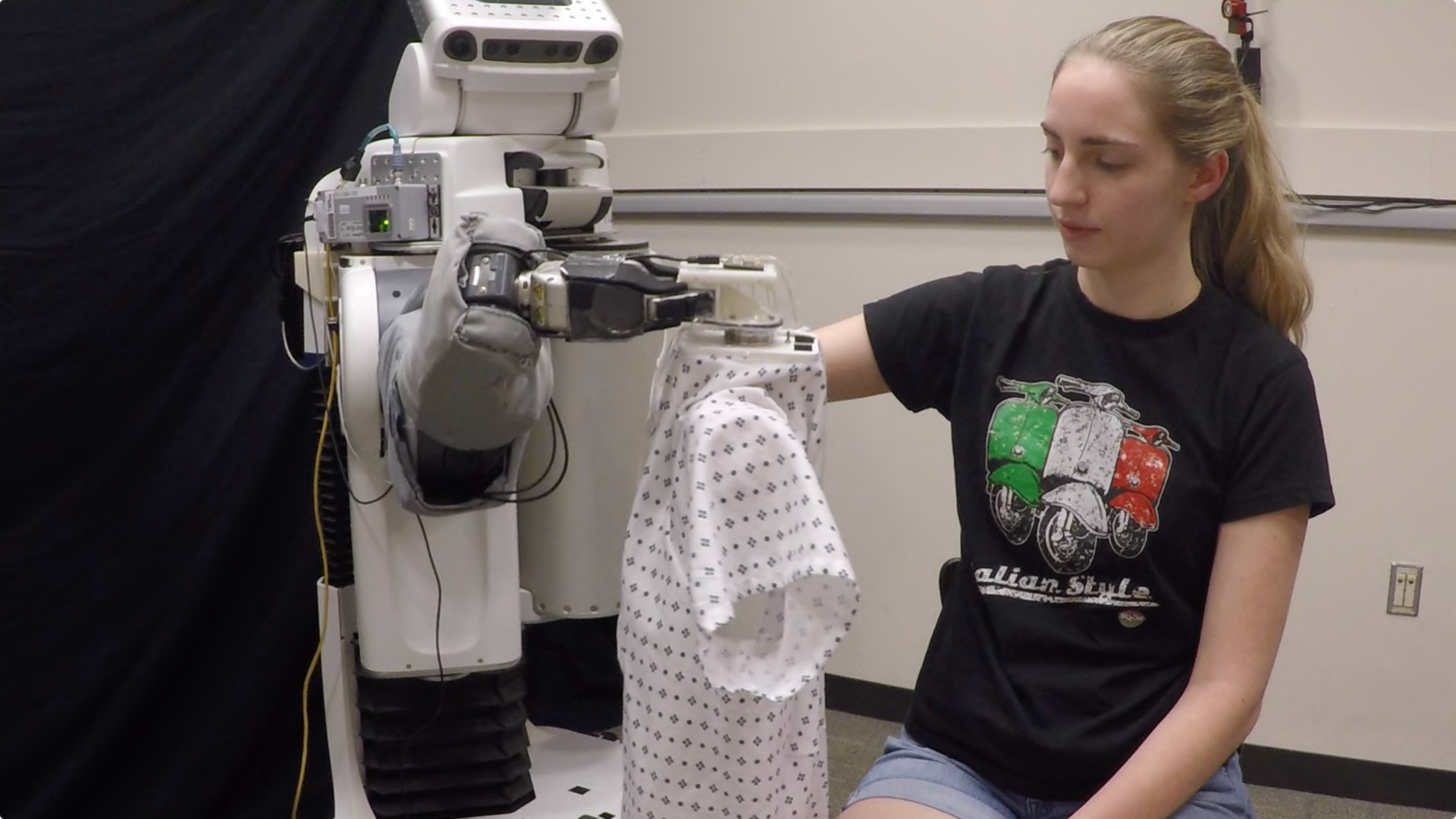}
\includegraphics[width=0.19\textwidth, trim={4cm 2cm 4cm 0cm}, clip]{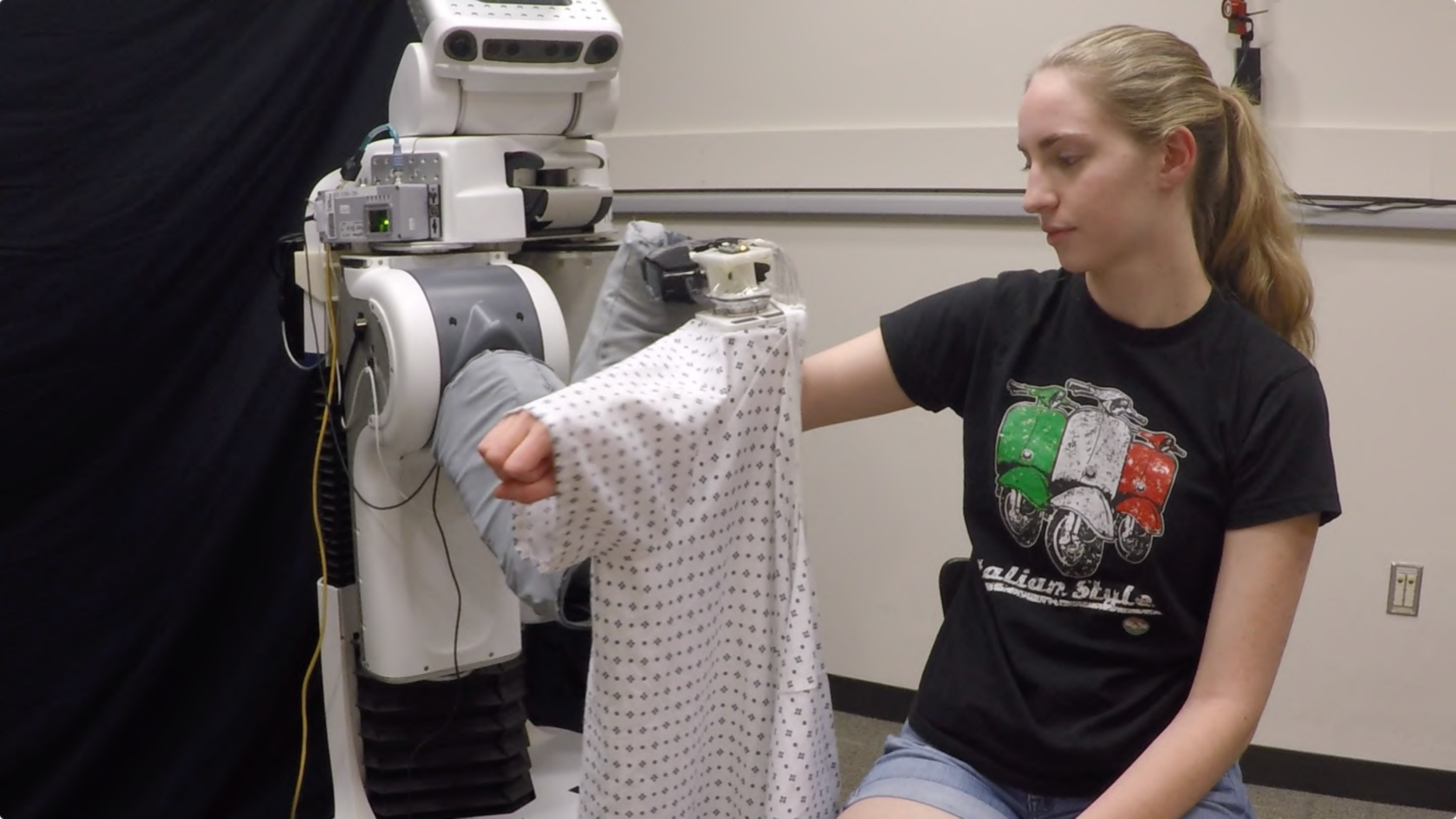}
\includegraphics[width=0.19\textwidth, trim={4cm 2cm 4cm 0cm}, clip]{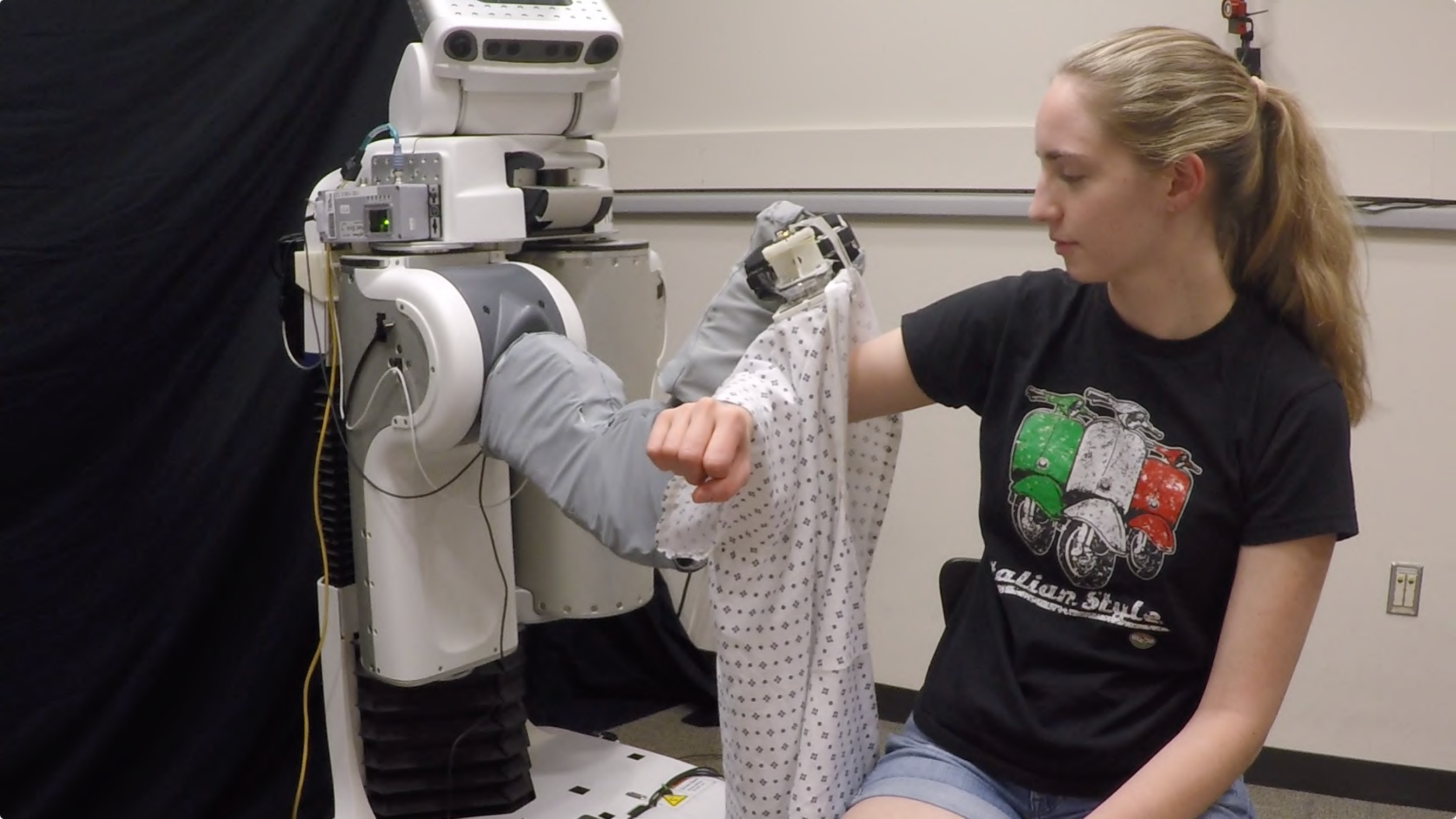}
\includegraphics[width=0.19\textwidth, trim={4cm 2cm 4cm 0cm}, clip]{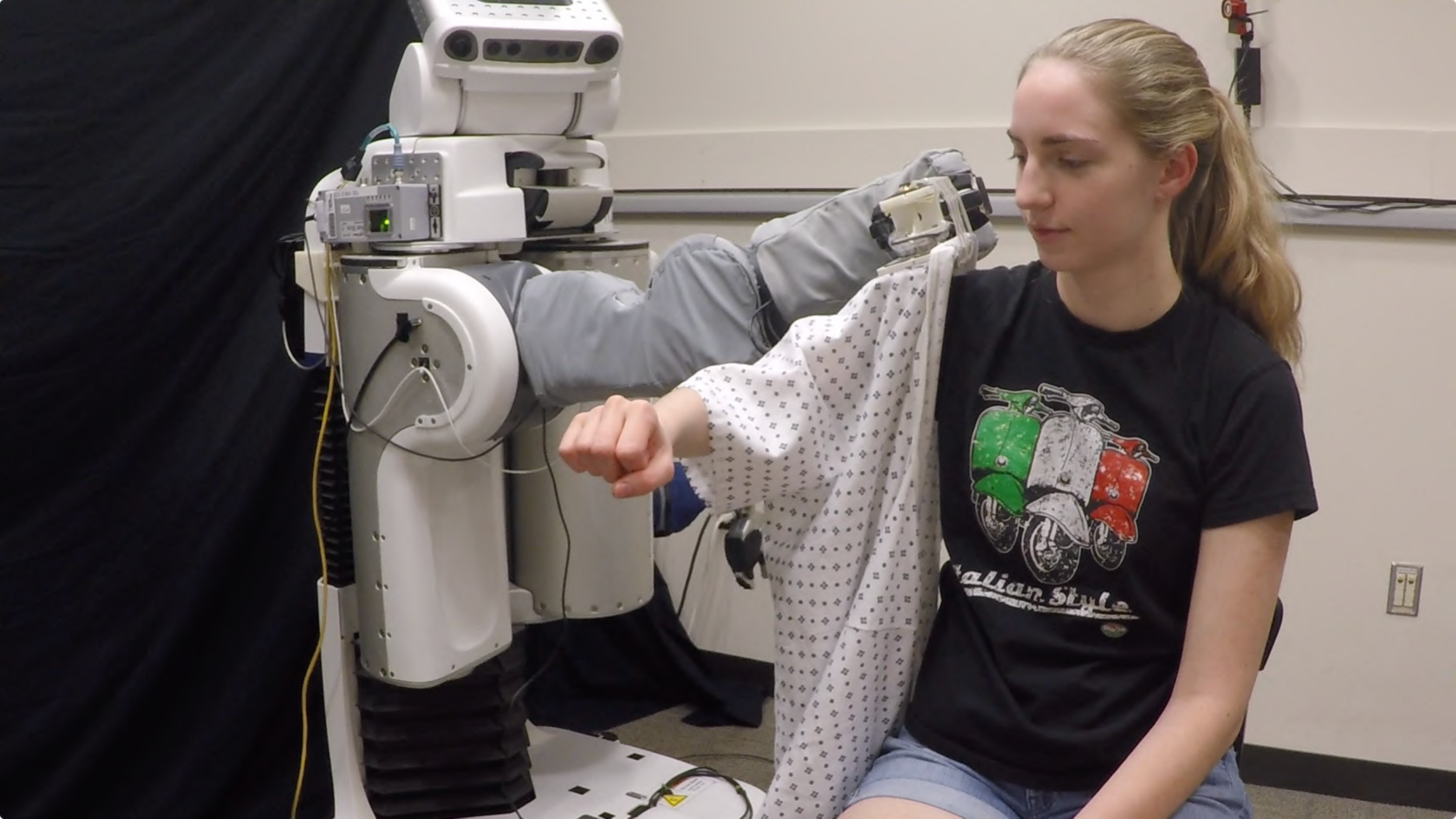}
\caption{\label{fig:dressing}The robot using capacitive servoing to sense and adapt to a participant's lateral arm motions during dressing assistance with a hospital gown sleeve. The PR2's end effector tracked both translational movements and orientation changes in the participant's arm pose in real time.}
\end{figure*}

\begin{figure}
\centering
\includegraphics[width=0.23\textwidth, trim={0cm 0cm 0cm 0cm}, clip]{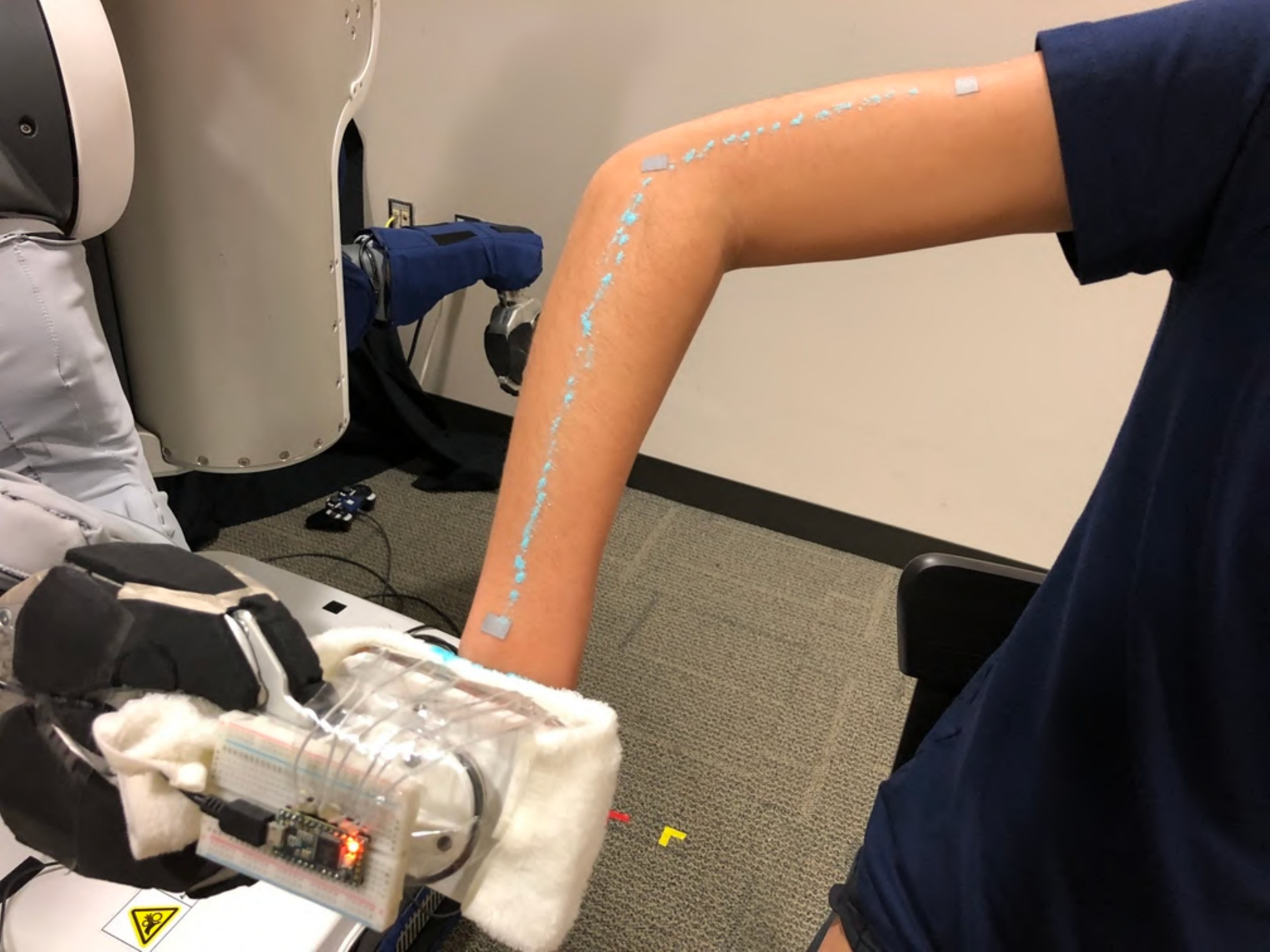}
\includegraphics[width=0.23\textwidth, trim={0cm 0cm 0cm 0cm}, clip]{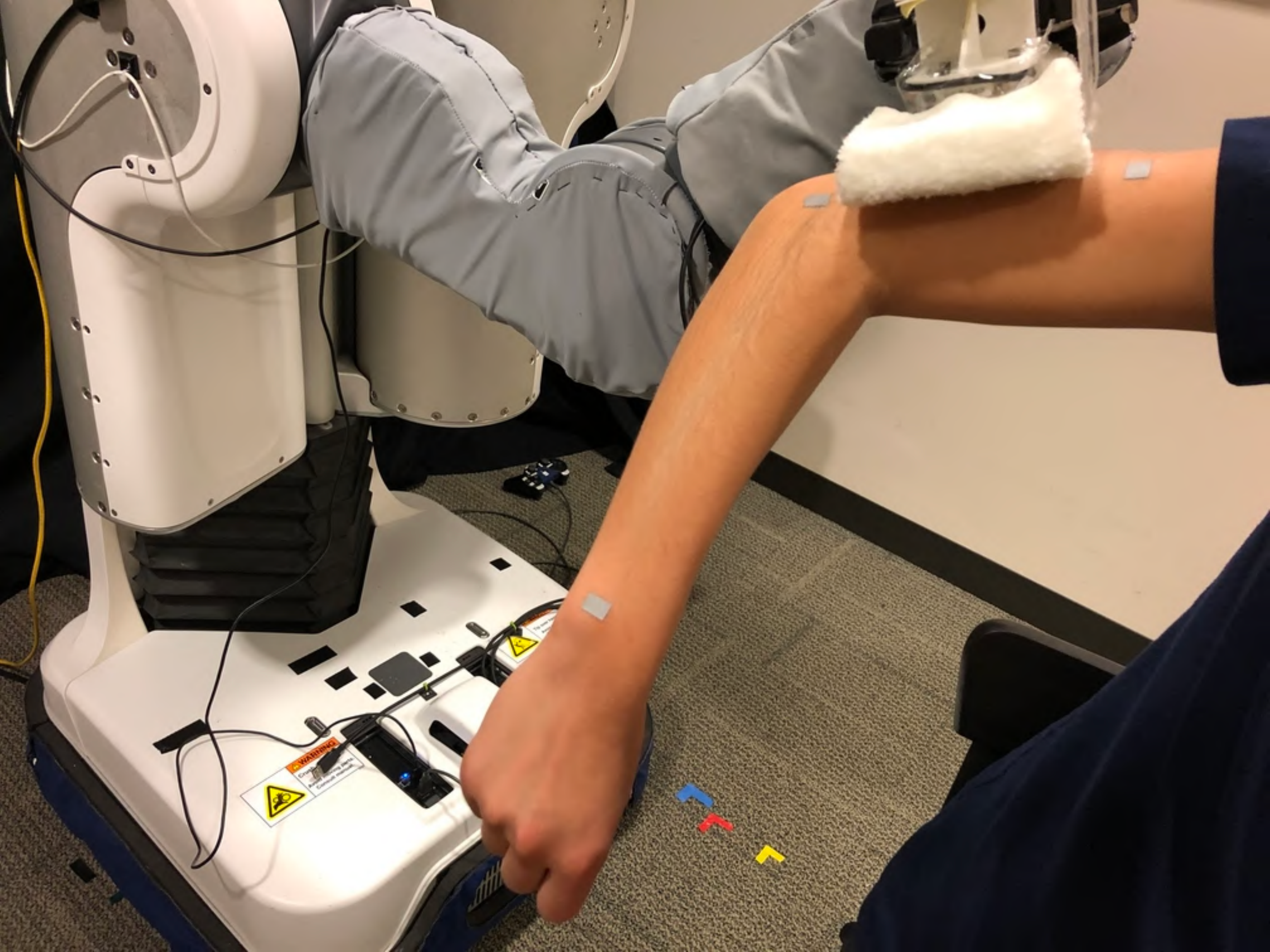}\\
\caption{\label{fig:bathing}A bathing task during which the PR2 used a wet washcloth to clean off blue powder from a participant's arm. The participant's elbow was bent 90 degrees and their forearm was tilted 60 degrees towards the ground. Similar to the \emph{Forearm Tilt} task from Section~\ref{sec:servo_around_limbs}, the robot used capacitive servoing to traverse the participant's arm, while maintaining contact with the arm.}
\end{figure}

We conducted two human-robot trials for each combination of task and joint rotation, where a trial is defined successful if the capacitive sensor has navigated past the joint and is within 8~cm of the participant's limb towards the end of the trial. Table~\ref{table:accuracies} provides the success rates for capacitive servoing across the three human limb tracking tasks.
While capacitive servoing worked well for most tasks and limb poses, there are cases in which the feedback controller would fail to track the limb contour correctly. An example of this is shown in Fig.~\ref{fig:knee_failure} where the robot struggled to rotate fast enough around a participant's 90 degree bent knee, resulting in the capacitive sensor navigating away from and losing track of the limb. When failures occurred for the \emph{Bent Elbow} and \emph{Bent Knee} tasks, they were mostly attributable to the feedback controller failing to rotate quickly enough around a sharp bend in a limb joint. In these cases, adaptive or more aggressive control algorithms may be able to rapidly adjust a robot's trajectories to more extreme human poses or human motion.

Figure~\ref{fig:estimated_distances} shows the estimated Euclidean distance $||\bm{\hat{D}}^{}_t||^{}_2$ between the capacitive sensor and human limb, averaged over all trials and joint rotations for each task. On average across all three tasks, the estimated capacitive sensor position remained within 1~cm from its target distance away from the human limb, namely $\bm{p}^{}_{desired} = (0, 5, 0, 0)$.

\subsection{Capacitive Servoing for Robotic Caregiving}
\label{sec:assistive_tasks}

In a prior study \cite{erickson2019multidimensional}, we have demonstrated that capacitive servoing can enable a robot to provide assistance for two activities of daily living: dressing and bathing. In the first part of the study, a PR2 used capacitive servoing to adapt to arm movements made by four able-bodied participants while helping dress on the sleeve of a hospital gown. Participants performed vertical and lateral arm motions by translating their arm and bending their elbow such that their hand remained within 20~cm of their starting pose. Fig.~\ref{fig:dressing} depicts an image sequence of the PR2 using capacitive servoing to track a participant's lateral arm motions during dressing assistance. Despite arm motions, the estimated capacitive sensor pose $\bm{\hat{p}}^{}_t$ remained on average within 3~cm from the desired pose $\bm{p}^{}_{desired}$ above each participants' arm throughout the duration of each trial.

This prior research also investigated how capacitive servoing can enable a robot to assist with bathing, a task that requires continuous contact with a person's body. The robot wiped off a blue powder from the top surface of participants' arms and legs using a wet washcloth which we attached to the bottom of the capacitive sensor array. Because of the change in electrical conductivity between air and wet cloth, we repeated data collection and trained a new pose estimation model (Sections~\ref{sec:learning} and \ref{sec:data_collection}) to better calibrate pose estimates for bathing assistance. Fig.~\ref{fig:bathing} demonstrates the robot using capacitive servoing to bathe a participant's arm. The capacitive sensor and washcloth moved proximally from the wrist to shoulder, while sensing and rotating around the elbow. In total, the robot succeeded to clean all powder off each participant's arm in 7 of the 8 trials from the study.



\section{Discussion}

Overall, capacitive servoing has shown to be a generalizable method for traversing a robot's end effector along the contours of human limbs and adapting to human limb motion. Yet, it is worth noting that the magnitude of capacitance measurements do vary depending on the circumference of a human limb. For both of the arm navigation tasks (\emph{Bent Elbow} and \emph{Forearm Tilt}) in Section~\ref{sec:servo_around_limbs}, the capacitive sensor array was on average slightly less than 5~cm above the arm. However, the sensor was on average slightly greater than 5~cm from the leg during the \emph{Bent Knee} task. This can be attributed to the larger surface area of a human leg. Based on the capacitance equation for a parallel plate capacitor $C=\frac{k\varepsilon_0A}{d}$, the capacitance $C$ is approximately proportional to the relative surface area $A$ of the electrode and sensed object (limb) (where $k, \varepsilon_0$ are constants and $d$ is the separation distance). Capacitance measurements typically have greater magnitude near the leg, causing the pose estimator to believe the sensor is closer to the limb than it actually is. For example, based on the data used to train our pose estimator (Section~\ref{sec:data_collection}), capacitance measurements had a 44.5\% increase in magnitude when the capacitive sensor array was 5~cm above the participant's leg as compared to above the arm. Future work could explore providing the pose estimator with a priori knowledge on the nearest limb segment (forearm, shin, thigh, etc.) or approximated limb circumference to help resolve this ambiguity.

Techniques for simulating capacitance signals around the human body may also serve as a way to improve capacitive servoing methods in future work. With the accessibility of physics simulation for physical human-robot interaction (see Assistive Gym~\cite{erickson2020assistive}), simulated capacitance measurements can serve as a feature for safely learning intelligent robot controllers across a diverse array of human body shapes and sizes. As a step towards simulated capacitive sensors, Clegg et al.~\cite{clegg2020learning} introduced a simulated multidimensional distance sensor for measuring the shortest Cartesian distance to the human body within a simulated robot-assisted dressing environment. 

In addition to being mounted to assistive tools, as shown in this work, capacitive sensors also provide significant flexibility in where they are mounted on a robot. For robotic caregiving tasks where a robot's end effector is in close proximity to the human body, capacitive sensors can be adhered directly to the outer surface of a robot's gripper. For example, several prior studies have integrated capacitive sensors onto anthropomorphic robotic hands for object manipulation~\cite{goger2013tactile, muhlbacher2015responsive}.

In this article, we evaluate capacitive servoing for controlling a robot's end effector along human limbs. Limbs provide an assistive robot an inherent direction of motion for which to move the robot's end effector along (e.g. from hip to ankle). Further development of capacitive servoing may also be extended to navigating a robot around the human torso. Doing so would present new challenges in adapting to the large distribution of human torso sizes and geometries, and the fusion of additional sensory modalities to replace the directional information that is inherent to limbs.



\section{Conclusion}

This article presents a capacitive servoing control scheme for robots to interact with the human body, including estimating relative human limb pose, navigating along human limbs, and tracking human motion while providing physical assistance for robotic caregiving tasks. 

We presented a design overview of multi-electrode capacitive sensors that can sense human limb pose when mounted to a robot end effector. Given a six-electrode capacitive sensor array, we also described a formal method for collecting capacitance data around human limbs to train human pose estimation models. Capacitive servoing couples these human pose estimates with feedback control, which enables a robot to sense and adapt to human limb pose during physical human-robot interaction.

Through a human-robot study with 12 human participants, we investigated the sensing ranges and pose estimation accuracy of a multidimensional capacitive sensor and human limb pose estimator. The trained pose estimation model provided accurate 4D limb pose estimates when the capacitive sensor was within 15~cm away from the limb and up to 30$^{\circ}$ from the orientation of the limb. Our results also indicated that pose estimation models used in capacitive servoing generalize well across varying human body sizes. With leave-one-participant-out cross-validation, we observed that a pose estimation model trained on capacitance measurements from only one human participant performed on-par with models trained on data from multiple human participants. Capacitive servoing runs in real time, and in our studies was able to compute pose estimates of the human limb at frequencies over 100~Hz, using only the robot's on-board CPUs. As part of our evaluation, a PR2 robot used multidimensional capacitive servoing to move its end effector proximally and distally along a human limb, which is the fundamental task for many robotic caregiving scenarios, including dressing and bathing assistance.

\appendix
\subsection{Data Collection}

To collect a dataset of capacitance measurements for training a pose estimator, we first position the robot's end effector and capacitive sensor above the participant's limb (Data Collection~\ref{alg:datacollect}, line~\ref{line:1_2}) with an initial relative pose of $\bm{p}^{}_0 = (0, 0, 0, 0)$. The robot begins by selecting a target end effector pose above the limb (line~\ref{line:1_6}), $\bm{p}^{}_T = (D^{}_{T,y}, D^{}_{T,z}, \theta^{}_{T,y}, \theta^{}_{T,z})$. In our implementation, both the position and orientation are selected from a uniform distribution such that $D^{}_{T,y}\in[-10~\text{cm},10~\text{cm}]$, $D^{}_{T,z}\in[0~\text{cm},15~\text{cm}]$, and $\theta^{}_{T,y},\theta^{}_{T,z}\in[-\frac{\pi}{8},\frac{\pi}{8}]$. We denote this bounded space of end effector poses above the person's static limb as $S$, which is depicted in Fig.~\ref{fig:datacollection}. The robot then selects target translational and rotational velocities for its end effector (lines~\ref{line:1_7}-\ref{line:1_8}) as it moves to the target pose $\bm{p}^{}_T$. We selected translational velocities $\bm{v}^{}_D$ and rotational velocities $\bm{v}^{}_\theta$ from a uniform distribution such that $||\bm{v}^{}_D||_2 \in [3~\text{cm/s}, 10~\text{cm/s}]$ and $||\bm{v}^{}_\theta||_2 \in [\frac{\pi}{20}, \frac{\pi}{8}]$ in radian/s.

We then compute the per time step change in the robot end effector's Cartesian position $\Delta\bm{\omega}$ and Euler orientation $\Delta\bm{\phi}$ (lines~\ref{line:1_9}-\ref{line:1_13}). $\Delta\bm{\omega}$ and $\Delta\bm{\phi}$ are defined such that the end effector progresses to the target pose $\bm{p}^{}_T$ with velocities $\bm{v}^{}_D$ and $\bm{v}^{}_\theta$ (lines~\ref{line:1_12}-\ref{line:1_13}), where $\odot$ defines the Hadamard product.

\begin{datacollect}[t]
\caption{Capacitive Sensor Data Collection for Pose Estimation}\label{alg:datacollect}
\begin{algorithmic}[1]
\State \textbf{Given:} $N$: number of trajectories,\newline
$\bm{p}_0$: initial pose,\newline
$S$: space of end effector poses,\newline
$\tau$: data collection frequency.
\State Move capacitive sensor to top of limb; to $\bm{p}_0$.\label{line:1_2}
\State $\bm{\omega}^{}_0, \bm{\phi}^{}_0 \gets$ GetCurrentEndEffectorPose().
\State $\mathcal{D} \gets \{\}$
\For{$i=1,\ldots, N$}\label{line:1_5}
    \State Select target $\bm{p}^{}_T \gets (D^{}_{T,y}, D^{}_{T,z}, \theta^{}_{T,y}, \theta^{}_{T,z})\in S$.\label{line:1_6}
    \State Select translational velocities $\bm{v}^{}_D \gets (0, v^{}_{D_y}, v^{}_{D_z})$.\label{line:1_7}
    \State Select rotational velocities $\bm{v}^{}_\theta \gets (0, v^{}_{\theta_y}, v^{}_{\theta_z})$.\label{line:1_8}
    
    \State $\bm{\omega}^{}_I, \bm{\phi}^{}_I \gets$ GetCurrentEndEffectorPose().\label{line:1_9}
    \State $\bm{\omega}^{}_T \gets \bm{\omega}^{}_0 + (0, D^{}_{T,y}, D^{}_{T,z})$.
    \State $\bm{\phi}^{}_T \gets \bm{\phi}^{}_0 + (0, \theta^{}_{T,y}, \theta^{}_{T,z})$.
    \State $\Delta\bm{\omega} \gets (\bm{\omega}^{}_T - \bm{\omega}^{}_I) \odot \frac{\bm{v}_D}{\tau}$.\label{line:1_12}
    \State $\Delta\bm{\phi} \gets (\bm{\phi}^{}_T - \bm{\phi}^{}_I) \odot \frac{\bm{v}_\theta}{\tau}$.\label{line:1_13}
    \State $\delta \gets \{\}$.
    \State $t \gets 1$.
    \While{$\bm{p}^{}_{t-1} \not\approx \bm{p}^{}_T$}
        \State $\bm{c}^{}_t \gets$ GetCapacitanceMeasurements().\label{line:1_17}
        \State $\bm{\omega}^{}_t, \bm{\phi}^{}_t \gets$ GetCurrentEndEffectorPose().\label{line:1_18}
        \State ($D^{}_{t,x}, D^{}_{t,y}, D^{}_{t,z}) \gets \bm{\omega}^{}_t - \bm{\omega}^{}_0$.
        \State $(\theta^{}_{t,x}, \theta^{}_{t,y}, \theta^{}_{t,z}) \gets \bm{\phi}^{}_t - \bm{\phi}^{}_0$.
        \State $\bm{p}^{}_t \gets (D^{}_{t,y}, D^{}_{t,z}, \theta^{}_{t,y}, \theta^{}_{t,z})$.\label{line:1_21}
        \State $\delta \gets \delta \cup \{(\bm{c}^{}_t, \bm{p}^{}_t)\}$.
        
        \State $\bm{\omega}^*_t \gets \bm{\omega}^{}_I+t\Delta\bm{\omega}$\label{line:1_23}
        \State $\bm{\phi}^*_t \gets \bm{\phi}^{}_I+t\Delta\bm{\phi}$
        \If{$\bm{\omega}^{}_t \approx \bm{\omega}^{}_T$}
            \State $\bm{\omega}^*_t \gets \bm{\omega}^{}_T$
        \ElsIf{$\bm{\phi}^{}_t \approx \bm{\phi}^{}_T$}
            \State $\bm{\phi}^*_t \gets \bm{\phi}^{}_T$
        \EndIf
        \State $\bm{\alpha}^{}_T \gets$ InverseKinematics($\bm{\omega}^*_t, \bm{\phi}^*_t$).
        \State SendToActuators($\bm{\alpha}^{}_t$).\label{line:1_30}
        \State $t \gets t+1$
    \EndWhile
    \State $\mathcal{D} \gets \mathcal{D} \cup \{\delta\}$\label{line:1_32}
\EndFor
\State \Return dataset $\mathcal{D}$
\end{algorithmic}
\end{datacollect}

For each time step $t$ during data collection, the robot records a measurement from all capacitive sensor electrodes $\bm{c}^{}_t$ (line~\ref{line:1_17}) and the current pose difference between the sensor and human limb $\bm{p}^{}_t$, which is calculated via forward kinematics (lines~\ref{line:1_18}-\ref{line:1_21}). Using inverse kinematics, the robot then executes a single step action (lines~\ref{line:1_23}-\ref{line:1_30}) to move the capacitive sensor towards the target state $\bm{p}^{}_T$ along a linear trajectory with velocities $\bm{v}^{}_D$ and $\bm{v}^{}_\theta$. Once the robot reaches the target state $\bm{p}^{}_{t} \approx \bm{p}^{}_T$, we store all recorded data from the trajectory into a dataset $\mathcal{D}$ (line~\ref{line:1_32}) and have the robot progress to the next randomly generated target pose for a total of $N=500$ trajectories (line~\ref{line:1_5}). This data collection process is then repeated for three locations above the arm (wrist, forearm, upper arm) and three locations above the leg (ankle, shin, knee), which resulted in a total of $N=3000$ end effector trajectories used to collect capacitance data around a participant's limbs.



\bibliographystyle{IEEEtran}
\bibliography{bibliography}

\begin{IEEEbiography}[{\includegraphics[width=1in,height=1.25in,clip,keepaspectratio]{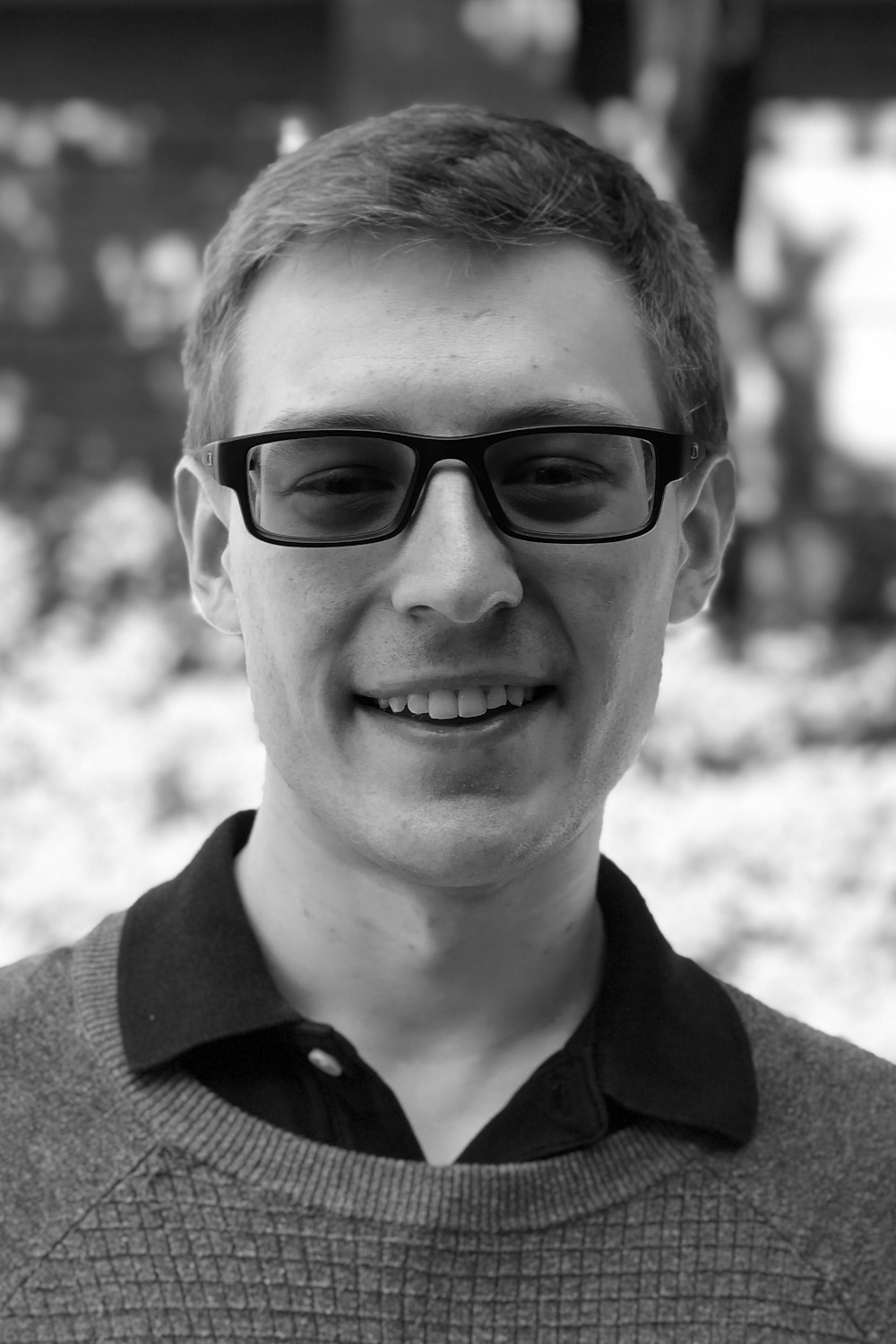}}]{Zackory Erickson} is an Assistant Professor in The Robotics Institute at Carnegie Mellon University. He received his Ph.D. in Robotics from the Georgia Institute of Technology. His work spans physical human-robot interaction, healthcare robotics, robot learning, physics simulation, multimodal perception, and mobile manipulation. His work has won the Best Student Paper Award at ICORR 2019 and a Best Paper in Service Robotics finalist at ICRA 2019.
\end{IEEEbiography}

\begin{IEEEbiography}[{\includegraphics[width=1in,height=1.25in,clip,keepaspectratio]{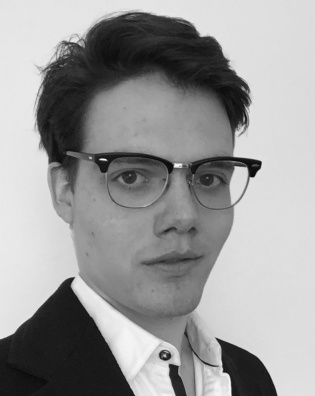}}]{Henry M. Clever}
received a B.S. in mechanical engineering from the University of Kansas, and a M.S. in mechanical engineering from New York University. Henry is currently pursuing a Ph.D. in robotics at the Georgia Institute of Technology in the Healthcare Robotics Lab. Henry's research interests include robot understanding in unstructured environments, haptic and vision perception of humans and robots, human-robot systems, physics simulation of humans and robots, and human pose estimation.
\end{IEEEbiography}

\begin{IEEEbiography}[{\includegraphics[width=1in,height=1.25in,clip,keepaspectratio]{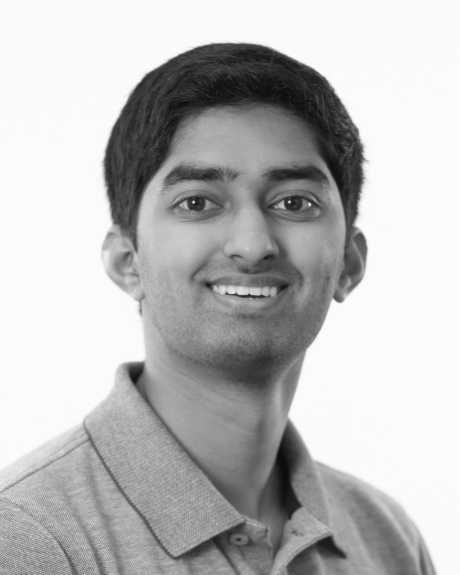}}]{Vamsee Gangaram}
received his Bachelors in Computer Science from Georgia Institute of Technology in 2020. He recently joined Microsoft Surface as a software engineer. His areas of work include robotics, simulation, and firmware.
\end{IEEEbiography}
\newpage

\begin{IEEEbiography}[{\includegraphics[width=1in,height=1.25in,clip,keepaspectratio]{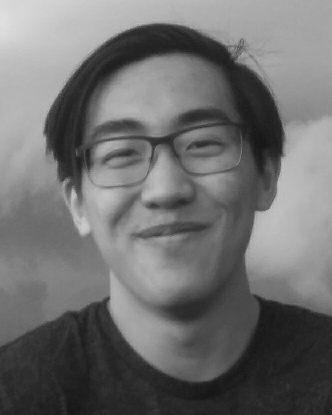}}]{Eliot Xing}
is an undergraduate at the Georgia Institute of Technology, pursuing B.S degrees in computer engineering and mathematics. His research interests lie in developing learning methods that enable robots to perceive and interact in the diverse real world, with minimal human supervision. 
\end{IEEEbiography}

\begin{IEEEbiography}[{\includegraphics[width=1in,height=1.25in,clip,keepaspectratio]{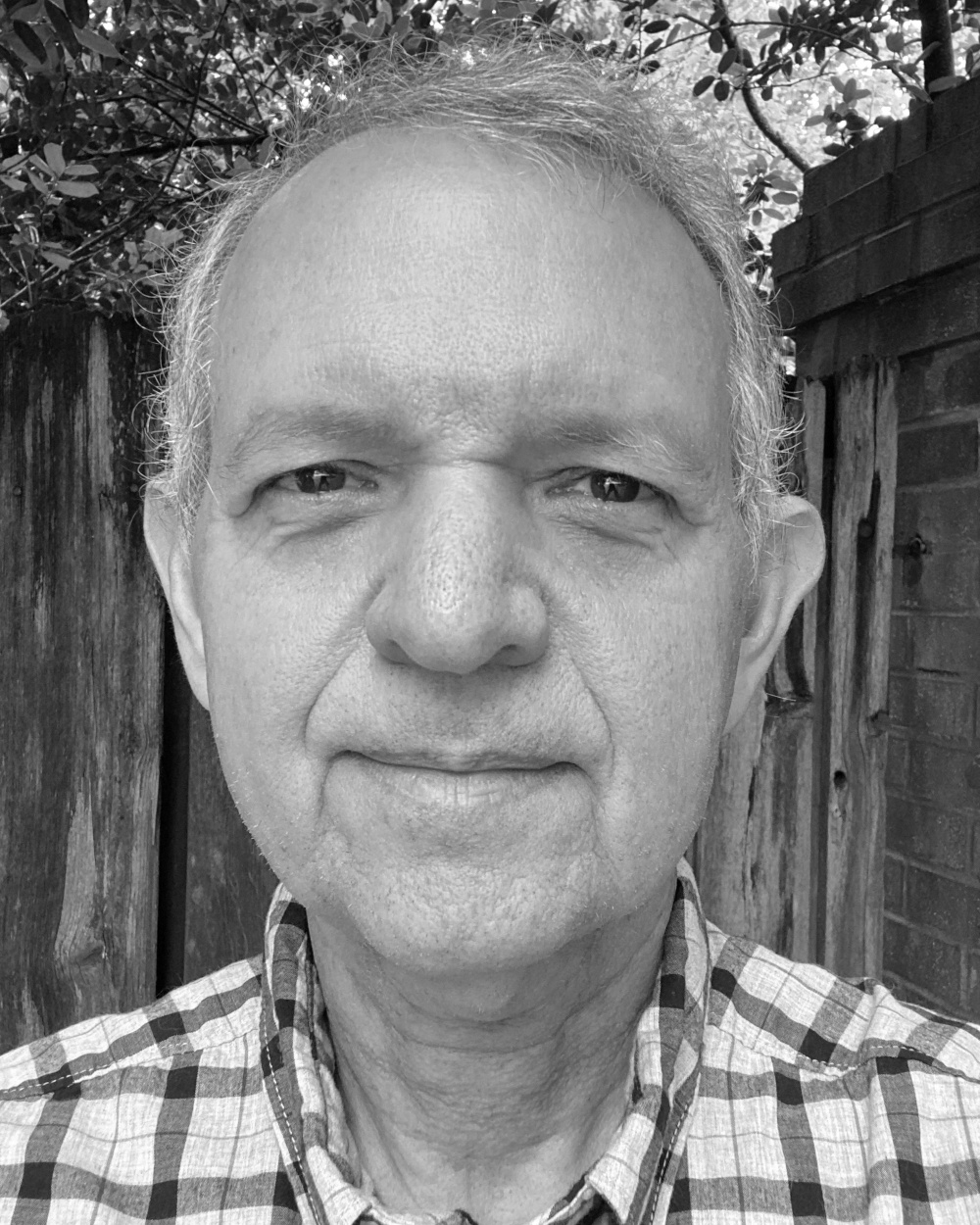}}]{Greg Turk}
received a Ph.D. in computer science in 1992 from the University of
North Carolina at Chapel Hill.  He was a postdoctoral researcher at Stanford
University for two years.  He is currently a Professor at the
Georgia Institute of Technology, where he is a member of the School of
Interactive Computing and the Graphics, Visualization and Usability Center.
His research interests include computer graphics, robotics, biological
simulation and machine learning.  He was the Technical Papers Chair for ACM
SIGGRAPH 2008.  In 2012 he received the Computer Graphics Achievement Award
from ACM SIGGRAPH for his computer graphics research. 
\end{IEEEbiography}

\begin{IEEEbiography}[{\includegraphics[width=1in,height=1.25in,clip,keepaspectratio]{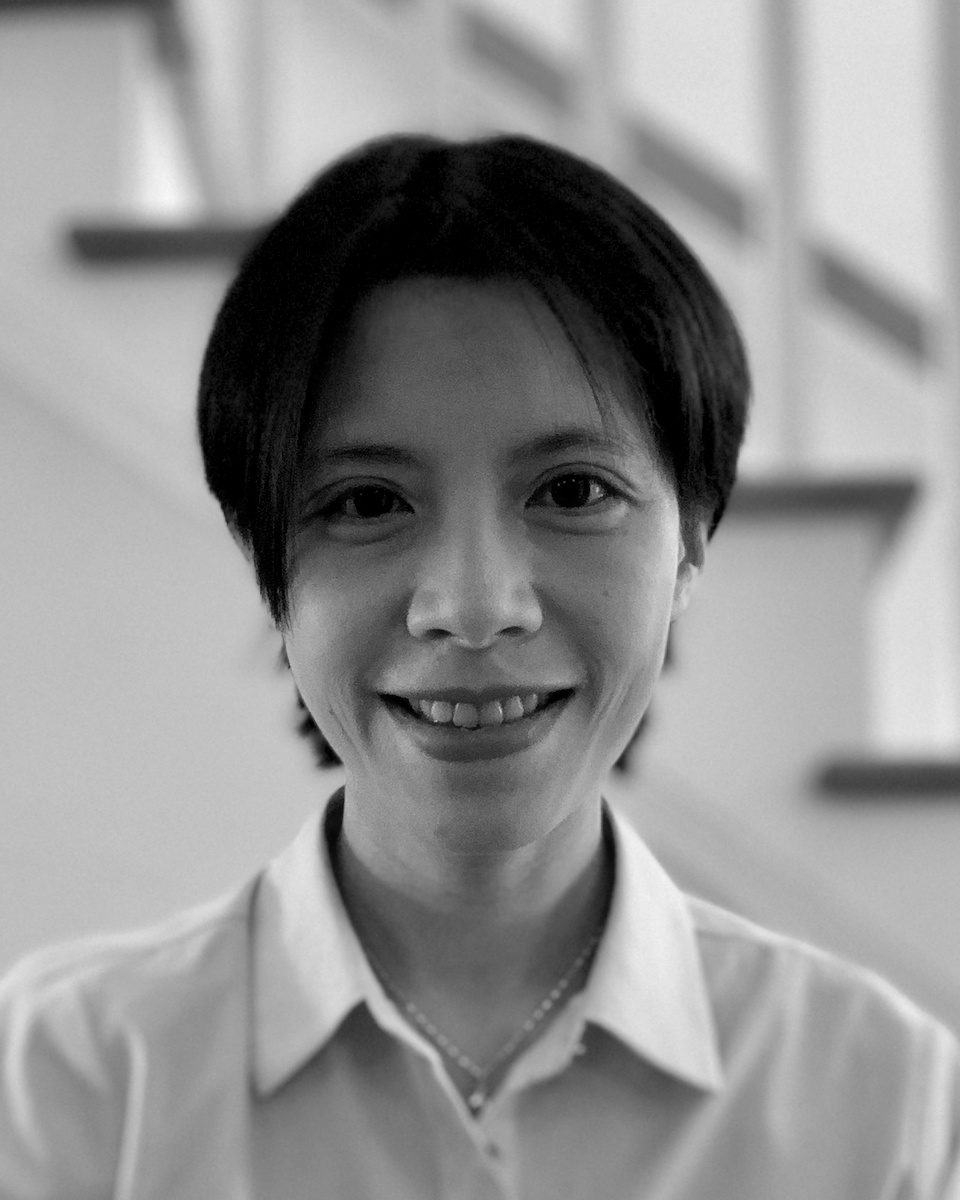}}]{C. Karen Liu}
is an associate professor in the Computer Science Department at Stanford University. She received her Ph.D. degree in Computer Science from the University of Washington. Liu's research interests are in computer graphics and robotics, including physics-based animation, character animation, optimal control, reinforcement learning, and computational biomechanics. She developed computational approaches to modeling realistic and natural human movements, learning complex control policies for humanoids and assistive robots, and advancing fundamental numerical simulation and optimal control algorithms. The algorithms and software developed in her lab have fostered interdisciplinary collaboration with researchers in robotics, computer graphics, mechanical engineering, biomechanics, neuroscience, and biology. Liu received a National Science Foundation CAREER Award, an Alfred P. Sloan Fellowship, and was named Young Innovators Under 35 by Technology Review. In 2012, Liu received the ACM SIGGRAPH Significant New Researcher Award for her contribution in the field of computer graphics.
\end{IEEEbiography}

\begin{IEEEbiography}[{\includegraphics[width=1in,height=1.25in,clip,keepaspectratio]{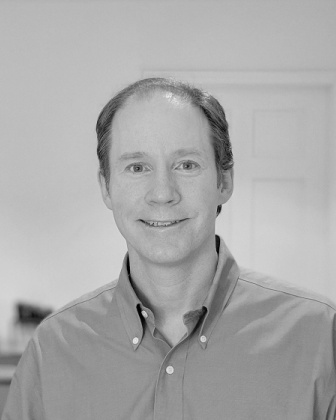}}]{Charles C. Kemp} (Charlie Kemp) is an Associate Professor at Georgia Tech in the Department of Biomedical Engineering with adjunct appointments in the School of Interactive Computing and the School of Electrical and Computer Engineering. In 2007, he founded the Healthcare Robotics Lab, which focuses on enabling robots to provide intelligent physical assistance in the context of healthcare. He earned a BS, an MEng, and a PhD from the Massachusetts Institute of Technology (MIT) in the areas of computer science and electrical engineering. 
\end{IEEEbiography}

\enlargethispage{-3cm}



\end{document}